\pdfoutput=1

\documentclass[11pt]{article}

\usepackage[margin=1in]{geometry}

\usepackage[T1]{fontenc}
\usepackage[utf8]{inputenc}
\usepackage{times}
\usepackage{latexsym}
\usepackage{microtype}
\usepackage{inconsolata}

\usepackage{amsmath}
\usepackage{amssymb}

\usepackage{graphicx}
\usepackage{booktabs}
\usepackage{multirow}
\usepackage{subfig}
\usepackage{wrapfig}

\usepackage[ruled,vlined,linesnumbered]{algorithm2e}
\usepackage{listings}

\usepackage[authoryear,round]{natbib}
\usepackage{xurl}
\usepackage[hidelinks]{hyperref}

\title{Detect Before You Leap: Mirage Detection in Vision--Language Models}

\author{
Md. Shaown Miah\thanks{Miah and Chowdhury contributed equally. Chowdhury also served as the faculty advisor for this work.}\\
Bangladesh University of Engineering and Technology\\
\texttt{1918018@bme.buet.ac.bd}
\and
S. M. Taiabul Haque\\
BRAC University\\
\texttt{taiabul.haque@bracu.ac.bd}
\and
Syed Ishtiaque Ahmed\\
University of Toronto\\
\texttt{ishtiaque@cs.toronto.edu}
\and
Sayeed Shafayet Chowdhury\footnotemark[1]\\
Indiana University Indianapolis\\
\texttt{saychow@iu.edu}
\and
}

\date{}
\begin{document}

\maketitle
\begin{abstract}
Vision--language models (VLMs) can produce confident answers without relevant visual evidence, a
failure mode known as \emph{mirage}~\citep{mirage2026}. We study pre-release mirage detection:
deciding whether a VLM answer should be released or withheld. Our model-agnostic method, \underline{T}ext-
\underline{C}onditioned \underline{L}ayer-wise \underline{I}nternal \underline{A}lignment (TC-LIA), tracks question--image alignment across the
layers of a frozen CLIP ViT-H/14 encoder, summarizing patch--text alignment by final similarity, late-
layer top-$k$ alignment, early-to-late gain, and slope. TC-LIA is training-free at deployment with fixed projections
and scoring weights, without any label-specific training, and delivers strong detection independently.
Additionally, when combined with blank/noise detection, domain routing, and VLM self-assessment,
it forms an ensemble whose supervised training improves performance but is an optional add-on. On
19,004 samples spanning 10 VQA domains, 14 state-of-the-art VLMs exhibit 57.3--75.0\% base mirage
rates. Our proposed TC-LIA alone cuts this to 7.5\% with 83.5\% Related/Unrelated/Blank-Noise
classification accuracy, and the ensemble reaches 84.5--88.4\% accuracy with 5.9--7.2\% mirage rates
(best joint result: 88.4\% accuracy, 6.4\% mirage rate). Notably, an ensemble trained on a single
backbone transfers well to unseen backbones, with the best-transferring source staying within 0.7\%
accuracy points of per-backbone training across 13 held-out VLMs.

\end{abstract}

\section{Introduction}

Multimodal models are increasingly used to answer questions about images,
radiographs, pathology slides, documents, and scene photographs, under the
implicit assumption that answers are grounded in the supplied image.
Recent work on \emph{mirage reasoning} challenges this assumption by showing that VLMs can
generate confident diagnoses even when images are absent, blank, or
irrelevant~\citep{mirage2026}, a failure most worrying in high-stakes settings,
where a missing or mismatched input produces a fluent but unsupported answer.

This motivates \emph{runtime mirage detection}, which involves deciding \emph{before release}
whether a VLM should answer a given image--question pair or abstain.
Given an image $x$ and question $q$, we classify the pair as
\textsc{Related}, \textsc{Unrelated-Real}, or \textsc{Blank/Noise}.
Conventional hallucination detection evaluates a completed answer, whereas we 
determine whether the visual input is appropriate \emph{a priori}.
\textsc{Out-of-Domain Unrelated} images are easy negatives, while
\textsc{In-Domain Unrelated} images share the question's same visual domain but show
different content, making them harder instance-level mismatches.

To resolve the release versus abstention question, we propose \textsc{TC-LIA}, a \underline{T}ext-
\underline{C}onditioned \underline{L}ayer-wise \underline{I}nternal \underline{A}lignment
method that exploits CLIP-style image--text representations and ViT patch
tokens~\citep{radford2021clip,dosovitskiy2021vit}.
Instead of relying solely on final global image--text similarity,
\textsc{TC-LIA} tracks how question-relevant visual evidence develops across the
vision encoder. We project patch tokens into the final CLIP embedding space
and compare with the question embedding at each layer, and the trajectory is
summarized by late-layer top-$k$ patch alignment, early-to-late gain, and slope. Our hypothesis is that
related pairs should develop increasingly localized, question-relevant alignment
in deeper layers, whereas unrelated or blank inputs show weaker or flatter
trajectories.

Our complete detector combines five signals:
(1) pixel-statistic blank/noise detection,
(2) zero-shot CLIP-based domain routing,
(3) \textsc{TC-LIA} alignment features,
(4) structured VLM self-assessment, and
(5) boosting fusion.
Self-assessment is only one of these five signals. The detector does not rely on the VLM refusing its own answer. Blank and noise detection also serves only as a simple pixel-level gate. The central challenge is \textsc{Unrelated-Real} detection. This is especially difficult for in-domain images that appear valid but contain no evidence relevant to the question. 

We evaluate on 19,004 records spanning ten diverse VQA domains (medical imaging,
document understanding, scene text, multi-discipline reasoning, and video
understanding) across 14 VLM backbones, from open-source models to
frontier systems. Base-prompt mirage rates span $57.3$--$75.0\%$. Notably,
\textsc{TC-LIA} alone reduces the rate to $7.5\%$ at $83.5\%$ accuracy, and the
full ensemble achieves $84.5$--$88.4\%$ accuracy with $5.9$--$7.2\%$ mirage
rates. We perform further diagnostic analyses to compare \textsc{TC-LIA} with final CLIP similarity,
GradCAM-style saliency, and prompting-based self-assessment. \textsc{TC-LIA} is fully training-free at deployment. It uses a frozen CLIP encoder, fixed projections, hand-set weights, and no labeled training data. It also detects mirages well on its own. The supervised ensemble further improves performance, but \textsc{TC-LIA} does not depend on it. When trained on a single VLM, the ensemble also transfers well to unseen backbones. The best source model stays 
within 0.7\% points of per-backbone training on the 13 unseen
backbones (Appendix~\ref{app:leave_all_but_one}).

\vspace{0mm}
\textbf{Contributions.}\\
(a) We formulate \emph{runtime mirage detection} as a pre-release decision
problem and introduce a 19,004-record benchmark spanning ten VQA domains and
four input conditions, including hard in-domain unrelated examples.

(b) We propose \textsc{TC-LIA}, a model-agnostic layer-wise alignment method
that uses intermediate CLIP ViT-H/14 patch tokens to quantify the emergence of
question-relevant visual evidence through interpretable features.

(c) We develop an ensemble detector combining pixel statistics, CLIP-based
domain routing, \textsc{TC-LIA}, and structured VLM self-assessment through
boosting fusion. It reduces base mirage rates from $57.3$--$75.0\%$ to
$5.9$--$7.2\%$ at $84.5$--$88.4\%$ accuracy across 14 state-of-the-art VLMs.

\begin{figure*}[t]
\centering
\includegraphics[width=\textwidth,height=0.52\textheight,keepaspectratio]{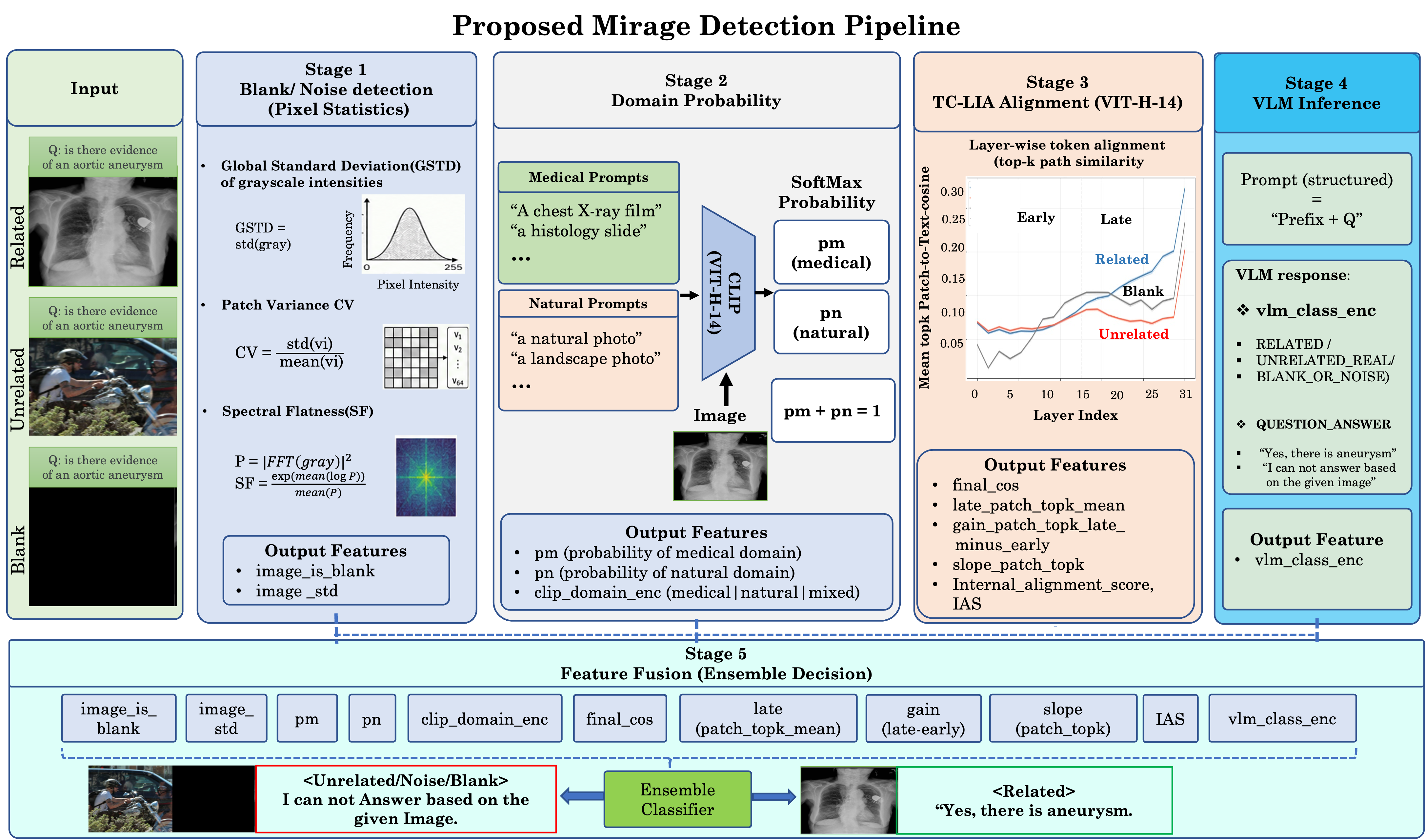}
\vspace{-6mm}
\caption{
Runtime pipeline: pixel statistics detect blank/noise inputs, CLIP-based routing
identifies the visual domain, \textsc{TC-LIA} measures layer-wise question--image
alignment, structured VLM self-assessment adds a further signal; a boosting
ensemble passes only \textsc{Related} pairs while rejecting
\textsc{Unrelated-Real} and \textsc{Blank/Noise}.
}
\label{fig:pipeline}
\vspace{-3mm}
\end{figure*}

\section{Related Work}

VLMs can generate fluent answers that are poorly grounded in the supplied image. POPE evaluates whether models mention objects unsupported by the image, and broader surveys examine the causes, benchmarks, and mitigation of multimodal hallucination \citep{li2023pope,bai2024hallucination}. VQA studies show that models often rely on language priors and dataset artifacts instead of visual evidence \citep{goyal2017making,agrawal2018dont}; and Winoground demonstrates failures in grounded compositional reasoning even in strong vision--language models \citep{thrush2022winoground}. Related selective-prediction work improves reliability by abstaining on uncertain inputs \citep{geifman2017selective,kadavath2022language,xiong2024uncertainty}, but answer confidence is not evidence that the image supports the question. These works diagnose errors or uncertainty around generation. On the other hand, we  decide \emph{before} generation whether an image--question pair contains enough evidence to answer.

A dedicated line of work targets unanswerable or unsolvable inputs directly: CLIP-UP injects a CLIP-based answerability prior into the VLM \citep{vardi2026clipup}. HALP probes VLM internal representations to flag likely hallucinations before generation \citep{kogilathota2026halp}. Reliable VQA formalizes risk-controlled abstention for VQA \citep{whitehead2022reliable} and MM-UPD \citep{miyai2025mmupd} and UNK-VQA \citep{guo2024unkvqa} benchmark abstention on unsolvable inputs. These methods share our pre-release goal but differ in evidence source---a single frozen alignment space or the target VLM's hidden states, whereas our detector couples layer-wise alignment trajectories with domain routing, pixel statistics, and structured self-assessment, evaluated across 10 heterogeneous domains with matched unrelated controls.

Intermediate representations capture structured, task-relevant information and can be probed to understand model behavior \citep{alain2016understanding, tenney2019bert, belinkov2022probing}. In vision transformers, semantic and spatial information emerges non-uniformly across layers \citep{caron2021dino,raghu2021vision}. Final CLIP similarity \citep{radford2021clip} is a useful image--text measure but too coarse for localized VQA evidence, particularly in medical, document, and infographic settings. \textsc{TC-LIA} therefore uses question-conditioned patch--text alignment across vision-encoder layers as a runtime detection signal. Grounding methods can localize phrases to image regions \citep{li2022grounded,liu2023groundingdino} but are not designed to determine pre-release abstention. Our framework targets that decision with lightweight layer-wise alignment and domain-routing features, evaluated across pathology, scene-text, document, and infographic domains \citep{he2020pathvqa,singh2019textvqa,mathew2021docvqa,mathew2022infographicvqa}. Furthermore, RAPT and Visual Evidence Augmentation show that VLM decoder layers may attend to relevant visual regions even when the final answer is wrong \citep{liu2026seeingbutnotbelieving}, focusing on how evidence is used \emph{during} answering, whereas we ask whether sufficient evidence exists \emph{before} answering. Our experiments find RAPT-style attention ratios useful as diagnostics but unreliable as standalone detectors (appendix~\ref{rapt-failed}).

\section{Problem Formulation}
\label{sec:problem}
\vspace{-\baselineskip}
Let $x \in \mathcal{X}$ denote an image and $q \in \mathcal{Q}$ a textual question. We assign a label $y\in\mathcal{Y}=\{R,U,B\}$, where $R$ indicates that $x$ is related to $q$ and the VLM may answer, $U$ that $x$ is a real but unrelated image, and $B$ blank or noise input. A mirage detector $g(x,q)$ outputs one of these three labels; if $g(x,q)=R$, a downstream VLM is allowed to answer, and otherwise the system returns a refusal.

\paragraph{Mirage rate.}
We define mirage rate as the fraction of all examples for which the detector incorrectly allows a non-related input to be answered:
\begin{equation}
\mathrm{MR}(g)=
\frac{1}{|\mathcal{D}|}
\sum_{(x,q,y)\in\mathcal{D}}
\mathbf{1}\!\left[y\neq R,\, g(x,q)=R\right].
\label{eq:mirage_rate}
\end{equation}
Here, $R$ denotes \textsc{Related}; $y\neq R$ covers \textsc{Unrelated-Real} and \textsc{Blank/Noise} inputs.
Unlike conventional false-positive rate, the denominator is the entire evaluation set, matching the deployment risk that a random incoming request may be wrongly passed to the VLM.
Because the four input conditions are balanced, non-related inputs constitute exactly $75.0\%$ of the evaluation set, so an unselective model attains $\mathrm{MR}(g)=75.0\%$ by construction---the zero-abstention ceiling---and $\mathrm{MR}(g)$ scales with the negative base rate of the deployment distribution. The conditional false-accept rate on invalid inputs, $P(g(x,q)=R \mid y\neq R)=\tfrac{4}{3}\,\mathrm{MR}(g)$, is recoverable from any reported mirage rate by this factor.


\begin{figure*}[t]
    \centering
    \includegraphics[width=\textwidth,height=0.48\textheight,keepaspectratio]{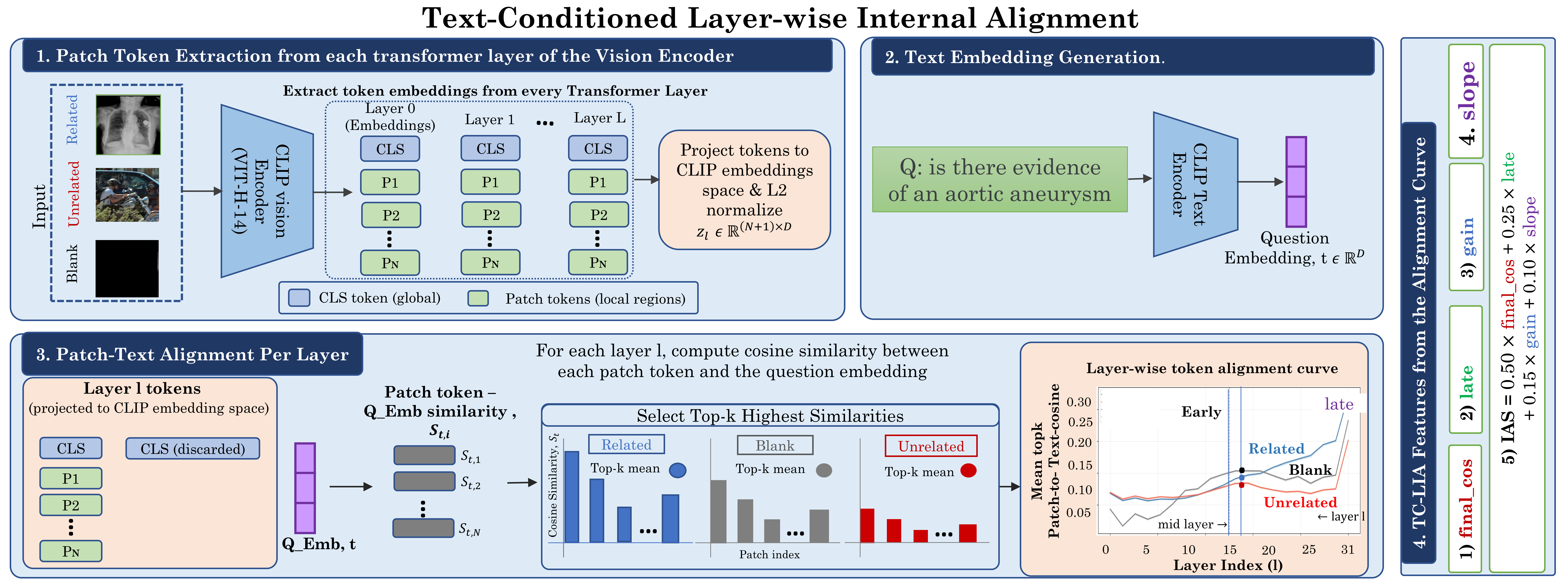}
    \vspace{-5mm}
    \caption{
    Text-conditioned layer-wise patch alignment from CLIP ViT-H/14: late-layer top-$k$ alignment, gain, slope, final cosine, and IAS as mirage detection features.
    }
    \label{fig:tc-lia}
    \vspace{-5mm}
\end{figure*}

\vspace{-4mm}

\section{Method}
\vspace{-4mm}
\label{sec:method}

The proposed system consists of five stages: blank/noise screening, domain-adaptive CLIP routing, TC-LIA feature extraction, structured VLM self-assessment, and ensemble feature fusion. In stage one, blank and noise inputs are detected using three signals, namely global grayscale standard deviation, patch-variance coefficient of variation, and spectral flatness---a high-recall safeguard that removes non-informative inputs before the harder ``related'' vs.\ ``unrelated-real'' distinction.

In stage two, because a single CLIP embedding space may not provide uniformly reliable image-text relevance across very different domains, a coarse routing step separates medical from natural images by computing zero-shot prompt probabilities against separate medical and natural prompt sets, producing two normalized probabilities. The pipeline remains a single-encoder system: these probabilities, plus the encoded routing decision, are strictly tabular ensemble features and never switch the encoder used for TC-LIA extraction, which always uses the same frozen CLIP ViT-H/14 (Appendix~\ref{app:domain_routing}).

In stage three, TC-LIA is extracted. \textsc{TC-LIA} is defined over any frozen contrastive vision--language model exposing per-layer patch tokens and a shared image--text embedding space. The main experiments instantiate it with CLIP ViT-H/14 (Appendix~\ref{app:encoder_gen} justifies the choice, and Section~\ref{sec:results} reports the encoder-swap ablation instantiating the signal with SigLIP-SO400M). For each image, the CLIP vision encoder produces one CLS token and a group of patch tokens at a given layer, alongside a normalized text embedding of the question. Each intermediate token is mapped into the shared image-text space by the terminal layer normalization (\texttt{ln\_post}) followed by the fixed visual projection matrix, and its cosine similarity to the text embedding is computed; this projection is never modified during detector training. For the four multi-image/video domains (38\% of the benchmark), pixel statistics, routing, and TC-LIA features are computed on the first image (for video, the first sampled frame), keeping the feature space identical across input modalities. Here, aggregation across frames or image lists is left to future work. This is not a bottleneck as the deployed detector restricted to those domains (37.3\% of test records) reaches $90.7\%$ accuracy, $3.9\%$ mirage rate, and $0.947$ non-blank AUROC, exceeding the six single-image domains ($85.3\%$, $7.2\%$, $0.896$; Appendix~\ref{app:diagnostics}).

\begin{figure}[t]
\centering
\includegraphics[width=0.85\textwidth]{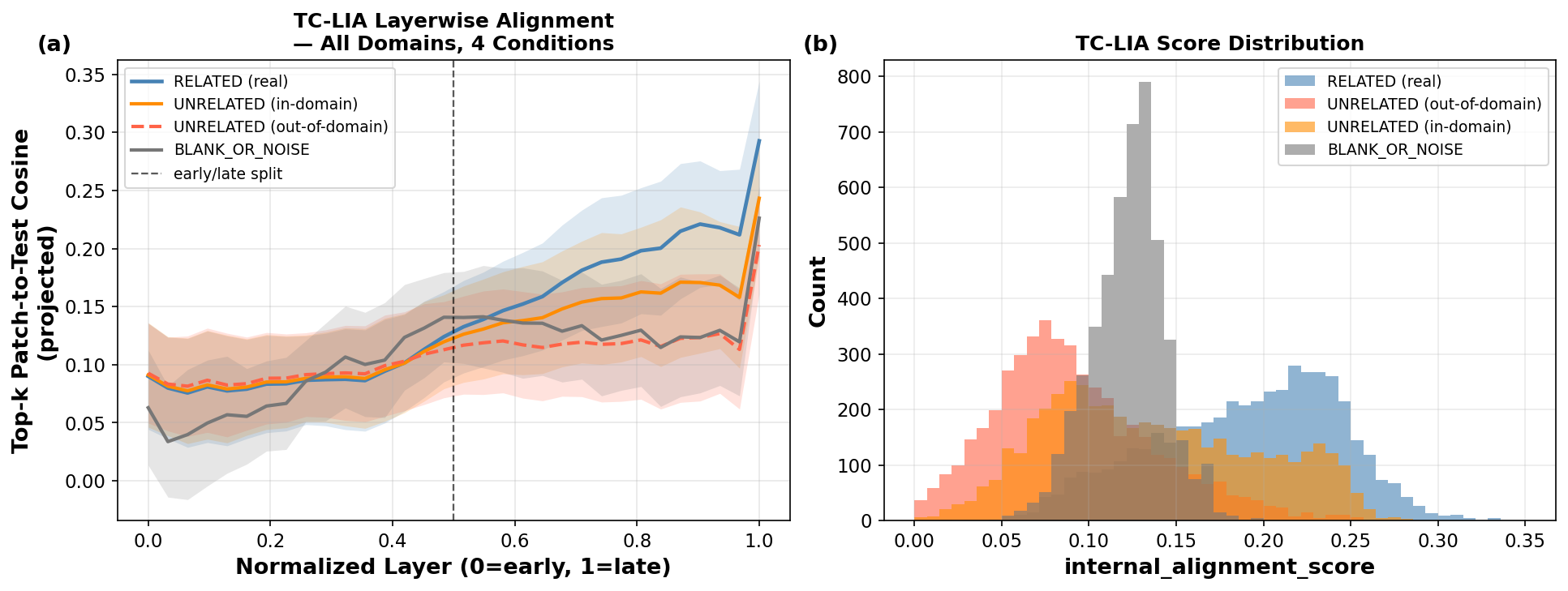}
\caption{
TC-LIA alignment across conditions.
(a) Mean top-$k$ patch--text alignment across domains: stronger late-layer rise for \textsc{Related} than for \textsc{Unrelated-Real} or \textsc{Blank/Noise}.
(b) IAS distributions: related inputs shift higher, non-related lower or intermediate.
}
\label{fig:tc-lia-alignment}
\end{figure}

A natural question is why a projection trained for the \emph{final} visual representation should work on \emph{intermediate} tokens. It is an approximate readout rather than a layer-optimal projection, but question-relevant patch tokens progressively shift toward semantically aligned directions in deeper layers, so TC-LIA relies on relative trends across layers---late-layer alignment, early-to-late gain, slope---rather than calibrated intermediate similarities. Supporting evidence, including layer-wise calibration and representational similarity analysis, is in the appendix ~\ref{app:layer_calibration}, ~\ref{app:rsa}.

Within this stage, for each layer (excluding the CLS token), the ten most text-relevant patch tokens are selected and their mean cosine similarity to the text embedding recorded. The layers are split into an early half and a late half, with the top-$k$ scores averaged separately to yield an ``early'' and a ``late'' score. Three scalar features follow: the late score, the gain (late minus early), and the slope (a linear trend across the per-layer top-$k$ scores). A final CLIP similarity (\texttt{final\_cos}, the final image--text cosine) combines with these three into a single composite, the Internal Alignment Score (\textbf{IAS}), a weighted sum ($50\%$ final similarity, $25\%$ late, $15\%$ gain, $10\%$ slope) with empirically set weights kept fixed across all experiments. A learned per-layer readout on the same rows reaches $0.846$, or $0.864$ with the final cosine appended ($+0.021$/$+0.039$ over the fixed IAS, $p<.001$), sacrificing the training-free property; per-domain re-weighting does not help either, degrading pooled AUROC from $0.825$ to $0.782$ (paired $\Delta$ $-0.043$, $p<.001$) with negligible within-domain benefit (Appendix~\ref{app:ias_sensitivity}).  This stage yields five features (\texttt{final\_cos}, \texttt{late}, \texttt{gain}, \texttt{slope}, \texttt{IAS}): related pairs build up localized patch-level evidence in deeper layers, unrelated or blank/noisy inputs never show this consistent late-layer buildup.

In stage four, structured VLM self-assessment: the VLM is prompted to produce both a class label and an answer. The class label is a feature in the ensemble, and the VLM's answer is revealed only when the input is classified as ``related''---otherwise a refusal is returned---so the detector does not depend on the VLM's own tendency to refuse irrelevant questions. This stage also delineates the two operational regimes: standalone TC-LIA (stage three alone) is a genuine pre-generation gate, cheap enough to run before VLM inference and able to bypass generation entirely on non-related pairs, while the full ensemble conditions on a completed generation pass and acts as a joint pre-release verification filter.

In stage five, feature fusion combines pixel-level statistics, domain routing, TC-LIA features, and the VLM's class encoding. XGBoost is the primary classifier \citep{chen2016xgboost}, compared against LightGBM \citep{ke2017lightgbm}, Gradient Boosting, AdaBoost, Random Forest \citep{breiman2001random}, rule-based fusion, and baselines using only CLIP, only TC-LIA, or only the VLM. Algorithm~\ref{alg:tclia_runtime} (Appendix~\ref{app:algorithm}) summarizes inference; Appendix~\ref{app:repro} gives implementation details. The full pipeline takes about $2{,}849.2$~ms against $2{,}675.1$~ms for the VLM alone ($1.06\times$ overhead): pixel statistics, routing, and layer-wise probing account for $173.5$~ms ($6.5\%$), XGBoost a negligible $0.6$~ms (Appendix~\ref{app:latency}, Fig.~\ref{fig:latency}).
\section{Theoretical Motivation}
\label{sec:theory}
\textsc{TC-LIA} rests on one assumption: related pairs show increasing localized semantic alignment in later vision layers, whereas non-related pairs may show generic similarity or spurious attention but lack consistent late-layer evidence.
\vspace{-\baselineskip}

\paragraph{Lemma 1: late-layer alignment separation.}
Let $a_L$ be the late-layer top-$k$ patch alignment. If $a_L|R$ and $a_L|U$ are sub-Gaussian with means $\mu_R$, $\mu_U$, common proxy variance $\sigma^2$, and margin $\Delta=\mu_R-\mu_U>0$, the threshold classifier $\hat{y}=R$ iff $a_L>(\mu_R+\mu_U)/2$ has error at most $\exp(-\Delta^2/(8\sigma^2))$ per class (sub-Gaussian tail bounds at $\tau=(\mu_R+\mu_U)/2$).  On benchmark
data these distributional assumptions are empirically consistent: the excess
kurtosis of every trajectory feature lies between $-0.94$ and $+0.18$ across
the two classes, and Cohen's $d$ for the related/unrelated separation of each
feature is $0.91$--$1.44$ (Appendix~\ref{app:diagnostics}).
\vspace{-\baselineskip}
\paragraph{Lemma 2: gain cancels layer-invariant shortcuts.}
If $a_\ell(x,q)=c(x,q)+r_\ell(x,q)+\epsilon_\ell$ with $c$ a layer-invariant global image--text prior, $r_\ell$ localized late-layer evidence, and $\epsilon_\ell$ noise, then $\mathrm{gain}=\mathrm{late}-\mathrm{early}$ cancels $c$ and estimates the emergence of localized evidence. This is an idealized first-order model rather than exact ViT algebra---intermediate representations drift and renormalize with depth and cosine similarities are nonlinear, so global priors scale with depth---so we read the gain as a bias-reducing contrastive feature that attenuates depth-dependent shortcut priors while averaging noise over the two layer halves, not an exact cancellation.

\paragraph{Proposition 1: staged blank gating decomposes mirage risk.}
Let $g_B$ be the blank/noise gate and $g_N$ be the non-blank related/unrelated detector. The total mirage risk satisfies
\begin{equation}
\mathrm{MR}(g) \leq P(B)\epsilon_B+P(U)\epsilon_U,
\end{equation}
where $\epsilon_B=P(g_B\ne B|B)$ and $\epsilon_U=P(g_N=R|U)$. This decomposition motivates the architecture in Fig.~\ref{fig:pipeline}: a lightweight high-recall blank/noise stage reduces the first term, while \textsc{TC-LIA} and ensemble fusion target the harder semantic mismatch term (proofs in Appendix~\ref{app:proofs}).
\section{Experiments}
\paragraph{Datasets.}
We evaluate on ten visually and semantically diverse VQA domains spanning medical
imaging (Chest VQA~\citep{lau2018dataset}, PathVQA~\citep{he2020pathvqa},
MedXpertQA-MM~\citep{tan2025medxpertqa}, MicroVQA~\citep{burgess2025microvqa},
ReXVQA~\citep{rajpurkar2025rexvqa}), document understanding
(DocVQA~\citep{mathew2021docvqa}, InfoVQA~\citep{mathew2022infographicvqa}),
scene text (TextVQA~\citep{singh2019textvqa}), multi-discipline reasoning
(MMMU-Pro~\citep{yao2024mmmupro}), and video understanding
(Video-MMMU~\citep{fu2025videommmu}).
Each sample is expanded into four input conditions: \textsc{Related} (matching image),
\textsc{In-Domain Unrelated} (hard negative, same domain),
\textsc{Out-of-Domain Unrelated} (easy negative, different domain), and
\textsc{Blank/Noise} (black image of matching resolution).
The ensemble's training set contains 100 samples per domain per condition
(400 records per domain); the remaining samples form the held-out test set, yielding
4,000 training and 15,004 test records.
Full details and provenance are in Appendix~\ref{app:dataset}.

\vspace{-3mm}
\paragraph{Models and Baselines.}
We evaluate 14 VLMs spanning open-source and commercial families:
Qwen2.5-VL-7B, 32B, and 72B~\citep{bai2025qwen25vl},
Llama-3.2-11B-Vision and Llama-4-Scout~\citep{meta2024llama},
LLaVA-OV-Qwen2-7B~\citep{liu2024llava},
InternVL2-8B~\citep{chen2024internvl},
Gemma3-4B~\citep{gemma2025},
Phi-3.5-Vision~\citep{abdin2024phi},
BLIP2-OPT-2.7B~\citep{li2023blip2},
MiniCPM-V-2.6~\citep{yao2024minicpm}, and Kimi-K3
(2.7B--72B parameters across the Qwen-VL, LLaMA, LLaVA, InternVL,
Gemma, Phi, BLIP, and MiniCPM families), plus the commercial
Claude-Opus-4.8~\citep{anthropic2026claude} and
GPT-5.4~\citep{openai2026gpt}.
All models use the same base and structured prompts; the TC-LIA encoder and
ensemble classifier are VLM-agnostic and shared across all fourteen backbones,
and---the ensemble's only VLM-specific input being the self-assessment
feature---a detector trained on one backbone deploys to backbones never seen
during training (Appendix~\ref{app:leave_all_but_one}, five source VLMs).

\vspace{-3mm}

\paragraph{Metrics.}
A trivial detector could minimize mirage rate by refusing every input, so we report three-class accuracy, macro-F1,
mirage rate, AUROC for binary related-versus-nonrelated detection, and answer quality 
of the eventual response using BLEU, ROUGE-L, and BERTScore F1 \citep{papineni2002bleu,lin2004rouge,zhang2020bertscore}. 
Implementation details are provided in Appendix~\ref{app:Implementation}.

\vspace{-4mm}
\section{Results}
\label{sec:results}

\vspace{-3mm}
\paragraph{Layer-wise alignment separates related and non-related inputs.}
Figure~\ref{fig:tc-lia-alignment}(a) visualizes the core empirical signal behind \textsc{TC-LIA}: averaged across domains, \textsc{Related} pairs develop a stronger late-layer top-$k$ patch--text alignment trajectory, whereas \textsc{Unrelated-Real} inputs remain comparatively flat and \textsc{Blank/Noise} inputs show unstable or non-semantic alignment---directly supporting late-layer alignment, early-to-late gain, and slope as 
features.

\begin{table*}[t]
  \centering
  \caption{\footnotesize Main mirage detection results across 14 VLM backbones
  (10 domains, 4 conditions). MR: mirage rate (\%).
  TC-LIA Only is VLM-agnostic (fixed IAS threshold on the shared CLIP encoder),
  so Acc and MR are identical across backbones.
  Clf: best classifier per backbone (RF = Random Forest, Ada = AdaBoost).
  $^\ddagger$InternVL2-8B: the two previously missing domains completed with
  base-prompt inference; row recomputed end-to-end with the released
  training-pipeline configuration over all 10 domains.}
  \label{tab:main}
  \vspace{-3mm}

  {\small
  \renewcommand{\arraystretch}{0.68}
  \setlength{\tabcolsep}{3.5pt}
  \setlength{\aboverulesep}{0.2ex}
  \setlength{\belowrulesep}{0.2ex}

  \begin{tabular*}{\textwidth}{@{\extracolsep{\fill}} l c cc cc cccc}
  \toprule
  \multirow{2}{*}{VLM Backbone}
  & \multirow{2}{*}{Base MR$\downarrow$}
  & \multicolumn{2}{c}{VLM Structured}
  & \multicolumn{2}{c}{TC-LIA Only}
  & \multicolumn{4}{c}{Best Ensemble} \\
  \cmidrule(lr){3-4}\cmidrule(lr){5-6}\cmidrule(lr){7-10}
  &
  & Acc$\uparrow$ & MR$\downarrow$
  & Acc$\uparrow$ & MR$\downarrow$
  & Acc$\uparrow$ & MR$\downarrow$ & Macro-F1$\uparrow$ & Clf \\
  \midrule
  Qwen2.5-VL-72B        & 73.6\% & 88.7\% &  9.9\% & 83.5\% & 7.5\% & \textbf{88.4\%} & 6.4\% & \textbf{0.885} & RF \\
  Claude-Opus-4.8         & 57.3\% & 87.8\% & 11.5\% & 83.5\% & 7.5\% & 88.2\% & 6.2\% & 0.883 & Ada \\
  Qwen2.5-VL-7B         & 73.7\% & 85.1\% &  7.2\% & 83.5\% & 7.5\% & 87.3\% & 6.0\% & 0.871 & RF \\
  Qwen2.5-VL-32B        & 71.4\% & 85.1\% & 13.5\% & 83.5\% & 7.5\% & 86.8\% & 7.2\% & 0.869 & Ada \\
  GPT-5.4               & 72.6\% & 81.9\% & 17.7\% & 83.5\% & 7.5\% & 86.1\% & 6.8\% & 0.860 & RF \\
  Llama-4-Scout         & 71.3\% & 56.8\% & 27.5\% & 83.5\% & 7.5\% & 85.7\% & 6.4\% & 0.854 & RF \\
  Llama-3.2-11B-Vision  & 74.7\% & 35.1\% & 59.6\% & 83.5\% & 7.5\% & 85.0\% & 5.9\% & 0.845 & RF \\
  LLaVA-OV-Qwen2-7B     & 75.0\% & 54.5\% & 32.9\% & 83.5\% & 7.5\% & 85.0\% & 6.1\% & 0.845 & Ada \\
  Gemma3-4B             & 73.9\% & 34.7\% & 60.7\% & 83.5\% & 7.5\% & 84.6\% & 6.1\% & 0.841 & RF \\
  Phi-3.5-Vision        & 74.9\% & 47.7\% &  8.9\% & 83.5\% & 7.5\% & 84.6\% & 5.9\% & 0.840 & RF \\
  BLIP2-OPT-2.7B        & 74.9\% & 25.0\% & 75.0\% & 83.5\% & 7.5\% & 84.5\% & 6.0\% & 0.839 & RF \\
  MiniCPM-V-2.6         & 75.0\% & 54.2\% & 25.2\% & 83.5\% & 7.5\% & 85.6\% & 6.2\% & 0.853 & RF \\
  InternVL2-8B$^{\ddagger}$ & 75.0\% & 59.1\% & 18.9\% & 83.5\% & 7.5\% & 85.9\% & 6.0\% & 0.856 & RF \\
  Kimi-K3               & 68.9\% & 87.3\% & 12.1\% & 83.5\% & 7.5\% & 87.9\% & 6.8\% & 0.880 & RF \\
  \bottomrule
  \end{tabular*}
  }
  \vspace{-3mm}
\end{table*}

\paragraph{Main mirage detection performance.}
Table~\ref{tab:main} reports mirage detection performance across 14 VLM backbones
under three evaluation conditions.
Mirage rate and accuracy capture orthogonal failure modes---refusing everything
gives near-zero mirage rate but poor accuracy on \textsc{Related}
inputs, over-confidence the reverse---so neither metric is
interpretable in isolation.

Base prompt mirage rates range from 57.3\% to 75.0\%, confirming widespread
over-confidence under standard prompting; 75.0\% is the zero-abstention ceiling
(Section~\ref{sec:problem}), and the
weaker backbones sit essentially at it.
Under structured prompting a strong capability dependency emerges: capable
backbones partially self-correct through explicit image classification, whereas
weaker ones stay over-confident, predicting \textsc{Related} for most inputs
(Tables~\ref{tab:mirage_baselines},~\ref{tab:main})---structured prompting alone
is unreliable as a general solution.

\begin{figure}[t]
\centering
\includegraphics[width=\columnwidth]{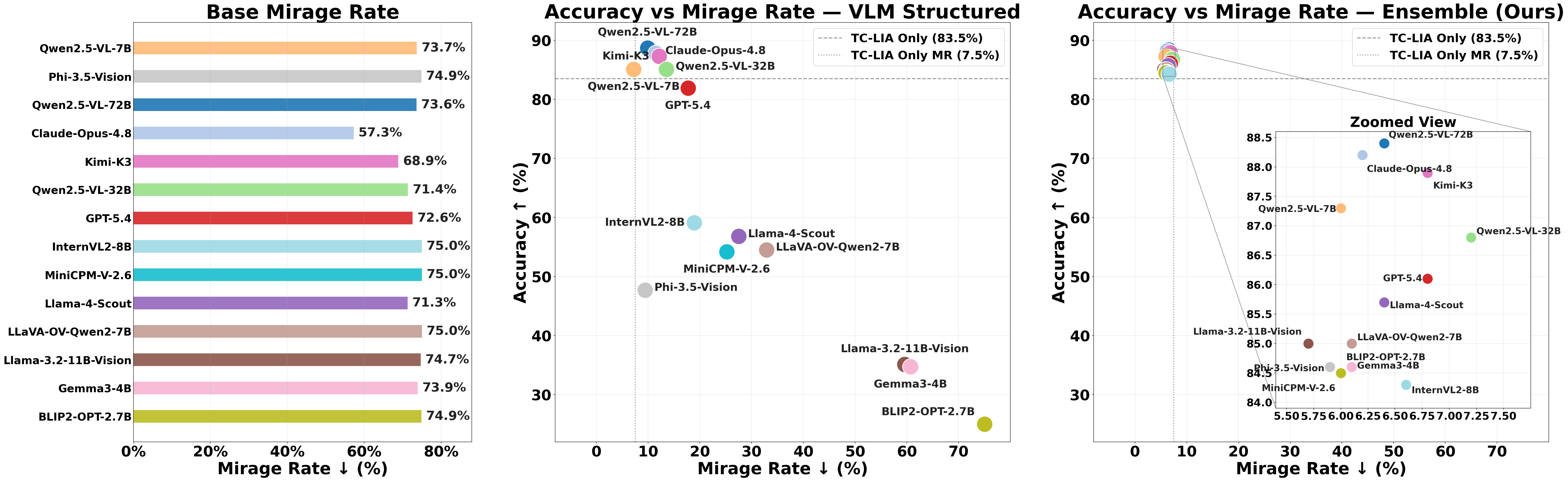}
\caption{Base mirage rate and accuracy--mirage trade-off across 14 VLM backbones.
\textit{(a) Base prompt:} mirage rate only (free-form responses have no defined accuracy;
uncertainty by a strong LLM judge, GPT-4o mini; prompt in Appendix Fig.~\ref{fig:judge_prompt}). 
\textit{(b) Structured prompt:} frontier models
self-correct into the upper-left; weaker models collapse to the lower-right with
high mirage rates or over-refuse (Phi-3.5-Vision: low MR, $\sim$47\% accuracy).
\textit{(c) Ensemble fusion (ours)}: all fourteen backbones converge into a tight
upper-left cluster (84.5--88.4\% accuracy, 5.9--7.2\% MR); zoomed inset resolves
individual models. Dashed/dotted lines: TC-LIA-only baselines (83.5\% accuracy,
7.5\% MR); fusion makes detection robust to backbone capability.}
\vspace{-5mm}
\label{fig:acc_vs_mirage}
\end{figure}

TC-LIA alone reduces mirage rate to 7.5\% at 83.5\% accuracy across all backbones,
a strong VLM-agnostic baseline.
Ensemble fusion improves both dimensions simultaneously (mirage rate
5.9--7.2\%, accuracy 84.5--88.4\%). On the strongest frontier backbones it is a
risk-reducing calibration filter: for Qwen2.5-VL-72B it trades a fraction of the
structured-prompt accuracy (88.7\% to 88.4\%) for a substantially lower mirage
rate (9.9\% to 6.4\%)---the best joint result.
The fusion gain is not reproducible by a single learned head:
logistic-regression selectors trained on the same $2{,}972$ non-blank
training rows with ground-truth labels reach at most $0.848$ held-out AUROC
versus $0.918$ for the ensemble on identical rows (Appendix~\ref{app:diagnostics}).
Figure~\ref{fig:acc_vs_mirage}(a)--(c) visualizes this progression across all
fourteen backbones.

\begin{table*}[t]
\centering
\setlength{\tabcolsep}{3pt}
\renewcommand{\arraystretch}{0.95}
\caption{Mirage detection under a shared protocol (canonical split; AUROC for related vs.\ unrelated). Encoder rows report our signal: pipeline = blank gate + routing + VLM self-assessment + CLIP ViT-H/14 trajectory readout; fusion is the full system. Basis counts: 19{,}004 full (including blank), 14{,}133 nonblank, 11{,}161 held-out test rows, and 7{,}500 S1 generation rows. Paired domain-stratified bootstrap (1{,}000 resamples) compares with raw IAS on each row's basis, except gated SigLIP, compared with gated TC-LIA. Gated rows are ranked at the bottom. Generation baselines use Qwen2.5-VL-7B-Instruct (NF4), 448\,px images, and the paper's prompts. $^{\dagger}$ IAS plus the blank hard gate. $^{\ddagger}$ The QT probe (HALP-style) works by extracting the last-prompt-token hidden state from specific intermediate transformer layers which is often inaccessible for the closed source models with api token.}
\label{tab:mirage_baselines}
\footnotesize
\begin{tabular*}{\textwidth}{@{\extracolsep{\fill}}llccc@{}}
\hline
Method & Access & Basis & AUROC [95\% CI] & id/od-HN \\
\hline
\textbf{Proposed (CLIP enc.): IAS + gate}$^{\dagger}$ & CLIP fwd & 19{,}004 & \textbf{0.880} [0.875, 0.885] & 0.740 / 0.905 \\
\textbf{Proposed (CLIP enc.): fusion (A+B+C+D)} & CLIP+VLM fwd & 11{,}161 & \textbf{0.918} [0.913, 0.923] & 0.876 / 0.961  \\
\textbf{Proposed (SigLIP enc.): cosine + gate} & SigLIP fwd & 19{,}004 & \textbf{0.908} [0.903, 0.912] & 0.790 / 0.938 \\
SigLIP-SO400M cosine (raw, no gate) & SigLIP fwd & 14{,}133 & 0.870 [0.864, 0.876] & 0.795 / 0.945  \\
VLM self-class & 1 VLM fwd & 19{,}004 & 0.824 [0.818, 0.831] & 0.760 / 0.840 \\
 CLIP-UP \citep{vardi2026clipup} & CLIP fwd &  11,161 &  0.782 [0.774, 0.791] &  0.693 / 0.872 \\
QT probe, layer 7, HALP-style \citep{kogilathota2026halp}$^{\ddagger}$ & VLM internals & 11{,}161 & 0.813 [0.806, 0.821] & 0.753 / 0.873 \\
Dual-prompt agr. & 2 VLM fwds & 19{,}004 & 0.635 [0.629, 0.641] & 0.604 / 0.646 \\
Semantic entropy (k=10) & 11 gens & 7{,}500 & 0.606 [0.592, 0.618] & 0.601 / 0.611  \\
Lexical entropy & 11 gens & 7{,}500 & 0.635 [0.623, 0.647] & 0.625 / 0.645 \\
Rephrasing consistency (SAC3) & 4 gens & 7{,}500 & 0.550 [0.538, 0.561] & 0.545 / 0.555 \\
\hline
\end{tabular*}
\end{table*}

\noindent\textbf{Comparison.} We compare TC-LIA against mirage-detection baselines spanning three access
regimes under a shared protocol: primary open backbone (Qwen2.5-VL-7B-Instruct,
quantized to 4-bit NormalFloat, NF4), byte-identical prompts, and the canonical split with domain-stratified
bootstrap 95\% CIs (Table~\ref{tab:mirage_baselines}). Self-report baselines
needing VLM forward passes approach but do not match TC-LIA: structured
self-classification~\citep{kadavath2022language} reaches 0.824, dual-prompt
agreement~\citep{manakul2023selfcheckgpt} 0.635. The strongest single baseline is
VLM-internal---a HALP-style query-token hidden-state
probe~\citep{kogilathota2026halp} fitted on Qwen2.5-VL-7B reaches 0.903,
$+0.078$ above raw TC-LIA IAS on identical rows---but it reads the target VLM's
internal activations, requiring white-box access and a separate probe fit per
backbone, unavailable for closed models (GPT-5.4, Claude-Opus-4.8);
at an untuned intermediate layer (layer~7) the same probe drops to 0.813
($-0.012$ vs.\ IAS on identical rows), so layer choice is itself a tuned
hyperparameter, whereas TC-LIA's single frozen CLIP forward is shared across all
fourteen backbones (Appendix~\ref{app:encoder_gen}).
Generation-time uncertainty is weaker at far higher cost: discrete semantic
entropy~\citep{kuhn2023semanticuncertainty,farquhar2024semanticentropy} over 11
sampled answers reaches 0.606 (lexical variant 0.635; rephrasing
consistency~\citep{zhang2023sac3} over 4 paraphrases 0.550)---mirage is an
input-side failure, so signals measured on generated text cannot recover evidence
the input never contained. Encoder swaps probe the mechanism, as ablations of our
own signal: the image--text cosine transfers across backbones
(SigLIP-SO400M~\citep{zhai2023siglip} 0.870 on the nonblank subset), while a
DINOv2~\citep{oquab2024dinov2} random-adapter control collapses to near chance
(0.481), confirming text-grounded alignment rather than a
generic image statistic. As single gated signals, SigLIP-SO400M's final cosine
reaches 0.908 on the full 19{,}004 records versus 0.880 for the
CLIP-instantiated IAS under the identical Stage-1 gate, the advantage
concentrating on out-of-domain negatives (0.938 vs.\ 0.905)
(Table~\ref{tab:mirage_baselines}); we therefore do not claim state-of-the-art
standalone AUROC for TC-LIA. Swapping the SigLIP cosine into the full pipeline
in place of the five trajectory features (gate, routing, self-assessment, pixel
statistics, and fusion protocol unchanged; A+B+C+E vs.\ A+B+C+D) performs on
par: held-out accuracy 87.35\% vs.\ 87.28\%, macro-F1 0.874 vs.\ 0.871,
nonblank AUROC 0.922 vs.\ 0.918 (paired $\Delta$ $+0.003$, $p=.064$),
deployed test AUROC 0.941 vs.\ 0.939---but the mirage rate rises from 6.0\% to
6.8\% under the released ensemble protocol (6.2\% vs.\ 6.0\% under a matched
RandomForest), so the released system keeps the CLIP trajectory readout.
Adding the cosine as a twelfth input instead yields a ranking and accuracy gain
without mirage suppression: held-out AUROC $0.918\to0.929$ and deployed AUROC
$0.939\to0.946$ (both $p<.001$), accuracy and macro-F1 $+0.5$ points
($p\le.006$), and a statistically unchanged mirage rate
($5.99\%\to5.95\%$, $p=.74$; Appendix~\ref{app:diagnostics}).  As single signals the trajectory features do not surpass final CLIP cosine
(Table~\ref{tab:tclia_feature_auroc}); same-domain ReXVQA negatives remain
near chance (IAS AUROC $0.497$, Appendix~\ref{app:hard_negatives}), motivating
complementary fusion with pixel, routing, and VLM self-assessment.

\vspace{-3mm}
\paragraph{Accuracy--mirage trade-off across backbones.}
Figure~\ref{fig:acc_vs_mirage}(a)--(c) shows the base-prompt mirage rate (a) and the
accuracy--mirage trade-off under structured prompting (b) and ensemble fusion (c).
Under the base prompt, mirage rates span 57.3--75.0\% (free-form responses,
judged only for uncertainty; no accuracy defined).
Under structured prompting, models split into two well-separated clusters:
frontier models (Qwen2.5-VL-72B, Claude-Opus-4.8, Qwen2.5-VL-32B, GPT-5.4) land
in the upper-left, meeting or exceeding the TC-LIA accuracy baseline through
self-assessment alone, while weaker models (BLIP2, Gemma3-4B, Llama-3.2-11B-Vision,
LLaVA-OV) cluster in the lower-right (mirage rates $\sim$33--75\%, accuracy
25--55\%)---over-confidence rather than over-refusal.
Phi-3.5-Vision is an instructive outlier: a low mirage rate at only
$\sim$47\% accuracy---predicting \textsc{Unrelated} for nearly all inputs games
the metric without genuine grounding---so structured prompting helps mainly
models that already have strong grounding.

Under ensemble fusion, this dispersion collapses entirely: all fourteen
backbones converge into a tight upper-left cluster (84.5--88.4\% accuracy,
5.9--7.2\% mirage rate), so tight that the zoomed inset is needed to tell models
apart. The ensemble lifts the weakest models by over 50 percentage points in
accuracy without degrading the already-strong frontier models; backbone choice no
longer determines outcome. As a selective predictor, the ensemble's confidence ordering also supports
graded release policies trading residual mirage risk against false abstention:
at 80\% coverage, mirage falls from 24--28\% to 13--17\%, while related-answer
recall is 50--63\% (Appendix~\ref{app:risk_coverage}).

\textbf{Per-backbone training is an upper bound, not a deployment requirement.} The ensemble's
only VLM-specific input is the self-assessment feature, so a detector trained
\emph{once} on a single backbone generalizes to unseen ones. Frozen after training on
InternVL2-8B (a mid-pack in-domain performer with a conservative self-assessment
style), the ensemble reaches a macro 85.3\% accuracy, 6.3\% mirage rate, and
0.894 AUROC across the 13 held-out backbones, versus 86.0\%/6.3\%/0.903 for
per-backbone training and 83.5\%/7.5\% for the unsupervised TC-LIA-only rule:
0.7 points of cost, over half of the ensemble's accuracy gain kept. Transfer
success depends on self-assessment style, not in-domain strength: conservative
self-assessors transfer best, while the strongest in-domain models transfer worst
(Appendix~\ref{app:leave_all_but_one}). Removing self-assessment from source
training yields a backbone-independent detector with 84.6\% accuracy and 6.1\%
mirage rate across 13 held-out backbones (Appendix~\ref{app:diagnostics}).

\begin{table}[t]
\centering
\small
\setlength{\tabcolsep}{3pt}
\renewcommand{\arraystretch}{0.9}
\caption{Build-up ablation over TC-LIA features (Random Forest, ten-domain benchmark;
\texttt{vlm\_class\_enc} in V10 queried from GPT-5.4). The 11-feature ensemble is
best on all metrics; V09 $=$ domain-adaptive IAS thresholds, no learned classifier
(AUROC undefined). $^\ddagger$ rows: released training-pipeline configuration
(same features, split, hyperparameters), harmonised snapshot: V10$^\ddagger$/V11$^\ddagger$
mutually comparable, absolute levels differ from other rows. V11 replaces the four
trajectory features with \texttt{final\_cos}, keeping domain routing, pixel
statistics, and self-assessment.}
\begin{tabular}{llcccc}
\hline
Variant & Features & Acc (\%)$\uparrow$ & F1$\uparrow$ & MR (\%)$\downarrow$ & AUROC$\uparrow$ \\
\hline
V01 & Late top-$k$ mean only          & 78.0 & 0.768 & 8.6 & 0.750 \\
V02 & Gain only                        & 76.8 & 0.745 & 7.6 & 0.725 \\
V03 & Slope only                       & 77.3 & 0.754 & 8.0 & 0.734 \\
V04 & Final cosine only                & 82.1 & 0.816 & 7.7 & 0.824 \\
V05 & Late $+$ Gain                    & 78.9 & 0.774 & 7.6 & 0.775 \\
V06 & Late $+$ Slope                   & 79.0 & 0.774 & 7.4 & 0.779 \\
V07 & Gain $+$ Slope                   & 79.3 & 0.770 & 6.0 & 0.755 \\
V08 & Late $+$ Gain $+$ Slope         & 80.5 & 0.793 & 7.2 & 0.798 \\
V09 & IAS threshold (TC-LIA only)    & 83.5 & 0.832 & 7.5 & --- \\
V10$^\ddagger$ & Full ensemble (recomputed) & 82.1 & 0.799 & 13.6 & 0.917 \\
V11$^\ddagger$ & Control: no trajectory features & 81.9 & 0.797 & 14.3 & 0.915 \\
\textbf{V10} & \textbf{Full ensemble} & \textbf{86.2} & \textbf{0.861} & \textbf{6.8} & \textbf{0.917} \\
\hline
\end{tabular}

\label{tab:tclia_feature_auroc}
\end{table}

\vspace{-3mm}

\paragraph{Ablation Study: TC-LIA Feature Contributions}

Table~\ref{tab:tclia_feature_auroc} presents a build-up ablation over TC-LIA alignment features for binary \textsc{Related} vs.\ \textsc{Unrelated-Real} discrimination across the full ten-domain benchmark. Individual trajectory features are weak in isolation (late top-$k$ mean $0.750$, gain $0.725$, slope $0.734$); final CLIP cosine similarity alone reaches $0.824$, stronger but still insufficient. Pairwise ($0.755$--$0.779$) and three-way ($0.798$) combinations add consistent incremental gains, where each feature captures a distinct, non-redundant aspect of evidence accumulation and the scalar IAS composite reaches $0.814$ AUROC. The TC-LIA-only threshold classifier (domain-adaptive IAS thresholds, no VLM call) achieves $83.5\%$ accuracy and $7.5\%$ MR, while the full 11-feature ensemble, adding VLM self-assessment, domain routing, and pixel statistics, reaches $86.2\%$ accuracy, $0.917$ AUROC, and $6.8\%$ MR: each component class contributes independently, and the complete feature set is needed for peak performance. In table~\ref{tab:tclia_feature_auroc}, the V11 control isolates the trajectory features' contribution: under the matched
recomputation (V10$^\ddagger$ vs.\ V11$^\ddagger$), replacing the four trajectory features
with \texttt{final\_cos} alone costs $0.2$ accuracy points and $0.002$ AUROC on GPT-5.4,
with consistent but modest gains across five recomputed backbones (up to $0.010$ AUROC on
Gemma3-4B). The trajectory features therefore add reliable but small signal in the
supervised ensemble; their larger contribution is the training-free at deployment regime, where the
IAS threshold rule (V09) reaches 83.5\% accuracy with no learned classifier or VLM
interaction.

\paragraph{IAS provides class-separating evidence.}
Figure~\ref{fig:tc-lia-alignment}(b): \textsc{Related} pairs shift toward higher
IAS---stronger question-conditioned evidence across layers.
\textsc{Out-of-Domain} negatives concentrate at consistently low IAS;
\textsc{In-Domain Unrelated} examples occupy an intermediate range---above
out-of-domain negatives via shared domain-level priors, below \textsc{Related}
pairs---and \textsc{Blank/Noise} inputs form a narrow low-IAS distribution
(absent semantic patch activations). This four-way separation makes IAS
pre-generation discriminative (domain detail in
Appendix~\ref{app:Domain-wise IAS}).

\vspace{-\baselineskip}
\paragraph{Qualitative results.}
Figure~\ref{fig:result_cards_main} contrasts the same medical question on matched and mismatched inputs: the ensemble keeps the answer on the related CT image but refuses the raw VLM's plausible medical-style answer on an unrelated image (further per-backbone cards in Figs.~\ref{fig:result_gemma}--\ref{fig:result_cards_internvl3_38b}).

\begin{figure*}[t]
\centering
\scriptsize
\setlength{\fboxsep}{3pt}
\setlength{\fboxrule}{0.4pt}

\fbox{%
\begin{minipage}[t][3.9cm][t]{0.48\linewidth}
\vspace{0pt}
\textbf{Example A: Medical Question / Related Image}\\
\hfill
\textbf{Model: \texttt{gpt-5.4}}\\[-0.1em]

\begin{minipage}{0.18\linewidth}
\centering
\fbox{\includegraphics[width=0.92\linewidth]{figures/result_card/images_chest_vqa_real_20.jpg}}
\end{minipage}
\hfill
\begin{minipage}{0.77\linewidth}
\textbf{Question:} ``what is the medium density close to the anterior abdominal wall?''

\textbf{Raw VLM output:} ``There is a crescent of low-density free intraperitoneal air immediately deep to the anterior abdominal wall.''

\textbf{Ensemble decision:} \textsc{Related}.

\textbf{Released output:} ``There is a crescent of low-density free intraperitoneal air immediately deep to the anterior abdominal wall.''
\end{minipage}
\end{minipage}}
\hfill
\fbox{%
\begin{minipage}[t][3.9cm][t]{0.48\linewidth}
\vspace{0pt}
\textbf{Example B: Medical Question / Unrelated-Real Image}\\
\hfill
\textbf{Model: \texttt{gpt-5.4}}\\[-0.1em]

\begin{minipage}{0.18\linewidth}
\centering
\fbox{\includegraphics[height=1.12\linewidth,keepaspectratio]{figures/result_card/images_chest_vqa_unrelated_20.jpg}}
\end{minipage}
\hfill
\begin{minipage}{0.77\linewidth}
\textbf{Question:} ``what is the medium density close to the anterior abdominal wall?''

\textbf{Raw VLM output:} ``The medium density close to the anterior abdominal wall is the liver.''

\textbf{Ensemble decision:} \textsc{Unrelated-Real}.

\textbf{Released output:} ``I cannot answer based on the given image.''\\
\end{minipage}
\end{minipage}}

\vspace{2mm}

\caption{Result cards for the same medical question and VLM (GPT-5.4): the ensemble preserves the answer on the related CT image but rejects the unrelated real image, where the VLM hallucinates a visually unsupported response (``\,the liver''), preventing a mirage.}

\label{fig:result_cards_main}
\vspace{-2mm}
\end{figure*}

\begin{table}[t]
\centering
\caption{Answer quality averaged over all three conditions (\textsc{Related}: vs.\
ground-truth answers; \textsc{Unrelated}/\textsc{Blank}: vs.\ six reference refusal
phrases). \textbf{Bold} ensemble scores improve on the base prompt. Unrelated/blank
inputs dominate the split and receive fixed refusal strings, so aggregate gains
reflect refusal compliance; on accepted \textsc{Related} queries lexical scores
can drop (Qwen2.5-VL-72B BLEU 0.180$\to$0.007) while BERTScore stays stable.
Per-condition breakdown in Appendix~\ref{app:quality_detail}. $^\ddagger$InternVL2-8B:
recomputed over all 10 domains with the released training-pipeline configuration;
BERTScore cells are raw roberta-large F1.}
\vspace{-2mm}
\label{fig:quality}
\small
\renewcommand{\arraystretch}{0.72}
\setlength{\tabcolsep}{2.5pt}
\begin{tabular*}{\linewidth}{@{\extracolsep{\fill}} l cc cc cc}
\toprule
& \multicolumn{2}{c}{BLEU} & \multicolumn{2}{c}{ROUGE-L} & \multicolumn{2}{c}{BERTScore} \\
\cmidrule(lr){2-3}\cmidrule(lr){4-5}\cmidrule(lr){6-7}
Model & Base & Ens. & Base & Ens. & Base & Ens. \\
\midrule
Qwen2.5-VL-72B     & 0.074 & \textbf{0.628} & 0.204 & \textbf{0.701} & 0.855 & \textbf{0.945} \\
Claude-Opus-4      & 0.025 & \textbf{0.630} & 0.252 & \textbf{0.681} & 0.851 & \textbf{0.937} \\
Qwen2.5-VL-7B      & 0.040 & \textbf{0.630} & 0.143 & \textbf{0.698} & 0.855 & \textbf{0.948} \\
Qwen2.5-VL-32B     & 0.031 & \textbf{0.620} & 0.146 & \textbf{0.659} & 0.838 & \textbf{0.931} \\
GPT-4.1            & 0.038 & \textbf{0.629} & 0.206 & \textbf{0.725} & 0.859 & \textbf{0.950} \\
Llama-4-Scout      & 0.025 & \textbf{0.630} & 0.132 & \textbf{0.729} & 0.873 & \textbf{0.953} \\
Llama-3.2-11B      & 0.014 & \textbf{0.633} & 0.117 & \textbf{0.753} & 0.877 & \textbf{0.960} \\
LLaVA-OV-Qwen2-7B  & 0.065 & \textbf{0.634} & 0.134 & \textbf{0.721} & 0.859 & \textbf{0.956} \\
Gemma3-4B          & 0.042 & \textbf{0.630} & 0.141 & \textbf{0.695} & 0.858 & \textbf{0.947} \\
Phi-3.5-Vision     & 0.065 & \textbf{0.648} & 0.141 & \textbf{0.743} & 0.868 & \textbf{0.965} \\
BLIP2-OPT-2.7B     & 0.012 & \textbf{0.628} & 0.101 & \textbf{0.684} & 0.863 & \textbf{0.956} \\
MiniCPM-V-2.6      & 0.012 & \textbf{0.633} & 0.101 & \textbf{0.724} & 0.861 & \textbf{0.961} \\
InternVL2-8B       & 0.028 & \textbf{0.632} & 0.105 & \textbf{0.726} & 0.862 & \textbf{0.959} \\
Kimi-K3            & 0.033 & \textbf{0.625} & 0.246 & \textbf{0.706} & 0.884 & \textbf{0.936} \\
\bottomrule
\end{tabular*}
\end{table}

\vspace{-\baselineskip}
\paragraph{Complementary roles of TC-LIA and VLM self-assessment.}
The two signals are non-redundant (Appendix
Figure~\ref{fig:complementary}): TC-LIA carries \textsc{Out-of-Domain} negatives
($0.913$ overall AUROC, $>0.93$ on seven of ten domains) while VLM
self-assessment is weaker there, but for \textsc{In-Domain} hard negatives
neither component alone suffices ($0.744$/$0.738$) and the ensemble lifts overall
AUROC to $0.884$, $+0.14$ over either. Feature importances mirror the split:
\texttt{vlm\_class\_enc} dominates in-domain (gain $0.585$), TC-LIA features the
out-of-domain weight. Per-condition AUROCs, $752$ near-miss rows (final cosine
$0.270$ vs.\ $0.149$), and importances: Appendix~\ref{app:diagnostics}.

\vspace{-3mm}
\paragraph{Error structure, answer quality, and negative probes.}
Blank/noise inputs are detected with near-perfect recall, so the dominant errors
are related--unrelated confusions---semantic mismatch, not image corruption.
Ensemble fusion lifts BLEU/ROUGE-L by ${\sim}+0.6$/${\sim}+0.55$ over the base
prompt while preserving BERTScore on accepted \textsc{Related} inputs
(Table~\ref{fig:quality}); rejected standalone probes (CLIP relevance,
GradCAM/attention, SAM3 grounding, RAPT ratios, easy small-scale
datasets) are detailed in Appendix~\ref{app:negative}.

\vspace{-3mm}

\section{Limitations}
\vspace{-3mm}
TC-LIA can flag relevance, however cannot provide answer correctness. In addition, it needs intermediate ViT tokens, and its fixed final-CLIP projection is diagnostic. The fixed weights may not generalize equally well across domains. Cross-backbone transfer can also vary with the VLM's self-assessment behavior (Appendix~\ref{app:leave_all_but_one}) and removing that feature during source training yields 84.6\% accuracy and 6.1\% mirage rate across 13 held-out backbones (Appendix~\ref{app:diagnostics}). Leave-one-domain-out testing also exposes a particular medical domain gap: ReXVQA reaches 73.4\% accuracy and 24.4\% mirage rate (Appendix~\ref{app:ablations}).

\vspace{-3mm}
\section{Conclusion}
\vspace{-3mm}
We introduced \textsc{TC-LIA}, a text-conditioned layer-wise alignment method that detects mirage-prone image--question pairs before generation. Using fixed projections and weights, the training-free at deployment version of the proposed method reduces mirages to 7.5\% at 83.5\% classification accuracy across 14 VLM backbones. Combined with blank/noise detection, domain-adaptive routing, and optional VLM self-assessment, the ensemble reaches 84.5--88.4\% accuracy with 5.9--7.2\% mirage rates while preserving answer quality on related inputs. An ensemble trained on a single backbone transfers to 13 unseen backbones, with the best source within 0.7 percentage points of per-backbone training. Together, these results support a simple safeguard for multimodal systems: verify that question-relevant visual evidence is present before releasing an answer. As VLMs are increasingly deployed in high-stakes and open-world settings, detecting when an answer is unsupported by the visual input is becoming an important AI safety requirement. Our results suggest that evidence verification can provide a lightweight safety layer that reduces unsupported multimodal reasoning before it propagates to downstream users or systems.

\clearpage
\section*{AI Use Statement}

Generative AI tools were used to critique theoretical framing, assist with code verification and quantitative result checking, identify relevant literature, and assist with manuscript composition and editing. They were not used to fabricate or alter experimental observations. Numerical claims were checked against the underlying numerical records and experimental figures; mathematical claims were re-derived and reviewed by the authors, who take responsibility for the final manuscript and released materials.


\section*{Ethics Statement}

This work targets safer deployment of VLMs by reducing unsupported answers in medical and document VQA. The detector should not be used as a standalone clinical decision system. It is a pre-release safety layer intended to trigger abstention or human review when visual evidence is missing or mismatched. All medical data should be de-identified and used according to applicable licenses and institutional requirements.




\clearpage
\appendix

\begin{center}
{\Large \textbf{Appendix}}
\end{center}
\vspace{1em}

\section{Reproducibility Details}
\label{app:repro}
\subsection{Dataset Composition}
\label{app:dataset}

Table~\ref{tab:dataset_appendix} provides the detailed dataset composition used in our mirage detection experiments. We evaluate across ten visually and semantically diverse VQA domains spanning medical imaging (Chest VQA, PathVQA, MedXpertQA-MM, MicroVQA, ReXVQA), document understanding (DocVQA, InfoVQA), scene text (TextVQA), multi-discipline reasoning (MMMU-Pro), and video understanding (Video-MMMU). Domains vary in answer format; five use free-form answers and five use multiple-choice (MCQ A--D or A--E). They also vary in image modality, including single images, multi-image lists, base64-encoded multi-image, and video frames.

For each domain, examples are organized into four input conditions:
\begin{itemize}
    \item \textsc{Related}: the image is semantically matched to the question (ground-truth positive).
    \item \textsc{In-Domain Unrelated}: a different image drawn from the \emph{same} domain, forming a hard negative: the image is visually coherent with the domain but does not contain evidence for the question.
    \item \textsc{Out-of-Domain Unrelated}: an image drawn uniformly at random from a \emph{different} domain, forming an easy negative: visually plausible but semantically irrelevant to the question.
    \item \textsc{Blank/Noise}: a black image of matching resolution, providing a non-informative visual input.
\end{itemize}

For all three unrelated conditions, the ground-truth answer is the standard refusal \emph{``I cannot answer based on the given image.''}  Hard negatives (in-domain) are constructed with an MD5-hash check (single-image domains) or sample-index check (multi-image/video domains) to prevent accidental duplicate positives within the same domain. Residual semantic overlap between an in-domain negative and the paired question cannot be excluded by construction; this ambiguity is deliberate: it is what makes the \textsc{id}-unrelated condition the hard-negative subset, and it manifests as near-chance single-signal separability on the medical domains (Table~\ref{tab:mirage_baselines}).

This four-way construction allows us to evaluate whether the detector (i) preserves answerability for genuinely matched image--question pairs, (ii) rejects low-level uninformative inputs, (iii) rejects easy cross-domain mismatches, and (iv) rejects the more challenging within-domain semantic mismatches.

\paragraph{Dataset split.}
Each domain contributes a training split of 100 samples per condition (400 records per domain), drawn uniformly at random (seed 42); all the remaining samples form the held-out evaluation set, yielding 4,000 training records and 15,004 test records across the full benchmark.

\begin{table}[h]
\centering
\small
\begin{tabular}{llcrrr}
\hline
Domain & Format & Answer type & Train & Test & Total \\
\hline
Chest VQA      & single image    & free-form & 400  & 1,404 & 1,804  \\
PathVQA        & single image    & free-form & 400  & 1,600 & 2,000  \\
TextVQA        & single image    & free-form & 400  & 1,600 & 2,000  \\
DocVQA         & single image    & free-form & 400  & 1,600 & 2,000  \\
InfoVQA        & single image    & free-form & 400  & 1,600 & 2,000  \\
MedXpertQA-MM  & multi-list      & MCQ A--E  & 400  & 1,600 & 2,000  \\
MicroVQA       & base64-multi    & MCQ A--E  & 400  & 1,600 & 2,000  \\
ReXVQA         & single image    & MCQ A--D  & 400  & 1,600 & 2,000  \\
MMMU-Pro       & multi-numbered  & MCQ A--D  & 400  & 1,600 & 2,000  \\
Video-MMMU     & video frames    & MCQ A--D  & 400  & 800   & 1,200  \\
\hline
Total          &                 &           & 4,000 & 15,004 & 19,004 \\
\hline
\end{tabular}
\caption{Dataset composition across domains and splits. Record counts equal samples $\times$ 4 conditions (\textsc{Related}, \textsc{In-Domain Unrelated}, \textsc{Out-of-Domain Unrelated}, \textsc{Blank/Noise}). Chest VQA uses all 451 of its available samples (351 test + 100 train); Video-MMMU uses all 300 of its available samples.}
\label{tab:dataset_appendix}
\end{table}

\subsection{Implementation Summary}
\label{app:Implementation}

The full system uses Python, PyTorch \citep{paszke2019pytorch}, OpenCLIP
\citep{ilharco2021openclip}, scikit-learn \citep{pedregosa2011sklearn},
XGBoost \citep{chen2016xgboost}, and LightGBM \citep{ke2017lightgbm}.

\paragraph{TC-LIA feature extraction.}
TC-LIA registers forward hooks on all 32 transformer blocks of a frozen
ViT-H/14 CLIP encoder.  For each non-blank image--question pair, the
image is passed through the vision encoder; intermediate patch tokens
are projected into the shared CLIP embedding space via the model's
\texttt{ln\_post} and projection matrix, and cosine similarity is
computed against the text embedding.  At each layer, the mean cosine
similarity of the top-$k$ ($k{=}10$) patch tokens is recorded, yielding
a 32-point \emph{top-$k$ curve}.  The scalar feature
\texttt{internal\_alignment\_score} is then computed as a fixed linear
combination:
\begin{equation}
\mathrm{IAS} = 0.50\, f_{\cos}
+ 0.25\, \bar{s}_{\mathrm{late}}
+ 0.15\, \Delta_{\mathrm{gain}}
+ 0.10\, \beta_{\mathrm{slope}}
\end{equation}
where $f_{\cos}$ is the final-layer cosine similarity, $\bar{s}_{\text{late}}$
is the mean of the top-$k$ curve over the second half of layers,
$\Delta_{\text{gain}}$ is the difference between late-layer and
early-layer means, and $\beta_{\text{slope}}$ is the slope of a
linear fit over the full 32-point curve.

Domain classification uses a zero-shot CLIP prompt bank (16 medical
templates and 13 natural-scene templates) with a softmax temperature of
0.07.  A domain is labelled \texttt{medical} if the aggregated medical
probability exceeds 0.65, \texttt{natural} if the natural probability
exceeds 0.65, and \texttt{mixed} otherwise.  Blank/noise images bypass
the cosine pipeline entirely; their features are derived from grayscale
pixel statistics (\texttt{image\_is\_blank}, \texttt{image\_std}).


\paragraph{Ensemble classifiers and hyperparameters.}
Five classifiers are trained independently on the 11 ensemble features
per VLM backbone:

\begin{itemize}
\item \textbf{XGBoost}: 300 estimators, max depth 6, learning rate 0.05, subsample 0.8, column subsample 0.8, log-loss evaluation metric.
\item \textbf{LightGBM}: 300 estimators, max depth 6, learning rate 0.05, subsample 0.8, column subsample 0.8.
\item \textbf{Gradient Boosting}: 200 estimators, max depth 5, learning rate 0.05, subsample 0.8.
\item \textbf{AdaBoost}: 200 estimators, learning rate 0.5, base learner is a decision stump with max depth 3.
\item \textbf{Random Forest}: 300 estimators, max depth 8.
\end{itemize}

Blank images are classified deterministically as \texttt{BLANK\_OR\_NOISE}
without invoking any classifier; the ensemble operates only on
non-blank samples.  The gate itself thresholds three grayscale pixel
statistics (standard deviation \texttt{s1\_std}, patch-variance
coefficient of variation \texttt{s1\_cv}, and a sharpness ratio
\texttt{s1\_sf}); on the benchmark the deployed decision is equivalent to
flagging an image when its grayscale standard deviation is at most $10.78$
or its coefficient of variation is at most $0.07$ ($99.85\%$ agreement with
the stored per-record labels).  Across all $19{,}004$ records the gate
flags $4{,}871$ images---$4{,}751$ true blank/noise inputs and $120$ real
images ($35$ \textsc{Related} and $57$ \textsc{Unrelated-Real} on the test
split)---yielding precision $97.5\%$ at recall $100\%$: every blank/noise
input is refused, and the false flags concentrate in genuinely low-contrast
content (median grayscale std $10.2$ for flagged real images vs.\ $51.1$
for passing ones).  Per-domain precision/recall and failure cases are
reported in Appendix~\ref{app:diagnostics}.  Model selection is performed via 5-fold
stratified cross-validation (StratifiedKFold, shuffle=True, seed 42)
on the training split; the classifier with the highest mean CV accuracy
is selected as the best ensemble for each VLM backbone.












\subsection{Prompt Conditions and Mirage Rate Measurement}
\label{sec:prompt_conditions}

We evaluate VLMs under two prompt conditions that differ in how mirage behavior is detected.

\subsubsection*{Base Prompt Condition}

In the base prompt condition, the model receives a minimal task-focused instruction (Fig.~\ref{fig:base_prompt}) with no explicit guidance to assess image validity. This mirrors realistic deployment scenarios where no special prompting is applied.

\begin{figure}[h]
\centering
\fbox{
\begin{minipage}{0.95\columnwidth}
\small
Directly answer the question based on the image, no explanation is needed.\\
If options are provided, respond with the option letter only.

\vspace{0.5em}
\textbf{Question:} \{question\} \{options\}
\end{minipage}
}
\caption{Base prompt used for standard VLM inference.}
\label{fig:base_prompt}
\end{figure}

Since the base prompt does not instruct the model to flag invalid images, responses cannot be evaluated by simple string matching. Instead, we apply an LLM judge (GPT-4o mini) to the model's free-form responses on the \textsc{Unrelated} and \textsc{Blank/Noise} conditions only: the \textsc{Related} condition is excluded from mirage measurement, as correct confident answers are expected. The judge prompt is shown in Fig.~\ref{fig:judge_prompt}.

\begin{figure}[h]
\centering
\fbox{
\begin{minipage}{0.95\columnwidth}
\small
Analyze the following response and determine if it mentions not seeing an image or if it shows any uncertainty or lack of confidence in the answer.

\vspace{0.5em}
\textbf{Response:} \{vlm\_response\}

\vspace{0.5em}
Does this response mention that it cannot see an image, that no image was provided, or that it needs an image to confidently answer?

\vspace{0.5em}
Provide brief reasoning, then respond with:\\
\texttt{<answer>true</answer>} if it mentions missing/unavailable image or uncertainty\\
\texttt{<answer>false</answer>} if it does not
\end{minipage}
}
\caption{LLM judge prompt (GPT-4o mini) used to assess uncertainty in base prompt responses.}
\label{fig:judge_prompt}
\end{figure}




\subsubsection*{Structured Prompt Condition}

In the structured prompt condition, the model is explicitly instructed to first classify the image into one of three categories before answering (Fig.~\ref{fig:structured_prompt}). This allows direct, deterministic evaluation without an external judge.

\begin{figure}[h]
\centering
\fbox{
\begin{minipage}{0.95\columnwidth}
\small
You are a visual analysis assistant. Examine the image carefully, then respond using EXACTLY this format:

\vspace{0.5em}
\texttt{IMAGE\_CLASS: <RELATED | UNRELATED\_REAL | BLANK\_OR\_NOISE>}\\
\texttt{ANSWER: <your response>}

\vspace{0.7em}
\textbf{Classification rules:}
\begin{itemize}
    \item \texttt{RELATED} $\rightarrow$ real image containing evidence relevant to the question
    \item \texttt{UNRELATED\_REAL} $\rightarrow$ real image whose content does NOT match the question topic
    \item \texttt{BLANK\_OR\_NOISE} $\rightarrow$ blank (black/white/grey/uniform) or random pixel noise
\end{itemize}

\vspace{0.3em}
\textbf{Answer rules:}
\begin{itemize}
    \item If \texttt{RELATED} $\rightarrow$ answer directly from what you see
    \item If \texttt{UNRELATED\_REAL} or \texttt{BLANK\_OR\_NOISE} $\rightarrow$ write exactly: ``I cannot answer based on the given image.''
\end{itemize}

\vspace{0.5em}
\textbf{Question:} \{question\} \{options\}
\end{minipage}
}
\caption{Structured prompt used for VLM evaluation with explicit image classification.}
\label{fig:structured_prompt}
\end{figure}



\subsection{Algorithm}
\label{app:algorithm}

Algorithm~\ref{alg:tclia_runtime} summarizes the runtime inference procedure for the proposed mirage detection framework. Given an image--question pair, the system first applies a lightweight blank/noise screen. Non-blank inputs are then processed through CLIP-based domain routing and \textsc{TC-LIA} feature extraction. The VLM is queried only for a structured preliminary class label and candidate answer, and the final decision is made by the ensemble classifier using pixel, routing, alignment, and VLM-derived features. If the ensemble predicts \textsc{Related}, the VLM answer is released; otherwise, the system returns a refusal response.

\begin{algorithm*}[t]
\small
\DontPrintSemicolon
\caption{Runtime Mirage Detection with TC-LIA}
\label{alg:tclia_runtime}

\KwIn{Image $x$, question $q$, VLM $M$, CLIP encoder $C$, trained ensemble $E$}
\KwOut{Decision $y \in \{\textsc{Related}, \textsc{Unrelated-Real}, \textsc{Blank/Noise}\}$ and released response}

Compute pixel statistics $b(x)$\;

\If{$b(x)$ indicates blank/noise with high confidence}{
    \Return{\textsc{Blank/Noise} and refusal response}\;
}

Compute domain probabilities $(p_m,p_n)$ using CLIP prompt routing\;

Extract layer-wise patch tokens $z_{\ell,i}$ from the CLIP ViT encoder\;

Compute patch--text similarities under the fixed terminal readout
$s_{\ell,i} \leftarrow \cos\bigl(W_v\,\texttt{ln\_post}(z_{\ell,i}),\, t(q)\bigr)$\;

Compute TC-LIA features: final cosine, late top-$k$, gain, slope, and IAS\;

Query the VLM for a structured class label and candidate answer\;

Fuse pixel, routing, TC-LIA, and VLM features using ensemble $E$\;

\If{$E(x,q)=\textsc{Related}$}{
    \Return{\textsc{Related} and the VLM answer}\;
}
\Else{
    \Return{predicted non-related class and refusal response}\;
}

\end{algorithm*}

\section{Additional TC-LIA Analyses}
\label{app:tclia_extra}

This section provides additional analyses that are useful for interpreting the main results but are placed in the appendix to preserve space in the main paper. These analyses explain why the five scalar \textsc{TC-LIA} features were chosen, how domain routing affects final cosine similarity, and how the detector affects answer quality on genuinely related inputs.

\begin{figure*}[h]
\centering
\includegraphics[width=0.98\textwidth]{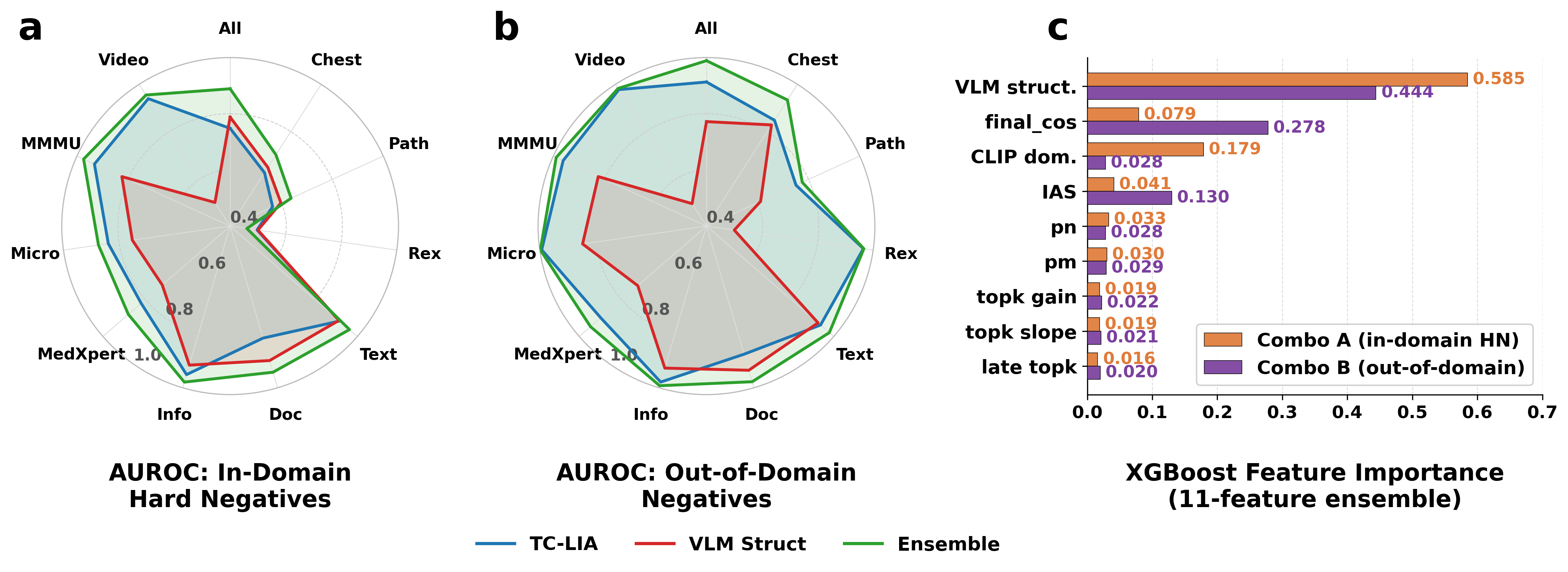}
\caption{Complementary roles of TC-LIA and VLM structured self-assessment
(Qwen2.5-VL-7B backbone).
\textit{(a)} AUROC for \textsc{Related} vs.\ \textsc{In-Domain Hard Negatives} and
\textit{(b)} AUROC for \textsc{Related} vs.\ \textsc{Out-of-Domain Negatives} across
ten domains; in both radar charts TC-LIA (blue), VLM structured self-assessment (red),
and the ensemble (green) are compared.
The ensemble consistently outperforms either component alone, with the largest gains
on in-domain hard negatives where neither TC-LIA nor VLM self-assessment is
individually sufficient.
\textit{(c)} XGBoost feature importance (gain) for the 11-feature ensemble on
in-domain hard negatives (Combo A, orange) and out-of-domain negatives (Combo B, purple).
\texttt{vlm\_class\_enc} is the dominant feature in both regimes;
TC-LIA features (\texttt{final\_cos}, \texttt{internal\_alignment\_score})
carry substantially greater weight for out-of-domain negatives, consistent
with TC-LIA's stronger raw AUROC in that setting.}
\label{fig:complementary}
\end{figure*}

\subsection{Rationale for the Five TC-LIA Features}

The five scalar TC-LIA features (Table~\ref{tab:tclia_feature_rationale}) were selected from nine candidate
features by evaluating each candidate's ability to separate related from
unrelated image--question pairs using AUROC on a controlled 200-pair
dataset (100 related, 100 unrelated, balanced across natural and medical
domains).
The nine candidates spanned three families: CLS-based similarity
(\texttt{late\_cls\_sim\_mean}, \texttt{slope\_cls\_sim}),
patch top-$k$ alignment
(\texttt{late\_patch\_topk\_mean}, \texttt{gain\_patch\_topk\_late\_minus\_early},
\texttt{slope\_patch\_topk}),
and attention-weighted patch alignment
(\texttt{late\_attn\_weighted\_sim\_mean},
\texttt{gain\_attn\_weighted\_late\_minus\_early},
\texttt{slope\_attn\_weighted},
\texttt{late\_attn\_topk\_semantic\_mean}).

CLS-based features were dropped because the CLS token aggregates global
image information at the final layer and is therefore highly correlated
with \texttt{final\_cos}; adding them introduced redundancy without
improving separability.
Attention-weighted features were dropped because they inherit the same
failure mode as raw RAPT: the encoder's self-attention is not conditioned
on the question text, so attention-weighted similarity reflects visual
salience rather than question relevance.
Patch top-$k$ features avoided this problem by selecting the patches with
the highest cosine similarity to the text embedding directly, bypassing
attention weights entirely.

The five retained features capture complementary and non-redundant evidence
signals.
\texttt{final\_cos} measures global image--text compatibility as a strong
baseline.
\texttt{late\_patch\_topk\_mean} measures whether localised question-relevant
evidence appears in the deeper visual layers.
\texttt{gain\_patch\_topk\_late\_minus\_early} measures whether that evidence
strengthens from early to late layers, suppressing layer-invariant shortcuts.
\texttt{slope\_patch\_topk} summarises the global trajectory of alignment
across all 32 layers.
\texttt{internal\_alignment\_score} (IAS) combines these signals into a
single compact scalar score:
\begin{equation}
\text{IAS} = 0.50 \cdot f_{\cos} + 0.25 \cdot \bar{s}_{\text{late}}
         + 0.15 \cdot \Delta_{\text{gain}} + 0.10 \cdot \beta_{\text{slope}}.
\end{equation}
An internal-only logistic-regression probe trained exclusively on the
patch top-$k$ features (excluding \texttt{final\_cos}) achieved AUROC
above 0.90, confirming that the layerwise trajectory carries separable
information beyond global cosine similarity alone.

\begin{table*}[h]
\centering
\small
\resizebox{\linewidth}{!}{%
\begin{tabular}{lll}
\hline
Feature & What it measures & Why it helps \\
\hline
\texttt{final\_cos} & final global CLIP similarity & strong baseline relevance signal \\
\texttt{late\_patch\_topk\_mean} & late local patch evidence & captures localized support \\
\texttt{gain} & late minus early alignment & suppresses global shortcuts \\
\texttt{slope} & trend across layers & captures evidence emergence \\
\texttt{IAS} & weighted composite & compact scalar decision score \\
\hline
\end{tabular}%
}
\caption{Interpretation of the five \textsc{TC-LIA} features used by the ensemble.}
\label{tab:tclia_feature_rationale}
\end{table*}

\subsection{Handling the No-Image Case}
\label{app:no_image}

When no image is supplied to the pipeline (for example, when an API call omits the image field or when the image file fails to load), rather than introducing a separate detection branch, we propose a simple unified treatment: define a zero tensor $\mathbf{0}$ of the expected spatial dimensions and compute $\tilde{x} = x \oplus \mathbf{0}$, where $\oplus$ denotes element-wise addition. For an informative image $x$, this operation is a no-op and the image passes through the pipeline unchanged. When no image is provided, $x$ itself is treated as $\mathbf{0}$, yielding an all-black representation whose pixel statistics (near-zero grayscale standard deviation and zero patch-variance coefficient of variation) are immediately flagged by the Stage~1 blank/noise gate, triggering a refusal response. This approach requires no additional detection module: the no-image case is subsumed by the existing blank/noise safeguard at no extra cost.

\subsection{TC-LIA Layer Projection Calibration}
\label{app:layer_calibration}

\begin{figure*}[h]
\centering
\includegraphics[width=\textwidth]{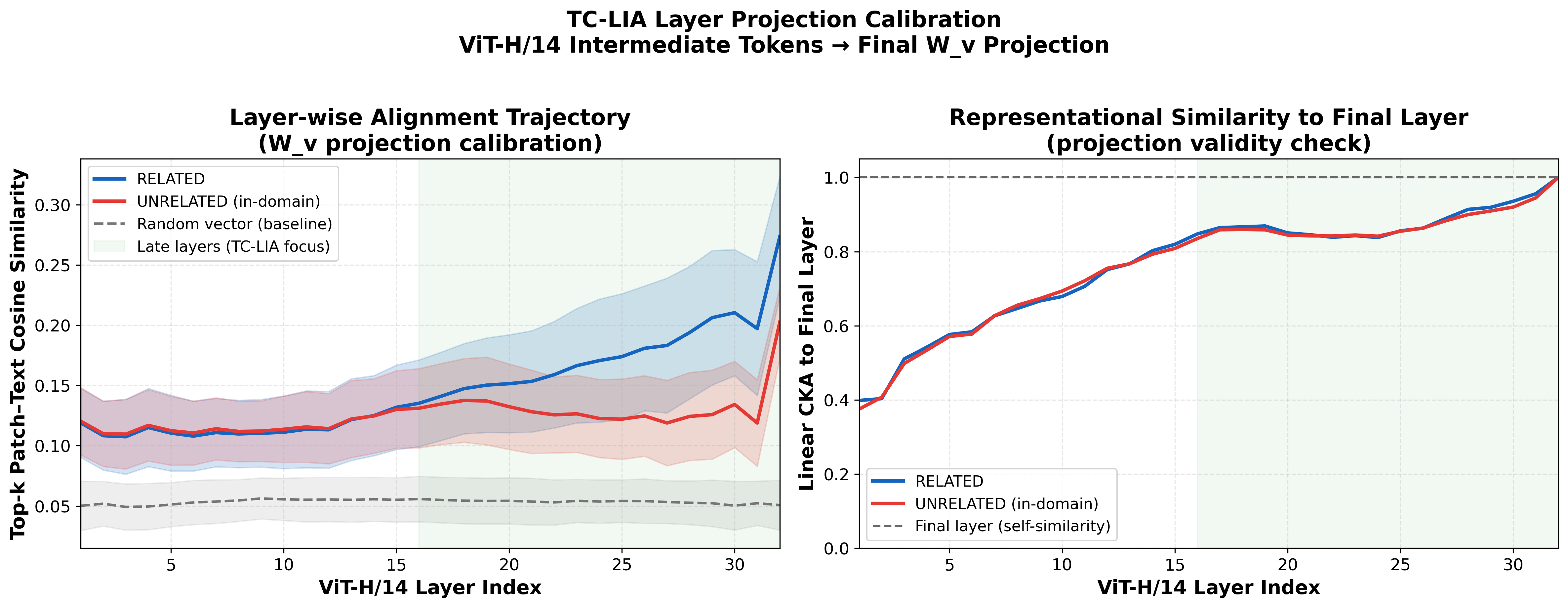}
\caption{%
\textbf{TC-LIA layer projection calibration (ViT-H/14).}
\textit{Left:} Layer-wise mean top-$k$ patch--text cosine similarity for
\textsc{Related} (blue) and \textsc{Unrelated} (red) pairs across the five
original benchmark domains (Chest VQA, PathVQA, TextVQA, DocVQA, InfoVQA).
Separation emerges in the TC-LIA focus window (shaded, layers 16--31);
a random-vector baseline (dashed) confirms the signal is semantic.
\textit{Right:} Linear CKA between projected intermediate-layer tokens and the
final-layer representation. The nearly identical curves for both conditions confirm
that $W_v$ projects into the same representational geometry regardless of image
content, validating the content-neutrality of the intermediate-token approximation.
}
\label{fig:layer_calibration}
\end{figure*}

Figure~\ref{fig:layer_calibration} directly validates the two core design assumptions
of TC-LIA: that the fixed final-layer projection $W_v$ provides a content-neutral
readout of intermediate tokens, and that the discriminative signal is concentrated
in the late vision-encoder layers.

\paragraph{Left panel: Layer-wise alignment trajectory.}
The mean top-$k$ patch--text cosine similarity is plotted at each ViT-H/14 layer for
\textsc{Related} (blue) and \textsc{Unrelated} (red) pairs pooled across the five
original benchmark domains (Chest VQA, PathVQA, TextVQA, DocVQA, InfoVQA), which
constitute the diagnostic subset on which this calibration analysis was run.
In early layers (0--14), both conditions produce nearly identical alignment values
($\approx$0.11--0.12), providing no discriminative signal.
In late layers (15--31), the \textsc{Related} curve rises steadily to
$\approx$0.20--0.21, while the \textsc{Unrelated} curve remains flat at
$\approx$0.12--0.14.
The separation emerges precisely in the TC-LIA focus window (shaded region, layers
16--31), directly justifying the use of \texttt{late\_patch\_topk\_mean} as a
discriminative feature.
A random-vector baseline ($\approx$0.05, flat throughout) confirms that the observed
separation is semantic rather than a geometric artifact of the $W_v$ projection.

\paragraph{Right panel: CKA validity check.}
To verify that $W_v$ does not favor one image type over the other when projecting
intermediate tokens, we compute Centered Kernel Alignment (CKA) between the projected
intermediate representations and the final-layer representations, separately for
\textsc{Related} and \textsc{Unrelated} pairs.
Both curves are essentially identical, rising from $\approx$0.40 at layer 0 to
$\approx$0.98 at layer 31, confirming that $W_v$ acts as a \emph{content-neutral
semantic ruler}: it does not introduce a systematic bias toward either condition, so
any alignment difference in the left panel is purely semantic. Together, these results
validate two properties required for TC-LIA: the $W_v$ approximation is
content-neutral and geometrically consistent across layers, and the discriminative
signal is concentrated in the late layers, motivating the late-layer focus of the
TC-LIA feature set.

\subsection{Representational Similarity Analysis: Late Layers Develop More Structured Representations}
\label{app:rsa}

\begin{figure*}[h]
\centering
\includegraphics[width=\textwidth]{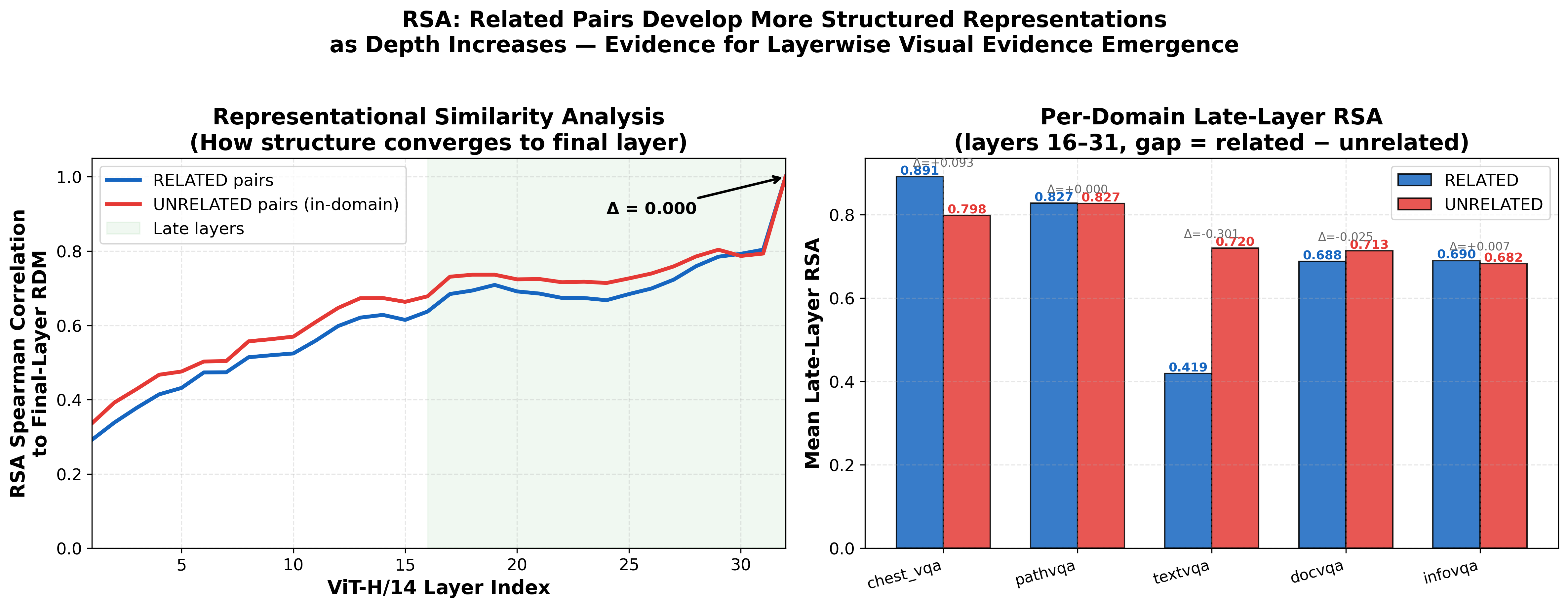}
\caption{%
\textbf{RSA evidence for layerwise visual evidence emergence.}
\textit{Left:} RSA Spearman correlation to the final-layer RDM for
\textsc{Related} (blue) and \textsc{Unrelated} (red) pairs across all layers.
Images sampled for unrelated pairs settle into their final representational
structure earlier (higher mid-layer RSA), while related-pair images are still
changing in late layers. Because the vision tower is unimodal (no question
conditioning), this reflects the representational dynamics of the two image
samples rather than question-conditioned processing.
\textit{Right:} Per-domain mean late-layer RSA (layers 16--31).
The gap $\Delta = \text{related} - \text{unrelated}$ varies substantially across
domains, with medical domains showing the largest related-image advantage.
}
\label{fig:rsa_analysis}
\end{figure*}

Figure~\ref{fig:rsa_analysis} provides complementary evidence for the TC-LIA
hypothesis through Representational Similarity Analysis (RSA).
For each ViT-H/14 layer, we compute the Spearman correlation between the pairwise
distance matrix of the projected patch representations and that of the final-layer
representations (\emph{RSA to final-layer RDM}), separately for \textsc{Related}
and \textsc{Unrelated} pairs.

\paragraph{Left panel: Overall RSA trajectory.}
Both curves converge to RSA $= 1.0$ at layer 31 by construction. We emphasise
that these RDMs are computed on the projected patch representations of the frozen
vision tower alone: the encoder is unimodal and never sees the question, so the
curves describe how quickly each image \emph{sample} settles into its final
representational geometry, not question-conditioned computation.
Across layers 5--25, the \textsc{Unrelated} curve (red) lies \emph{above} the
\textsc{Related} curve (blue): images sampled for unrelated pairs reach their final
representational structure \emph{earlier}, while images sampled for related pairs
continue to change in late layers. Because both conditions draw on the same domain
pools, this depth-dependent difference in representational dynamics is the visual-side
correlate of the late-layer alignment rise that TC-LIA measures, rather than evidence
that the encoder reasons about the question.

\paragraph{Right panel: Per-domain late-layer RSA.}
The mean late-layer RSA (layers 16--31) is reported per domain.
Medical domains (\texttt{chest\_vqa}, $\Delta = +0.093$; \texttt{pathvqa},
$\Delta \approx 0$) show the sharpest domain-level contrast.
\texttt{textvqa} shows a negative gap ($\Delta = -0.301$), as same-domain
text-rich images share structural properties regardless of semantic relevance,
consistent with TC-LIA's reduced discriminative power in document-style domains.
This per-domain variability motivates ensemble fusion rather than a single universal
IAS threshold. Taken together, RSA provides convergent evidence at the representational
level that related-pair images undergo more active late-layer transformation than
unrelated-pair images, i.e., the TC-LIA alignment rise tracks a representational
change specific to related-pair images rather than a projection artifact.

\subsection{In-Domain Hard Negatives: TC-LIA Scope and Ensemble Motivation}
\label{app:hard_negatives}

\begin{figure*}[h]
\centering
\includegraphics[width=\textwidth]{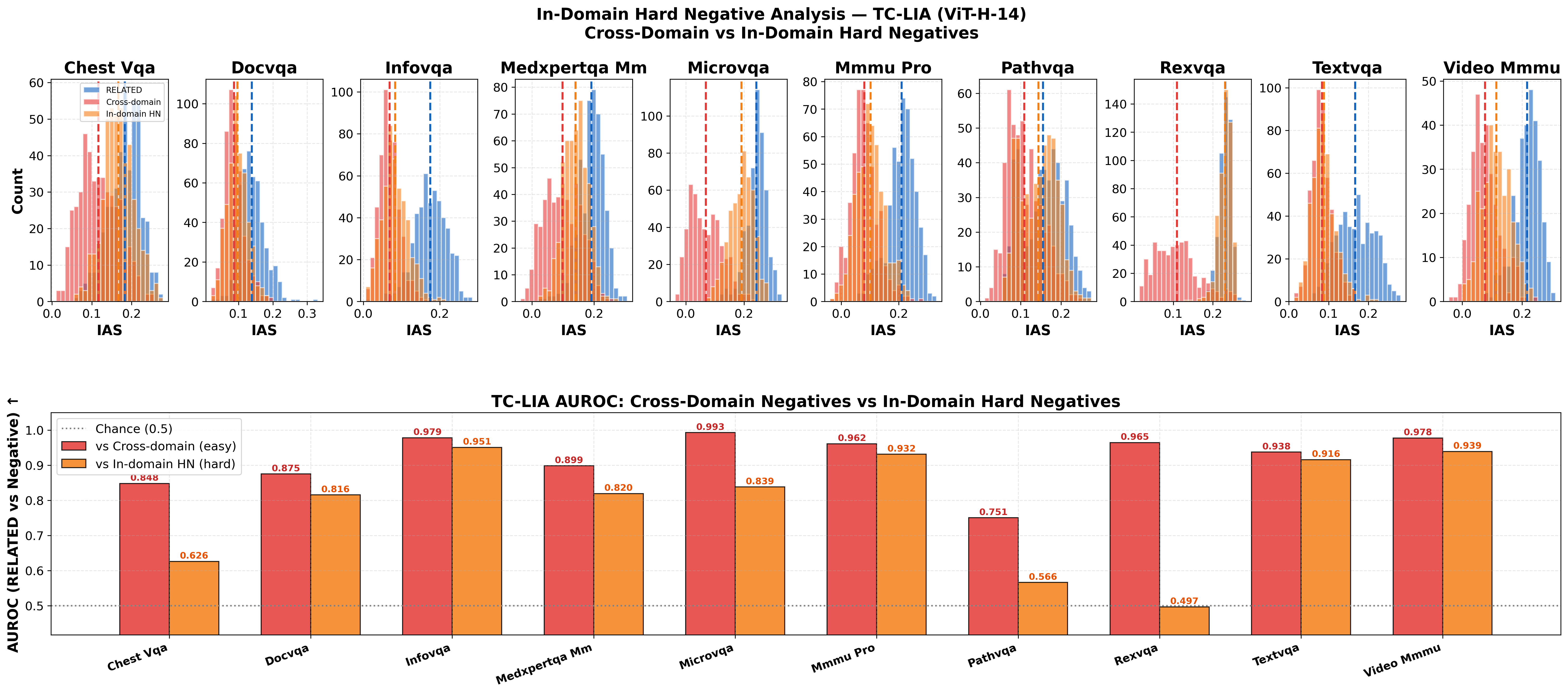}
\caption{%
\textbf{TC-LIA IAS on cross-domain versus in-domain hard negatives (10 domains).}
\textit{Top:} Per-domain IAS histograms for \textsc{Related} (blue),
cross-domain negatives (red), and in-domain hard negatives (orange).
\textit{Bottom:} TC-LIA AUROC against each negative type.
Cross-domain negatives are strongly separated across all domains
($0.751$--$0.993$).
In-domain hard negatives are also well-separated in 7 of 10 domains
($0.816$--$0.951$); the three exceptions (\texttt{chest\_vqa}, \texttt{pathvqa},
\texttt{rexvqa}) are visually homogeneous medical imaging domains where
domain-level priors dominate instance-level alignment signals.
This residual gap motivates the inclusion of structured VLM self-assessment as a
complementary component in the ensemble.
}
\label{fig:hard_negatives}
\end{figure*}

A key question for mirage detection is whether TC-LIA can distinguish a related image
from an \emph{in-domain} hard negative, an unrelated image drawn from the same visual
domain (e.g., a different chest X-ray paired with a chest VQA question).
Figure~\ref{fig:hard_negatives} reports this evaluation across all ten domains.

\paragraph{Top histograms: IAS distributions.}
Across all ten domains, cross-domain negatives (red) are clearly left-shifted relative
to \textsc{Related} (blue), confirming strong IAS separation for easy negatives.
In-domain hard negatives (orange) are more challenging by construction, yet on the full
dataset most domains show meaningful separation: the orange distribution is visibly
shifted from \textsc{Related} in domains such as \texttt{infovqa}, \texttt{mmmu\_pro},
\texttt{textvqa}, and \texttt{video\_mmmu}.
The hardest cases are \texttt{chest\_vqa}, \texttt{pathvqa}, and \texttt{rexvqa}, where
in-domain negatives overlap substantially with related images.

\paragraph{Bottom chart: AUROC by negative type.}
TC-LIA IAS achieves strong AUROC against cross-domain negatives across all ten domains
(range $0.751$--$0.993$, mean $\approx 0.921$).
Against in-domain hard negatives the picture is more nuanced: seven of ten domains
remain above $0.80$ (range $0.816$--$0.951$), while three domains show substantially
weaker performance: \texttt{pathvqa} ($0.566$), \texttt{rexvqa} ($0.497$), and
\texttt{chest\_vqa} ($0.626$).
All three are single-modality medical imaging domains in which images share a
narrow visual vocabulary, making instance-level discrimination from patch-level
alignment alone particularly difficult.

\paragraph{Interpretation and ensemble motivation.}
IAS provides reliable separation not only for cross-domain negatives but also for the
majority of in-domain hard negatives.
The residual gap is concentrated in a small number of visually homogeneous medical
domains where domain-level priors are unusually strong.
This characterization directly motivates the full ensemble design: TC-LIA provides a
powerful, computationally efficient signal that generalises well across most conditions,
while structured VLM self-assessment (Stage~4) supplies semantic instance-level
discrimination for the harder medical-imaging cases where patch-level alignment falls
short.
Figure~\ref{fig:complementary}(a,b) shows this complementary coverage.

\subsection{Encoder Generalization: TC-LIA Across Visual Backbones}
\label{app:encoder_gen}

\begin{figure*}[h]
\centering
\includegraphics[width=\textwidth]{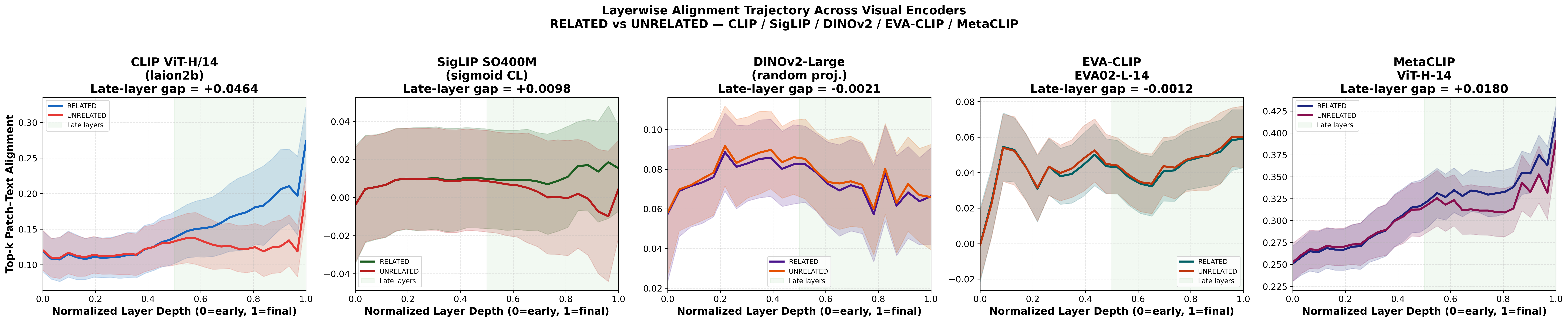}
\caption{%
\textbf{Layerwise alignment trajectories across five visual encoders.}
Each panel shows the mean top-$k$ patch--text alignment for \textsc{Related} and
\textsc{Unrelated} pairs as a function of normalized layer depth.
The late-layer gap $\Delta$ is annotated in each panel title.
Softmax-contrastive encoders (CLIP ViT-H/14, MetaCLIP) produce a positive
discriminative gap consistent with the TC-LIA hypothesis.
The sigmoid-loss encoder (SigLIP) shows marginal signal.
The text-free encoder (DINOv2) and the architecturally incompatible encoder
(EVA-CLIP) show near-zero gaps, serving as negative controls.
CLIP ViT-H/14 (laion2b) is the recommended backbone for TC-LIA.
}
\label{fig:encoder_gen}
\end{figure*}

TC-LIA's design rests on two structural properties of the visual encoder:
softmax contrastive training that aligns patch tokens with text embeddings, and a
$W_v$ projection that faithfully maps intermediate tokens into the joint CLIP space.
Figure~\ref{fig:encoder_gen} tests both properties by comparing the late-layer
alignment gap ($\Delta = \mu_{\text{related}} - \mu_{\text{unrelated}}$ over the
final 20\% of layers) across five encoders spanning different training objectives
and architectures.

\begin{itemize}

\item \textbf{CLIP ViT-H/14 (laion2b), $\Delta = +0.0464$.}
Strongest signal. \textsc{Related} alignment rises sharply after normalized depth
$\approx$0.6 while \textsc{Unrelated} remains flat, producing the widening late-layer
gap across samples. This result justifies CLIP ViT-H/14 as the primary TC-LIA backbone.

\item \textbf{MetaCLIP ViT-H-14, $\Delta = +0.0180$.}
Positive but weaker signal; separation appears only in the final 20\% of layers.
MetaCLIP uses the same softmax contrastive loss as CLIP on curated data: the same
training objective reproduces the late-layer phenomenon, albeit attenuated.

\item \textbf{SigLIP SO400M, $\Delta = +0.0098$.}
Marginal signal despite being a large contrastively trained model.
SigLIP uses sigmoid binary loss rather than softmax, which fundamentally changes how
patch tokens align to text and suppresses the late-layer amplification TC-LIA depends
on.

\item \textbf{DINOv2-Large, $\Delta = -0.0021$.}
Negative control. Both curves are nearly identical throughout all layers.
DINOv2 has no text supervision; the random projection into text space produces noise.
The near-zero gap confirms that TC-LIA requires text-visual co-training.

\item \textbf{EVA-CLIP EVA02-L-14, $\Delta = -0.0012$.}
Despite contrastive training, near-zero signal with very wide confidence intervals.
The $W_v$ approximation is likely lossy for EVA-CLIP's \texttt{TimmModel}
architecture, where intermediate token geometry may not map cleanly into the text
embedding space via \texttt{ln\_post} and \texttt{proj}.

\end{itemize}

These results establish that TC-LIA's signal comes from softmax contrastive
co-training (CLIP-style loss): the method generalizes to MetaCLIP
($\Delta = +0.0180$) but degrades with sigmoid loss (SigLIP) or without text
supervision (DINOv2), while EVA-CLIP's architecture likely breaks the $W_v$
projection assumption. CLIP ViT-H/14 (laion2b) is empirically the optimal backbone,
justified by direct comparison across encoder families.

\paragraph{Scope of encoder generality.}
The trajectory-level readout above degrades under sigmoid-contrastive objectives,
yet the SigLIP-SO400M \emph{final cosine} transfers strongly (0.870 on the nonblank
subset vs.\ 0.838 for CLIP ViT-H/14 on identical rows;
Section~\ref{sec:results}), which is why the main-text encoder ablation
instantiates the alignment signal with SigLIP's final cosine rather than its
trajectory. Pixel statistics, domain routing, structured VLM self-assessment,
and the boosting-fusion protocol remain unchanged; the fusion-stage ablation
reported in Section~\ref{sec:results} replaces only the five trajectory
features with SigLIP's final cosine and performs on par with the CLIP pipeline
(AUROC parity, no mirage-rate reduction), which is why the released system
instantiates the alignment signal with CLIP ViT-H/14.

\subsection{Domain-wise TC-LIA Score and IAS Distributions}
\label{app:Domain-wise IAS}

\begin{figure*}[h]
\centering
\includegraphics[width=\textwidth]{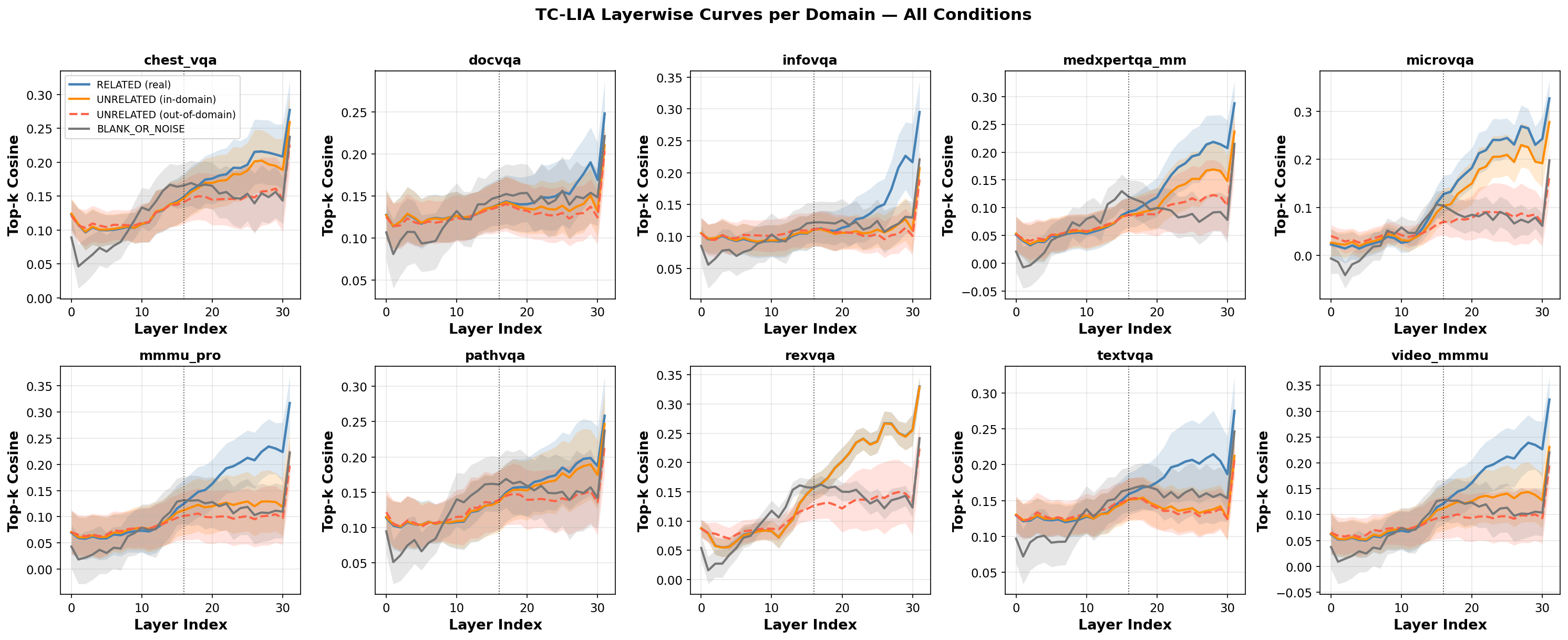}
\caption{
Per-domain TC-LIA layerwise alignment curves for all four input conditions across ten
domains ($5{\times}2$ grid).
Each panel shows the mean top-$k$ patch--text alignment (shaded band: $\pm$1 std)
across 32 ViT-H/14 layers.
\textsc{Related} inputs (blue) exhibit a consistent late-layer rise;
\textsc{Unrelated} out-of-domain inputs (red dashed) remain flat and low throughout.
In-domain hard negatives (orange) track below \textsc{Related} in most domains but
rise sharply in \texttt{chest\_vqa}, \texttt{pathvqa}, and \texttt{rexvqa}, where
domain-level visual priors dominate instance-level alignment signals.
\textsc{Blank/Noise} inputs (gray) show non-semantic intermediate behavior without
the sustained late-layer rise characteristic of related pairs.
}
\label{fig:layerwise_tclia_per_domain}
\end{figure*}

Figure~\ref{fig:layerwise_tclia_per_domain} provides a domain-wise view of the TC-LIA
alignment trajectory across all ten domains and all four input conditions.
\textsc{Related} examples (blue) consistently show a pronounced late-layer rise in
top-$k$ patch--text cosine similarity: question-relevant visual evidence emerges in
deeper ViT layers.
\textsc{Unrelated} out-of-domain inputs (red dashed) remain flat and low throughout,
indicating minimal semantic correspondence between question and image.
\textsc{Blank/Noise} inputs (gray) exhibit variable or non-semantic intermediate
behavior, consistently lacking the sustained late-layer rise seen in related pairs.

The in-domain hard negatives (orange) reveal a domain-dependent pattern.
In most domains (\texttt{docvqa}, \texttt{infovqa}, \texttt{medxpertqa\_mm},
\texttt{mmmu\_pro}, \texttt{textvqa}, \texttt{video\_mmmu}), the in-domain curve
remains clearly below the \textsc{Related} curve in late layers, indicating that
TC-LIA can still discriminate them.
However, in \texttt{chest\_vqa}, \texttt{pathvqa}, and \texttt{rexvqa} the orange
curve rises steeply in late layers and approaches or exceeds the \textsc{Related}
curve, directly explaining the near-chance in-domain AUROC observed for those domains
(see Appendix~\ref{app:hard_negatives}).
These three are single-modality medical imaging domains where images share a narrow
visual vocabulary, so patch-level alignment is driven by domain-level priors rather
than instance-level question relevance.
The remaining domains benefit from greater visual diversity, allowing late-layer
alignment to distinguish the specific image--question pairing.
These patterns support the use of late alignment, gain, and slope as complementary
TC-LIA features, while also motivating the ensemble's reliance on structured VLM
self-assessment for the hardest medical-imaging cases.

\begin{figure*}[h]
\centering
\includegraphics[width=\textwidth]{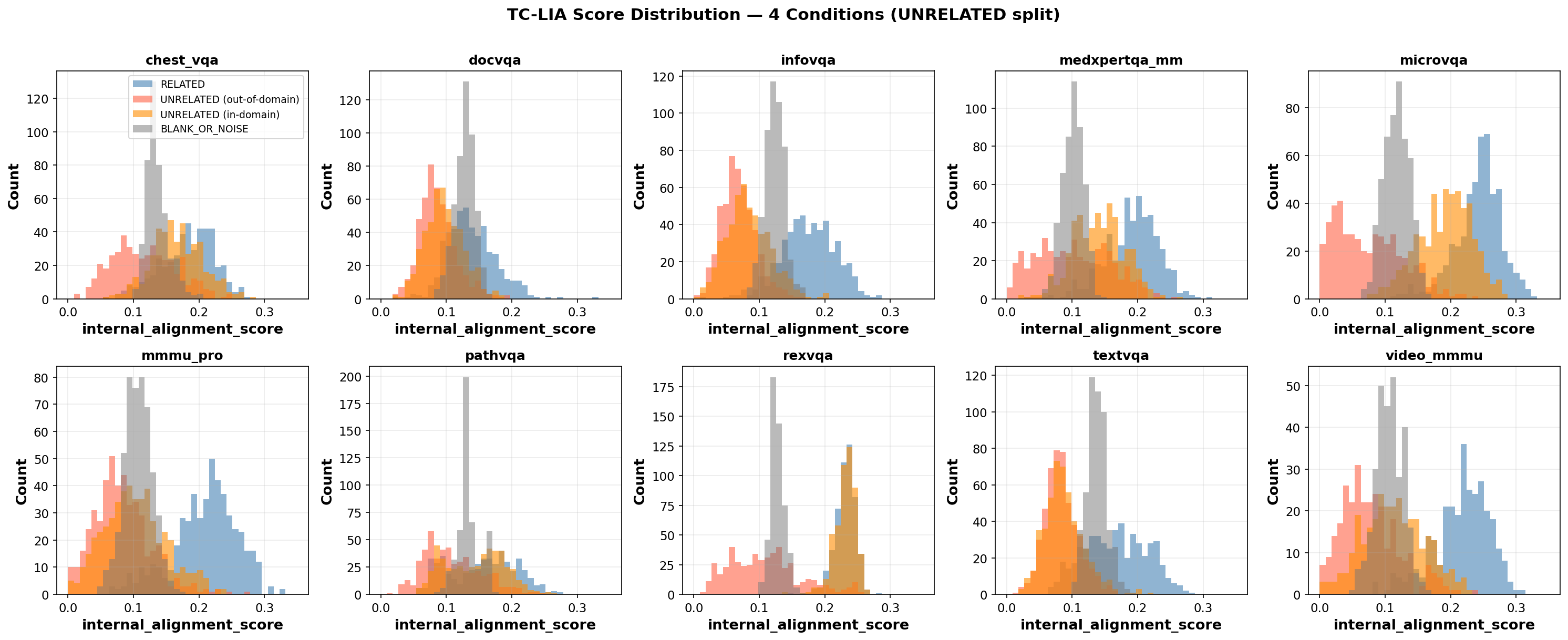}
\caption{
Per-domain distributions of the TC-LIA Internal Alignment Score (IAS) across all four
input conditions and ten domains.
\textsc{Related} image--question pairs generally shift toward higher IAS values, while
\textsc{Unrelated} out-of-domain examples concentrate at lower scores.
In-domain hard negatives overlap more with \textsc{Related}, consistent with the
layerwise curves in Fig.~\ref{fig:layerwise_tclia_per_domain}.
\textsc{Blank/Noise} examples form a narrower intermediate distribution.
These domain-wise patterns confirm that TC-LIA provides discriminative evidence,
computed prior to answer generation, for separating answerable inputs from
mirage-prone cases, with separation strength varying by domain.
}
\label{fig:layerwise_ias_per_domain}
\end{figure*}

Figure~\ref{fig:layerwise_ias_per_domain} expands the aggregate IAS distribution by
showing each of the ten domains separately across all four conditions.
The main pattern is consistent: \textsc{Related} examples shift toward larger IAS
values, \textsc{Unrelated} out-of-domain examples concentrate at low scores, and
\textsc{Blank/Noise} examples occupy a narrower intermediate range.
In-domain hard negatives overlap more with \textsc{Related}; the layerwise
curves above and the AUROC analysis in Appendix~\ref{app:hard_negatives}
show the same pattern.
The degree of separation varies by domain.

\subsection{Domain-Adaptive Routing}
\label{app:domain_routing}
Domain-adaptive routing improves medical-domain cosine similarity by switching to BioMedCLIP when the image is confidently medical, but it can slightly degrade document or natural-image domains when mixed-content images are routed imperfectly. We therefore use routing probabilities and the encoded routing decision as features rather than treating routing as a hard final decision. We confirm explicitly that the evaluated pipeline is a single-encoder system: all TC-LIA features are extracted with the frozen CLIP ViT-H/14 encoder for every non-blank input, and the routing outputs $(p_m, p_n, \texttt{clip\_domain\_enc})$ enter the Stage-5 ensemble purely as tabular features; no intermediate representation is ever taken from BioMedCLIP at benchmark time.


\subsection{Classifier Comparison}
Table~\ref{tab:appendix_classifier_comparison} compares ensemble classifier choices
on Qwen2.5-VL-72B across the full ten-domain benchmark.
All five boosting and tree-based classifiers perform within a narrow band
($88.0$--$88.4\%$ accuracy, $5.8$--$6.4\%$ mirage rate), indicating that the
ensemble's performance is robust to classifier choice rather than sensitive to
any single algorithm.
Random Forest achieves the best accuracy ($88.4\%$) and macro-F1 ($0.885$) with
the strongest cross-validation score ($85.7\% \pm 1.8\%$), and is used as the
default throughout the paper.
GradientBoosting shows the best cross-validation mean among boosting methods
($85.6\% \pm 2.2\%$), suggesting slightly better generalisation under distribution
shift.

\begin{table}[t]
\centering
\small
\caption{Classifier comparison on Qwen2.5-VL-72B (10 domains). CV Acc is
    5-fold cross-validation accuracy. }
\label{tab:appendix_classifier_comparison}
\resizebox{.85\columnwidth}{!}{%
\setlength{\tabcolsep}{5pt}
\begin{tabular}{lcccc}
\toprule
Classifier & Acc (\%)$\uparrow$ & MR (\%)$\downarrow$ & Macro-F1$\uparrow$ & CV Acc \\
\midrule
RandomForest      & \textbf{88.4} & 6.4          & \textbf{0.885} & $85.7 \pm 1.8\%$ \\
XGBoost           & 88.1          & \textbf{5.8} & 0.880          & $84.6 \pm 2.6\%$ \\
GradientBoosting  & 88.1          & 6.0          & 0.880          & $85.6 \pm 2.2\%$ \\
AdaBoost          & 88.1          & 6.1          & 0.881          & $85.2 \pm 1.7\%$ \\
LightGBM          & 88.0          & 5.9          & 0.879          & $84.7 \pm 2.5\%$ \\
\bottomrule
\end{tabular}%
}
\end{table}

\subsection{Answer Quality}
\label{app:quality_detail_sub}

Table~\ref{tab:quality_detail} reports per-condition answer quality across all fourteen
VLM backbones under base prompt and ensemble fusion.
\textsc{Unrelated}
and \textsc{Blank} inputs are scored against six canonical refusal phrases, so higher
scores indicate correct refusal.
\textbf{Bold} ensemble scores indicate improvement over the base prompt.

On \textsc{Unrelated} inputs, ensemble fusion raises BLEU and ROUGE-L from near zero
to $0.858$--$0.887$ and $0.871$--$0.896$ respectively.
On \textsc{Blank} inputs, the ensemble achieves perfect scores across all fourteen
models, confirming that blank detection is effectively deterministic.

\begin{table*}[h]
\centering
\small
\caption{Per-condition answer quality across 14 VLM backbones.
\textsc{Unrelated} \& \textsc{Blank}: scored vs.\ six canonical refusal phrases.
\textbf{Bold} ensemble scores indicate improvement over the base prompt.
$^\ddagger$InternVL2-8B: recomputed over all 10 domains with the released
training-pipeline configuration; its BERTScore cells are raw roberta-large F1.}
\label{tab:quality_detail}
\setlength{\tabcolsep}{4pt}
\renewcommand{\arraystretch}{0.78}
\begin{tabular*}{\textwidth}{@{\extracolsep{\fill}} ll cc cc cc}
\toprule
& & \multicolumn{2}{c}{BLEU} & \multicolumn{2}{c}{ROUGE-L} & \multicolumn{2}{c}{BERTScore} \\
\cmidrule(lr){3-4}\cmidrule(lr){5-6}\cmidrule(lr){7-8}
Condition & Model & Base & Ens. & Base & Ens. & Base & Ens. \\
\midrule
\multirow{14}{*}{\rotatebox[origin=c]{90}{\textsc{Unrelated}}}
& Qwen2.5-VL-72B    & 0.024 & \textbf{0.876} & 0.068 & \textbf{0.885} & 0.857 & \textbf{0.982} \\
& Claude-Opus-4.8     & 0.031 & \textbf{0.882} & 0.124 & \textbf{0.896} & 0.859 & \textbf{0.982} \\
& Qwen2.5-VL-7B     & 0.020 & \textbf{0.883} & 0.061 & \textbf{0.892} & 0.859 & \textbf{0.983} \\
& Qwen2.5-VL-32B    & 0.013 & \textbf{0.858} & 0.040 & \textbf{0.871} & 0.852 & \textbf{0.977} \\
& GPT-5.4           & 0.018 & \textbf{0.870} & 0.050 & \textbf{0.878} & 0.859 & \textbf{0.982} \\
& Llama-4-Scout     & 0.036 & \textbf{0.876} & 0.072 & \textbf{0.881} & 0.866 & \textbf{0.982} \\
& Llama-3.2-11B     & 0.017 & \textbf{0.887} & 0.038 & \textbf{0.888} & 0.867 & \textbf{0.985} \\
& LLaVA-OV-Qwen2-7B & 0.001 & \textbf{0.881} & 0.001 & \textbf{0.882} & 0.846 & \textbf{0.982} \\
& Gemma3-4B         & 0.027 & \textbf{0.881} & 0.072 & \textbf{0.889} & 0.863 & \textbf{0.983} \\
& Phi-3.5-Vision    & 0.004 & \textbf{0.882} & 0.008 & \textbf{0.883} & 0.843 & \textbf{0.981} \\
& BLIP2-OPT-2.7B    & 0.017 & \textbf{0.880} & 0.070 & \textbf{0.884} & 0.851 & \textbf{0.982} \\
& MiniCPM-V-2.6     & 0.007 & \textbf{0.881} & 0.012 & \textbf{0.881} & 0.844 & \textbf{0.981} \\
& InternVL2-8B      & 0.014 & \textbf{0.883} & 0.018 & \textbf{0.888} & 0.848 & \textbf{0.982} \\
& Kimi-K3           & 0.014 & \textbf{0.867} & 0.042 & \textbf{0.876} & 0.852 & \textbf{0.963} \\
\midrule
\multirow{14}{*}{\rotatebox[origin=c]{90}{\textsc{Blank}}}
& Qwen2.5-VL-72B    & 0.018 & \textbf{1.000} & 0.051 & \textbf{1.000} & 0.854 & \textbf{1.000} \\
& Claude-Opus-4.8     & 0.029 & \textbf{1.000} & 0.122 & \textbf{1.000} & 0.865 & \textbf{1.000} \\
& Qwen2.5-VL-7B     & 0.001 & \textbf{1.000} & 0.002 & \textbf{1.000} & 0.841 & \textbf{1.000} \\
& Qwen2.5-VL-32B    & 0.002 & \textbf{1.000} & 0.005 & \textbf{1.000} & 0.843 & \textbf{1.000} \\
& GPT-5.4           & 0.009 & \textbf{1.000} & 0.031 & \textbf{1.000} & 0.851 & \textbf{1.000} \\
& Llama-4-Scout     & 0.033 & \textbf{1.000} & 0.062 & \textbf{1.000} & 0.875 & \textbf{1.000} \\
& Llama-3.2-11B     & 0.014 & \textbf{1.000} & 0.032 & \textbf{1.000} & 0.868 & \textbf{1.000} \\
& LLaVA-OV-Qwen2-7B & 0.000 & \textbf{1.000} & 0.001 & \textbf{1.000} & 0.845 & \textbf{1.000} \\
& Gemma3-4B         & 0.002 & \textbf{1.000} & 0.007 & \textbf{1.000} & 0.856 & \textbf{1.000} \\
& Phi-3.5-Vision    & 0.029 & \textbf{1.000} & 0.033 & \textbf{1.000} & 0.847 & \textbf{1.000} \\
& BLIP2-OPT-2.7B    & 0.015 & \textbf{1.000} & 0.063 & \textbf{1.000} & 0.852 & \textbf{1.000} \\
& MiniCPM-V-2.6     & 0.004 & \textbf{1.000} & 0.007 & \textbf{1.000} & 0.842 & \textbf{1.000} \\
& InternVL2-8B      & 0.012 & \textbf{1.000} & 0.014 & \textbf{1.000} & 0.844 & \textbf{1.000} \\
& Kimi-K3           & 0.011 & \textbf{1.000} & 0.047 & \textbf{1.000} & 0.852 & \textbf{1.000} \\
\bottomrule
\end{tabular*}
\end{table*}

\subsection{Per-VLM Feature Importance}
\label{app:feat_importance}

Feature importance is reported as a diagnostic rather than a causal explanation;
nevertheless, the rankings provide useful evidence about which signals the ensemble
relies on and how that reliance shifts across VLM backbones.

\begin{figure*}[h]
\centering
\includegraphics[width=\textwidth]{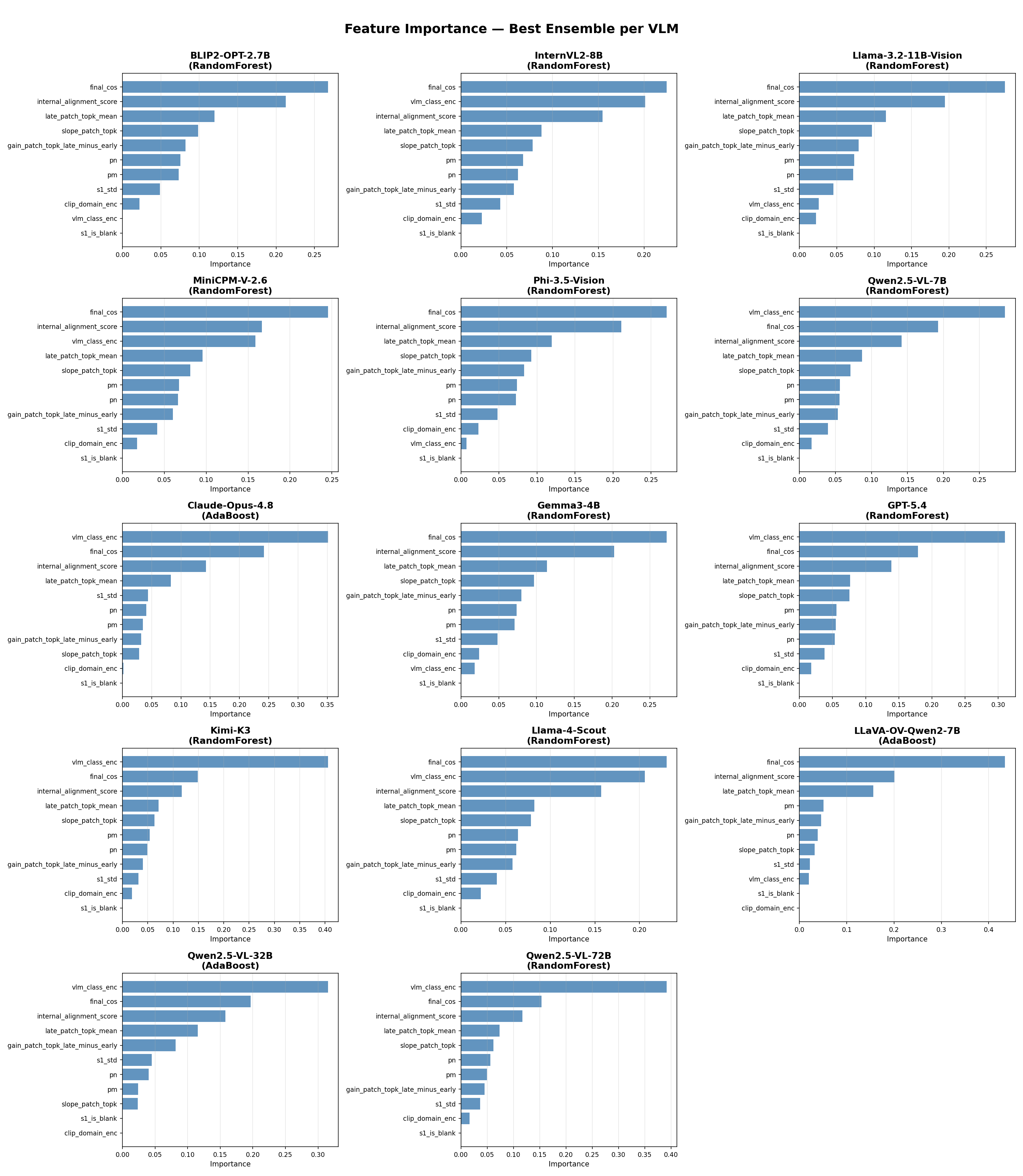}
\caption{Feature importance (gain) for the best ensemble classifier per VLM backbone
($5{\times}3$ grid, 14 models; InternVL2-8B recomputed over all 10 domains).
Two patterns emerge: stronger models (Qwen family, Kimi-K3, Claude-Opus-4.8,
GPT-5.4) are dominated by \texttt{vlm\_class\_enc}, while weaker models (BLIP2,
Gemma3-4B) are dominated by TC-LIA features (\texttt{final\_cos},
\texttt{internal\_alignment\_score}), reflecting the ensemble's adaptive reliance
on whichever signal is more reliable for a given backbone.}
\label{fig:feature_importance_all}
\end{figure*}

Figure~\ref{fig:feature_importance_all} shows feature importance for the best
ensemble classifier per VLM backbone across all fourteen models.
Two distinct patterns emerge, splitting along the reliability of the VLM's
structured self-assessment.

For the backbones with the most reliable structured self-assessment (Qwen2.5-VL-72B,
Qwen2.5-VL-32B, Qwen2.5-VL-7B, Claude-Opus-4.8, and GPT-5.4), \texttt{vlm\_class\_enc}
is the top feature, indicating that the
ensemble primarily relies on the VLM's own image classification signal.
\texttt{final\_cos} and \texttt{internal\_alignment\_score} follow as the second
and third most important features, providing a complementary TC-LIA check.

For the remaining backbones (BLIP2-OPT-2.7B, MiniCPM-V-2.6, Phi-3.5-Vision,
Gemma3-4B, Llama-3.2-11B, LLaVA-OV-Qwen2-7B, Llama-4-Scout, and
InternVL2-8B), \texttt{final\_cos}
takes the top position, with \texttt{internal\_alignment\_score} ranking second
on most panels and \texttt{vlm\_class\_enc} a close second for InternVL2-8B and
Llama-4-Scout.
When the VLM's structured output is less reliable, the ensemble
shifts weight toward TC-LIA's vision-encoder-based signals.

Across all backbones, \texttt{s1\_is\_blank} consistently ranks last, confirming
that blank detection is near-deterministic and adds little discriminative gain
beyond the other features.
The pixel-statistic features (\texttt{pn}, \texttt{pm}, \texttt{s1\_std}) and
domain routing (\texttt{clip\_domain\_enc}) occupy mid-range importance in all
panels, providing consistent but secondary contributions.
No single feature suffices alone across all backbones, confirming that ensemble
fusion over complementary signals is necessary for robust detection.

  \section{Additional Ablation Studies}
  \label{app:ablations}


\subsection{Leave-One-Domain-Out Generalization}

\begin{figure*}[t]
  \centering
  \includegraphics[width=\linewidth]{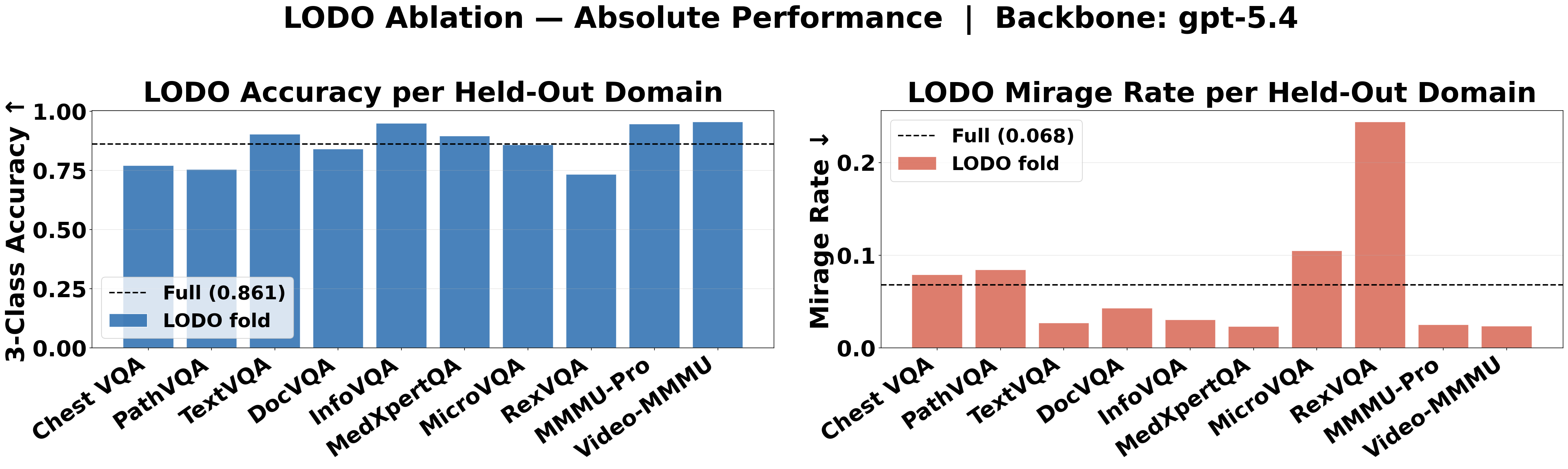}
  \caption{%
    LODO ablation (backbone: GPT-5.4).
    \textit{Left:} 3-class accuracy on the held-out domain; dashed line = full
    model (86.1\%).
    \textit{Right:} Mirage rate on the held-out domain; dashed line = full
    model (6.8\%).
    ReXVQA is the hardest held-out domain (MR = 24.4\%); Video-MMMU achieves
    the highest accuracy (95.5\%) and lowest mirage rate (2.4\%).
  }
  \label{fig:abl_lodo}
\end{figure*}

We evaluate cross-domain generalization by training the TC-LIA ensemble on
nine domains and testing on the held-out tenth, using GPT-5.4 as the VLM
backbone (full-model baseline: 86.1\% accuracy, 6.8\% mirage rate).
As Figure~\ref{fig:abl_lodo} and Table~\ref{tab:lodo} show, held-out accuracy
ranges from 73.4\% (ReXVQA) to 95.5\% (Video-MMMU), and five of the ten held-out
domains exceed the full-model baseline, so TC-LIA features transfer across
diverse visual domains without domain-specific fine-tuning.
ReXVQA is the hardest held-out domain: its 24.4\% mirage rate (more than three
times the full-model baseline) reflects the close-up radiograph distribution
that differs sharply from the other nine training domains, and ChestVQA and
PathVQA also fall below the full-model baseline in accuracy (77.1\% and 75.4\%)
as their medical-imaging visual statistics are underrepresented when left out.
Document and multi-modal reasoning domains (InfoVQA, MMMU-Pro, Video-MMMU,
TextVQA) generalise well or improve when held out, with mirage rates as low as
2.4\%, suggesting that TC-LIA patch-level features are particularly
discriminative for these domain types.

\begin{table}[h]
  \centering
  \caption{%
    LODO results per held-out domain (backbone: GPT-5.4).
    Full-model baseline: Acc = 86.1\%, MR = 6.8\%, Macro-F1 = 0.860.
    \textbf{Bold} = best; \underline{underline} = worst.
  }
  \label{tab:lodo}
  \setlength{\tabcolsep}{6pt}
  \begin{tabular}{lccc}
    \toprule
    \textbf{Held-Out Domain} & \textbf{Acc (\%)\,$\uparrow$} & \textbf{MR (\%)\,$\downarrow$}  & \textbf{Macro-F1\,$\uparrow$} \\
    \midrule
    Chest VQA      & 77.1          & 7.9           & 0.751 \\
    PathVQA        & 75.4          & 8.4           & 0.731 \\
    TextVQA        & 90.4          & 2.7           & 0.899 \\
    DocVQA         & 84.1          & 4.3           & 0.826 \\
    InfoVQA        & 94.9          & 3.1           & 0.950 \\
    MedXpertQA     & 89.6          & 2.3           & 0.889 \\
    MicroVQA       & 85.9          & 10.5          & 0.862 \\
    ReXVQA         & \underline{73.4} & \underline{24.4} & 0.763 \\
    MMMU-Pro       & 94.6          & 2.5           & 0.945 \\
    Video-MMMU     & \textbf{95.5} & \textbf{2.4}  & \textbf{0.955} \\
    \midrule
    \textit{Full model} & \textit{86.1} & \textit{6.8} & \textit{0.860} \\
    \bottomrule
  \end{tabular}
\end{table}

\subsection{Leave-One-VLM-Out Generalization}

\begin{figure*}[t]
  \centering
  \includegraphics[width=\linewidth]{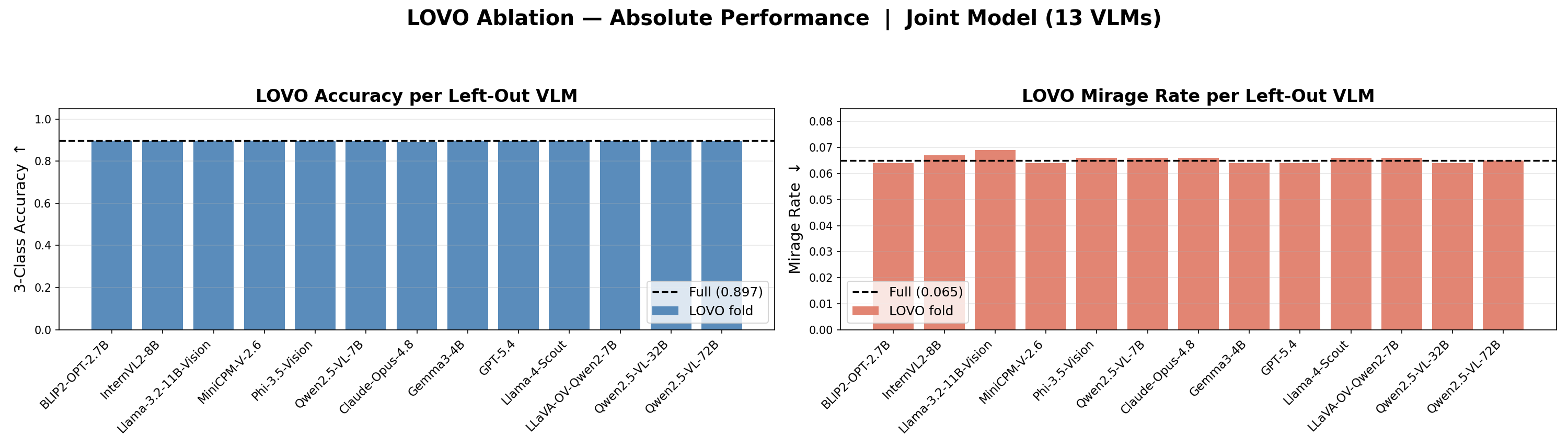}
  \caption{%
    LOVO ablation (joint model, 13 VLMs).
    \textit{Left:} 3-class accuracy when each VLM is removed; dashed line =
    full model (89.7\%).
    \textit{Right:} Mirage rate when each VLM is removed; dashed line = full
    model (6.5\%).
    Performance is stable across all 13 held-out VLMs ($<$0.6\% accuracy drop
    in the worst case), demonstrating that the ensemble does not depend
    critically on any single backbone.
  }
  \label{fig:abl_lovo}
\end{figure*}

We assess the contribution of individual VLMs to the joint ensemble by
constructing a single model that incorporates all 13 VLM class encodings as
features alongside the TC-LIA signal, then removing one VLM at a time
(full-model baseline: 89.7\% accuracy, 6.5\% mirage rate).
As shown in Figure~\ref{fig:abl_lovo} and Table~\ref{tab:lovo}, performance
is remarkably stable across all 13 held-out VLMs: accuracy ranges narrowly
from 89.1\% to 89.7\%, and mirage rate varies by less than 0.5 percentage
points in either direction.
The largest single-VLM contribution belongs to Claude-Opus-4.8, whose removal
causes the greatest accuracy drop ($\Delta\text{acc} = -0.61\%$,
$\Delta\text{F1} = -0.62\%$), consistent with its substantially lower base
mirage rate (57.3\%), which provides a distinctive training signal not replicated by
the other 12 models.
The near-zero deltas for most VLMs indicate that no single backbone is a critical
dependency: with 12 other collinear self-assessment features available, a withheld
VLM's prediction is readily compensated by the remaining ones. This experiment
therefore primarily demonstrates mutual redundancy across the joint feature space
rather than the dominance of the TC-LIA features themselves; the leave-all-but-one
study (Appendix~\ref{app:leave_all_but_one}), which removes all VLM self-assessment
signals at once, is the stronger test of feature reliance and cross-backbone
transfer.

\begin{table}[h]
  \centering
  \caption{%
    LOVO results per left-out VLM (joint model, 13 VLMs).
    Full-model baseline: Acc = 89.7\%, MR = 6.5\%, Macro-F1 = 0.898.
    $\Delta$Acc = accuracy change vs.\ full model.
    \textbf{Bold} = largest drop (most critical VLM).
    The LOVO study covers the 13-backbone joint-model cohort, which predates
    the final roster adjustment (Kimi-K3 was added to the main table
    afterwards and has no fold in this ablation).
  }
  \label{tab:lovo}
  \setlength{\tabcolsep}{5pt}
  \begin{tabular}{lcccc}
    \toprule
    \textbf{Left-Out VLM} & \textbf{Acc (\%)\,$\uparrow$} & \textbf{MR (\%)\,$\downarrow$} & \textbf{Macro-F1\,$\uparrow$} & \textbf{$\Delta$Acc (\%)} \\
    \midrule
    BLIP2-OPT-2.7B                 & 89.7 & 6.4 & 0.899 & $+0.03$ \\
    InternVL2-8B                   & 89.6 & 6.7 & 0.898 & $-0.09$ \\
    Llama-3.2-11B-Vision           & 89.7 & 6.9 & 0.899 & $-0.01$ \\
    MiniCPM-V-2.6                  & 89.7 & 6.4 & 0.899 & $+0.03$ \\
    Phi-3.5-Vision                 & 89.4 & 6.6 & 0.896 & $-0.23$ \\
    Qwen2.5-VL-7B                  & 89.6 & 6.6 & 0.898 & $-0.08$ \\
    Claude-Opus-4.8                & \textbf{89.1} & 6.6 & \textbf{0.892} & $\mathbf{-0.61}$ \\
    Gemma3-4B                      & 89.7 & 6.4 & 0.899 & $+0.04$ \\
    GPT-5.4                        & 89.6 & 6.4 & 0.897 & $-0.12$ \\
    Llama-4-Scout                  & 89.6 & 6.6 & 0.898 & $-0.06$ \\
    LLaVA-OV-Qwen2-7B              & 89.6 & 6.6 & 0.897 & $-0.12$ \\
    Qwen2.5-VL-32B                 & 89.7 & 6.4 & 0.898 & $-0.02$ \\
    Qwen2.5-VL-72B                 & 89.4 & 6.5 & 0.896 & $-0.27$ \\
    \midrule
    \textit{Full model (all 13)}   & \textit{89.7} & \textit{6.5} & \textit{0.898} & \textit{---} \\
    \bottomrule
  \end{tabular}
\end{table}

\subsection{Leave-All-But-One-VLM Generalization}
\label{app:leave_all_but_one}

The Leave-One-VLM-Out analysis above removes one backbone from a jointly trained
model. We now study the mirror-image direction, the one that matters for
deployment: the full ensemble is trained \emph{once} on a single source VLM's
training split and then evaluated, frozen, on every other backbone. At evaluation
time the only VLM-specific input is the held-out VLM's own structured
self-assessment feature; all TC-LIA features come from the frozen CLIP ViT-H/14
Metrics follow the paper's definitions (three-class accuracy including
gated blank rows; mirage rate; macro one-vs-rest AUROC on non-blank rows).
The held-out cohort of this study is the 13 backbones of the
Table~\ref{tab:main} roster excluding the source.
The per-backbone upper-bound macro
averages the per-backbone bounds over all evaluated backbones.

Table~\ref{tab:leave_all_but_one_macro} summarizes five source backbones. A
single-backbone ensemble does transfer to unseen VLMs, but transfer quality is
governed by the source's self-assessment style rather than by its in-domain
strength. The two conservative self-assessors, InternVL2-8B and GPT-5.4, beat the
training-free at deployment TC-LIA-only rule on both macro accuracy and macro mirage rate;
the strongest in-domain backbones (Qwen2.5-VL-72B, Claude-Opus-4.8,
Qwen2.5-VL-7B) over-trust their own self-assessment patterns, expecting a style
that weaker held-out backbones do not reproduce, which pushes held-out mirage
rates above the unsupervised 7.5\%. The ensemble
trained on InternVL2-8B sits within 0.7 percentage points of the per-backbone
upper bound while keeping over half of the ensemble's gain over the unsupervised
detector, and it beats that unsupervised rule on the majority of individual
backbones (e.g., 86.8\%/5.3\% on Qwen2.5-VL-72B and 86.2\%/4.9\% on
Qwen2.5-VL-7B). The classifier choice itself barely matters (a roughly one-point
spread across the five classifiers), so the transfer behavior is a property of
the feature set and the source's self-assessment distribution, not of classifier
selection.

\begin{table}[h]
  \centering
  \caption{%
    Leave-all-but-one-VLM generalization: the full ensemble is trained once on
    the source VLM (rows) and evaluated frozen on the 13 remaining backbones of
    Table~\ref{tab:main} (macro over held-out VLMs).
    Per-backbone upper bound and TC-LIA-only rows are from the main
    results (Table~\ref{tab:main}); TC-LIA-only is a threshold rule, so its
    AUROC is undefined.
  }
  \label{tab:leave_all_but_one_macro}
  \setlength{\tabcolsep}{4pt}
  \renewcommand{\arraystretch}{0.95}
  \begin{tabular}{llcccc}
    \toprule
    \textbf{Train VLM} & \textbf{Clf} & \textbf{AUROC$\uparrow$} & \textbf{Acc (\%)$\uparrow$} & \textbf{MR (\%)$\downarrow$} & \textbf{In-domain Acc/MR} \\
    \midrule
    Qwen2.5-VL-7B   & RF  & 0.878 & 81.7 & 13.2 & 87.3 / 6.0 \\
    GPT-5.4         & RF  & 0.880 & 84.2 & 6.6 & 86.1 / 6.8 \\
    Qwen2.5-VL-72B  & RF  & 0.873 & 82.0 & 11.2 & 88.4 / 6.4 \\
    Claude-Opus-4.8   & Ada & 0.869 & 82.3 & 10.1 & 88.2 / 6.2 \\
    InternVL2-8B    & RF  & \textbf{0.894} & \textbf{85.3} & \textbf{6.3} & 85.9 / 6.0 \\
    \midrule
    \textit{Per-backbone upper bound} & best-of-5 & \textit{0.903} & \textit{86.0} & \textit{6.3} & \textit{---} \\
    \textit{TC-LIA-only (unsupervised)} & threshold & --- & \textit{83.5} & \textit{7.5} & \textit{---} \\
    \bottomrule
  \end{tabular}
\end{table}

\begin{table*}[h]
  \centering
  \caption{%
    Leave-all-but-one-VLM per-backbone results for the best-transferring source
    (ensemble trained on InternVL2-8B only, then frozen). Each cell: Acc (\%) /
    MR (\%) / AUROC. TC-LIA Only is the VLM-agnostic unsupervised rule
    (Table~\ref{tab:main}); the upper bound is the per-backbone-trained ensemble
    of Table~\ref{tab:main}.
  }
  \label{tab:leave_all_but_one_detail}
  \setlength{\tabcolsep}{4pt}
  \renewcommand{\arraystretch}{0.85}
  \begin{tabular*}{\textwidth}{@{\extracolsep{\fill}} l cc ccc ccc}
    \toprule
    & \multicolumn{2}{c}{TC-LIA Only} & \multicolumn{3}{c}{Transferred (Train: InternVL2-8B)} & \multicolumn{3}{c}{Per-Backbone Upper Bound} \\
    \cmidrule(lr){2-3}\cmidrule(lr){4-6}\cmidrule(lr){7-9}
    \textbf{Held-Out VLM} & Acc$\uparrow$ & MR$\downarrow$ & Acc$\uparrow$ & MR$\downarrow$ & AUROC$\uparrow$ & Acc$\uparrow$ & MR$\downarrow$ & AUROC$\uparrow$ \\
    \midrule
    Qwen2.5-VL-72B        & 83.5 & 7.5 & 86.8 & 5.3 & 0.922 & 88.4 & 6.4 & 0.937 \\
    Claude-Opus-4.8         & 83.5 & 7.5 & 86.6 & 5.6 & 0.922 & 88.2 & 6.2 & 0.936 \\
    Kimi-K3               & 83.5 & 7.5 & 86.7 & 5.6 & 0.923 & 87.9 & 6.8 & 0.937 \\
    Qwen2.5-VL-7B         & 83.5 & 7.5 & 86.2 & 4.9 & 0.910 & 87.3 & 6.0 & 0.918 \\
    Qwen2.5-VL-32B        & 83.5 & 7.5 & 86.5 & 5.6 & 0.919 & 86.8 & 7.2 & 0.922 \\
    GPT-5.4               & 83.5 & 7.5 & 85.5 & 6.8 & 0.906 & 86.1 & 6.8 & 0.917 \\
    Llama-4-Scout         & 83.5 & 7.5 & 85.6 & 6.7 & 0.903 & 85.7 & 6.4 & 0.905 \\
    Llama-3.2-11B-Vision  & 83.5 & 7.5 & 84.7 & 7.4 & 0.883 & 85.0 & 5.9 & 0.884 \\
    LLaVA-OV-Qwen2-7B     & 83.5 & 7.5 & 84.7 & 5.8 & 0.863 & 85.0 & 6.1 & 0.878 \\
    MiniCPM-V-2.6         & 83.5 & 7.5 & 85.5 & 6.2 & 0.892 & 85.6 & 6.2 & 0.896 \\
    Gemma3-4B             & 83.5 & 7.5 & 84.0 & 8.3 & 0.874 & 84.6 & 6.1 & 0.878 \\
    Phi-3.5-Vision        & 83.5 & 7.5 & 82.5 & 5.1 & 0.843 & 84.6 & 5.9 & 0.876 \\
    BLIP2-OPT-2.7B        & 83.5 & 7.5 & 83.6 & 8.7 & 0.867 & 84.5 & 6.0 & 0.875 \\
    \bottomrule
  \end{tabular*}
\end{table*}

\section{Proofs for Theoretical Motivation}
\label{app:proofs}

This appendix provides full proofs for the theoretical statements in Section~\ref{sec:theory}. These results are intended as formal motivation for the \textsc{TC-LIA} feature design rather than as guarantees for the full nonlinear ensemble classifier.

\subsection{Proof of Lemma 1: Late-layer Alignment Separation}

Let $a_L$ denote the late-layer top-$k$ patch alignment statistic. Assume that $a_L|R$ and $a_L|U$ are sub-Gaussian with means $\mu_R$ and $\mu_U$, respectively, and with common proxy variance $\sigma^2$. Thus, for any $t>0$,
\[
P(a_L-\mu_R \leq -t \mid R)
\leq
\exp\!\left(-\frac{t^2}{2\sigma^2}\right),
\]
and
\[
P(a_L-\mu_U \geq t \mid U)
\leq
\exp\!\left(-\frac{t^2}{2\sigma^2}\right).
\]
Let $\Delta=\mu_R-\mu_U>0$ and define the decision threshold
\[
\tau=\frac{\mu_R+\mu_U}{2}.
\]
\emph{Scope.} The positive-separation assumption $\Delta>0$ is load-bearing: the
proof below only quantifies how quickly the error decays in the margin
$\Delta$, and the bound is vacuous whenever $\Delta \leq 0$ (i.e., when the two
class-conditional distributions overlap so strongly that the late-layer
statistic alone provides no usable margin). This collapsed-margin regime is not
hypothetical: it is exactly what the in-domain hard-negative analysis observes
(Section~\ref{sec:results}), and it is the formal reason the threshold rule is
augmented by the learned ensemble rather than deployed standalone.
The threshold classifier predicts $\hat{y}=R$ if $a_L>\tau$ and predicts $\hat{y}=U$ otherwise.

For a related pair, an error occurs when $a_L\leq \tau$. Therefore,
\[
\begin{aligned}
P(\hat{y}\neq R \mid R)
&= P(a_L\leq \tau \mid R)\\
&= P(a_L-\mu_R\leq \tau-\mu_R \mid R).
\end{aligned}
\]
Since
\[
\tau-\mu_R
=
\frac{\mu_R+\mu_U}{2}-\mu_R
=
-\frac{\Delta}{2},
\]
we have
\[
P(\hat{y}\neq R \mid R)
=
P(a_L-\mu_R\leq -\Delta/2 \mid R).
\]
Applying the sub-Gaussian lower-tail bound with $t=\Delta/2$ gives
\[
P(\hat{y}\neq R \mid R)
\leq
\exp\!\left(-\frac{\Delta^2}{8\sigma^2}\right).
\]

Similarly, for an unrelated pair, an error occurs when $a_L>\tau$. Thus,
\[
\begin{aligned}
P(\hat{y}\neq U \mid U)
&= P(a_L>\tau \mid U)\\
&= P(a_L-\mu_U>\tau-\mu_U \mid U).
\end{aligned}
\]
Since
\[
\tau-\mu_U
=
\frac{\mu_R+\mu_U}{2}-\mu_U
=
\frac{\Delta}{2},
\]
we obtain
\[
P(\hat{y}\neq U \mid U)
=
P(a_L-\mu_U>\Delta/2 \mid U).
\]
Applying the sub-Gaussian upper-tail bound with $t=\Delta/2$ gives
\[
P(\hat{y}\neq U \mid U)
\leq
\exp\!\left(-\frac{\Delta^2}{8\sigma^2}\right).
\]
Therefore, the per-class error of the threshold classifier is bounded by
\[
\exp\!\left(-\frac{\Delta^2}{8\sigma^2}\right).
\]
This proves Lemma 1.

In summary, Lemma 1 characterizes the \emph{separated} regime $\Delta>0$, where
the threshold rule's per-class error decays as $\exp(-\Delta^2/8\sigma^2)$. When
$\Delta \leq 0$ the bound degenerates to the trivial statement that the error is
at most $1$, and no threshold on $a_L$ can separate the classes; empirically this
corresponds to the near-chance in-domain hard-negative AUROCs reported in
Section~\ref{sec:results}. The system-level answer to margin collapse is not a
better threshold but additional evidence sources: the learned ensemble
(Section~\ref{sec:method}) combines the alignment trajectory with domain routing,
pixel statistics, and structured VLM self-assessment, which restores separability
precisely where the standalone alignment statistic fails.

\subsection{Proof of Lemma 2: Gain Cancels Layer-invariant Shortcuts}

Assume that the layer-wise alignment statistic can be written as
\[
a_\ell(x,q)=c(x,q)+r_\ell(x,q)+\epsilon_\ell,
\]
where $c(x,q)$ is a layer-invariant global image--text prior, $r_\ell(x,q)$ is a localized evidence term that varies across layers, and $\epsilon_\ell$ is noise. Let $\mathrm{early}$ and $\mathrm{late}$ denote the average alignment over the early and late layer sets:
\[
\mathrm{early}
=
\frac{1}{|\mathcal{L}_{E}|}
\sum_{\ell\in\mathcal{L}_{E}} a_\ell,
\qquad
\mathrm{late}
=
\frac{1}{|\mathcal{L}_{L}|}
\sum_{\ell\in\mathcal{L}_{L}} a_\ell.
\]
Substituting the decomposition of $a_\ell$ gives
\[
\mathrm{early}
=
c(x,q)
+
r_E
+
\epsilon_E,
\]
and
\[
\mathrm{late}
=
c(x,q)
+
r_L
+
\epsilon_L,
\]
where $r_E$ and $r_L$ are the average localized evidence terms in the early and late layers, and $\epsilon_E$ and $\epsilon_L$ are the corresponding average noise terms.

The gain feature is
\[
\begin{aligned}
\mathrm{gain}
&=
\mathrm{late}-\mathrm{early}\\
&=
\left(c(x,q)+r_L+\epsilon_L\right)
-
\left(c(x,q)+r_E+\epsilon_E\right)\\
&=
r_L-r_E+\epsilon_L-\epsilon_E.
\end{aligned}
\]
Thus, the layer-invariant term $c(x,q)$ cancels exactly. Consequently, the gain feature suppresses global similarity shortcuts that persist across layers and instead emphasizes the emergence of localized evidence from early to late layers. This proves Lemma 2.

\subsection{Proof of Proposition 1: Staged Blank Gating Decomposes Mirage Risk}

Let $B$ denote the event that the input is blank/noise, and let $U$ denote the event that the input is unrelated-real. A mirage error occurs when a non-related input, one whose ground-truth label
$Y$ is blank ($B$) or unrelated-real ($U$), is incorrectly passed as related. Therefore,
\[
\mathrm{MR}(g)
=
P(g(x,q)=R,\; Y\in\{B,U\}).
\]
Since $B$ and $U$ are disjoint classes,
\[
\begin{aligned}
\mathrm{MR}(g)
&=
P(g(x,q)=R,\;Y=B)\\
&\quad+
P(g(x,q)=R,\;Y=U).
\end{aligned}
\]

For a blank/noise input to be passed as related, it must first fail to be rejected by the blank/noise gate. Therefore,
\[
P(g(x,q)=R,\;Y=B)
\leq
P(g_B\neq B,\;Y=B).
\]
Using the product rule,
\[
P(g_B\neq B,\;Y=B)
=
P(B)P(g_B\neq B\mid B).
\]
By definition, $\epsilon_B=P(g_B\neq B\mid B)$, so
\[
P(g(x,q)=R,\;Y=B)
\leq
P(B)\epsilon_B.
\]

For unrelated-real inputs, the relevant failure mode is that the non-blank related/unrelated detector passes the input as related. Thus,
\[
P(g(x,q)=R,\;Y=U)
\leq
P(g_N=R,\;Y=U).
\]
Again applying the product rule,
\[
P(g_N=R,\;Y=U)
=
P(U)P(g_N=R\mid U).
\]
By definition, $\epsilon_U=P(g_N=R\mid U)$, so
\[
P(g(x,q)=R,\;Y=U)
\leq
P(U)\epsilon_U.
\]

Combining the two bounds gives
\[
\mathrm{MR}(g)
\leq
P(B)\epsilon_B+P(U)\epsilon_U.
\]
This proves Proposition 1. The result shows that the overall mirage risk can be reduced by separately controlling blank/noise failures through $g_B$ and semantic mismatch failures through $g_N$, which matches the staged design of the proposed detector.

\section{Negative and Developmental Experiments}
\label{app:negative}
In this section, we describe orthogonal approaches to TC-LIA that were attempted before TC-LIA. 

\subsection{Toy CLIP/BioMedCLIP Relevance Scoring}
We first constructed a small image--text relevance dataset containing natural images, medical images, related prompts, and unrelated prompts. Standard OpenCLIP/BioMedCLIP final cosine similarity performed perfectly on the tiny 16-pair setting. However, this was not evidence of a robust solution: the dataset was too small, the negatives were often easy cross-domain mismatches, and final cosine did not test whether internal localized evidence emerged across layers.

\textbf{Takeaway.} Initial CLIP-only experiments were useful as a sanity check but were not sufficient as a main method because they overestimated performance on small/easy negatives and did not explain layer-wise evidence emergence.

\subsection{Layer-wise Attention and GradCAM-style Metrics}
\label{appsec:gradcam}
We next tested several layer-wise attention and GradCAM-style metrics. These included attention concentration, late-layer attention mass, Gini-like attention sparsity, text-conditioned GradCAM maps, and visualizations over related, unrelated, and blank images. These metrics produced interpretable plots, but they were not stable enough as standalone classifiers. Attention maps could be sparse for both correct and incorrect cases, and blank or unrelated cases could still produce visually salient but semantically meaningless hotspots.

\begin{figure*}[t]
\centering
\includegraphics[width=\linewidth]{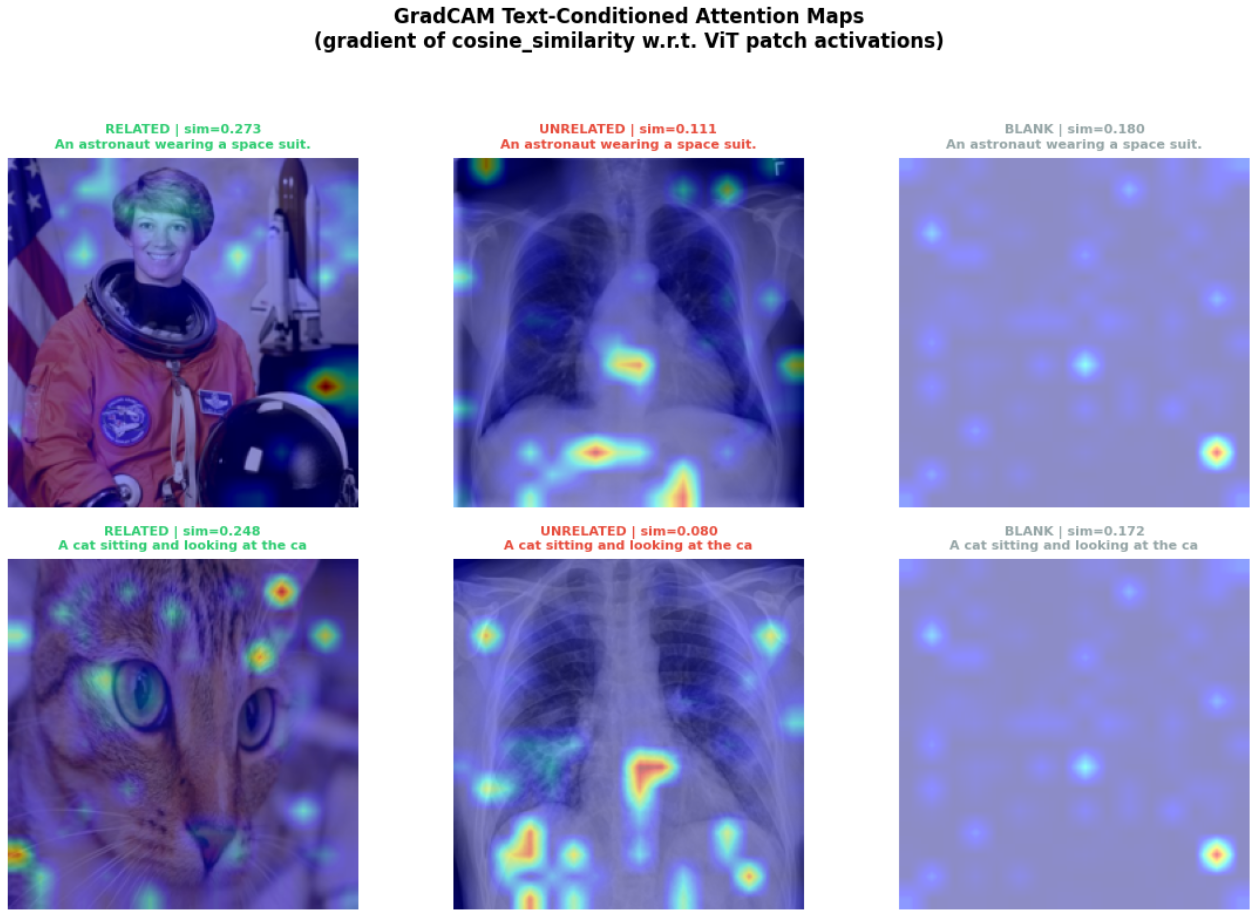}
\caption{%
  \textbf{Text-conditioned GradCAM maps across three input conditions}
  (columns: related, unrelated-real, blank).
  For unrelated inputs, GradCAM fires on spurious visually salient regions
  that bear no relation to the question.
  For blank inputs, activations are near-uniform or randomly scattered.
  GradCAM cannot separate related from unrelated inputs (AUROC 0.543),
  motivating the move to patch--text cosine alignment in TC-LIA.
}
\label{fig:gradcam_grid}
\end{figure*}

Figure~\ref{fig:gradcam_grid} shows text-conditioned GradCAM maps computed
over the late six ViT-H/14 layers for image--question pairs across three
conditions: related, unrelated-real, and blank.
Two failure patterns are immediately visible.
First, for unrelated inputs the GradCAM map fires on visually salient but
semantically irrelevant regions, for example, activating on structural
edges of a chest X-ray when the question concerns a rocket launch, or
highlighting fur texture on a cat image when the question asks about lung
opacity.
The model attends to whatever is visually prominent, not to what is
question-relevant.
Second, for blank images GradCAM produces near-uniform or randomly scattered
activations, confirming there is no stable grounding signal to detect.

These observations are confirmed quantitatively, with the caveat that this suite is
an exploratory diagnostic pilot over a small cohort (nine prompts $\times$ three
conditions, 27 samples), so the reported AUROC point estimates carry high
estimation variance and should be read qualitatively rather than as precise
measurements; the larger 400-probe RAPT study in Appendix~\ref{rapt-failed}
supports the same conclusion.
GradCAM achieved
only 22\% three-way accuracy and AUROC of 0.543, barely above chance.
The other six attention-only metrics we tested (auxiliary variants of RAPT-style
image-attention concentration, released with our code)%
all scored below 0.640 AUROC, with most below 0.450.
Attention rollout reached AUROC 0.750 but still failed to reliably separate
unrelated-real from related inputs in per-prompt evaluation.
These results confirm that visual attention in a CLIP-style encoder is not
text-conditioned at the feature level: the encoder attends to visually
salient patches regardless of whether they are semantically matched to the
query.
This motivated the shift to TC-LIA, which replaces attention weights with
direct patch--text cosine similarity computed against the question embedding.

\textbf{Conclusion:} Attention/saliency is useful for qualitative diagnosis, but a scalar mirage detector needs text-conditioned semantic alignment rather than attention concentration alone.

\begin{figure*}[t]
\centering
\includegraphics[width=\linewidth]{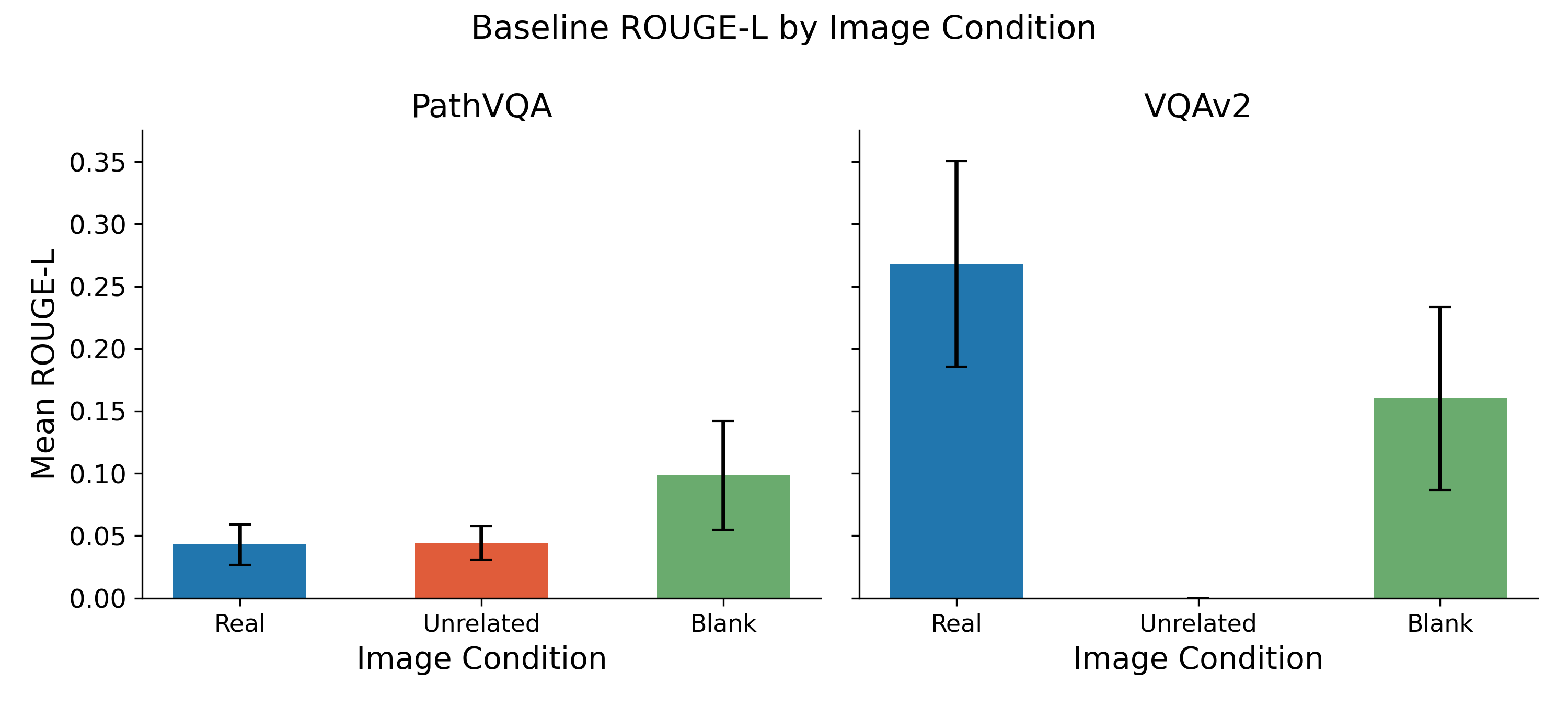}
\caption{Baseline ROUGE-L by condition across representative datasets. The plot shows that answer-overlap can remain nonzero even when the image is blank or unrelated, especially in easier settings, which makes ROUGE-L alone an unreliable signal of true visual grounding.}
\label{fig:appendix_baseline_rouge_condition}
\end{figure*}

Figure~\ref{fig:appendix_baseline_rouge_condition} further illustrates why output-overlap metrics alone are insufficient for mirage detection. Even when the visual evidence is absent or mismatched, the model can retain nontrivial lexical overlap with reference answers, suggesting that answer quality does not by itself certify groundedness.

\begin{figure*}[t]
\centering
\includegraphics[width=\textwidth]{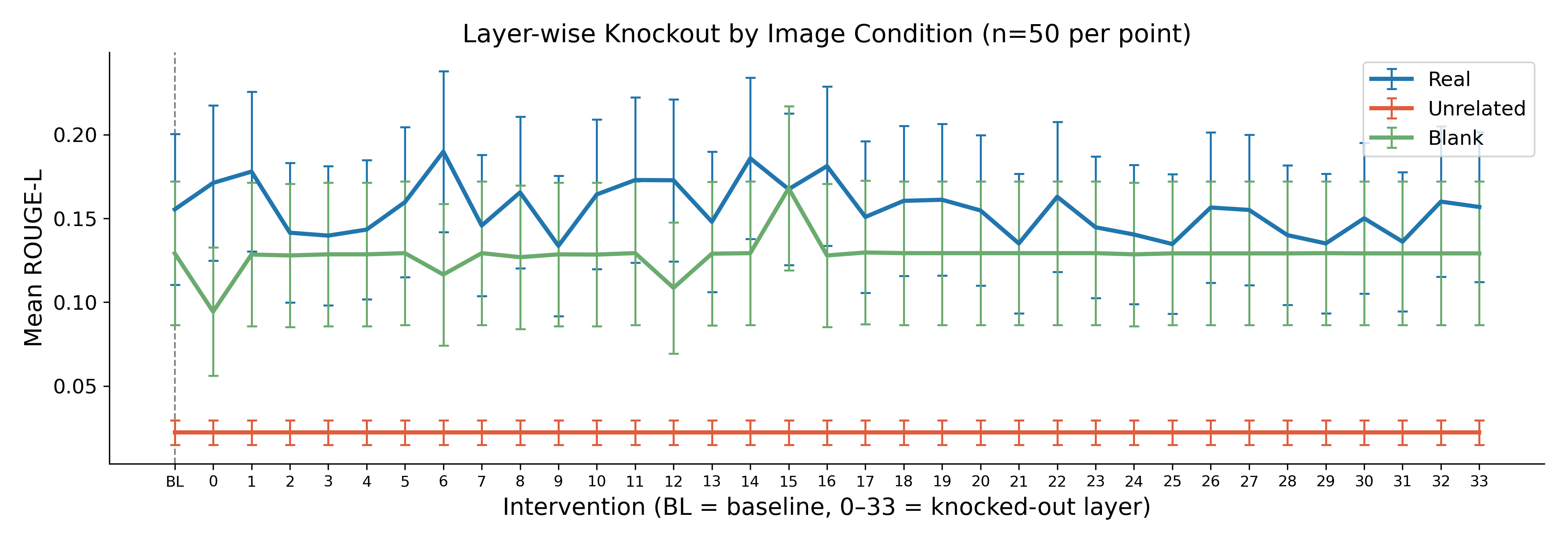}
\caption{Layer-wise answer-quality changes under attention knockout, grouped by condition. The effect varies across layers and input settings, indicating that intervention sensitivity is interpretable but not sufficiently stable to serve as a standalone mirage detector.}
\label{fig:appendix_layerwise_by_condition}
\end{figure*}

Figure~\ref{fig:appendix_layerwise_by_condition} shows that the impact of layer-wise intervention is highly condition-dependent. While some layers exhibit stronger degradation under perturbation, the overall patterns are not cleanly separable enough to define a reliable decision rule for mirage detection.

\subsection{RAPT Probe Replication}
\label{rapt-failed}
We also tested a RAPT-style probe inspired by the ``seeing but not believing'' line of work \citep{liu2026seeingbutnotbelieving}. The probe used Gemma3-4B and extracted relative attention per token for image and question spans over a 400-example, five-condition experiment. The conditions were: matched real image/question, unrelated image with real question, blank image with real question, real image with unrelated question, and real image with no question.

\begin{table*}[h]
\centering
\small
\begin{tabular}{llrrrrr}
\hline
Image condition & Question condition & $n$ & EM & F1 & Image RAPT & Question RAPT \\
\hline
Blank & Real & 80 & 6.2 & 11.2 & 0.105 & 1.211 \\
Real & None & 80 & 0.0 & 5.1 & 0.277 & 0.000 \\
Real & Real & 80 & 27.5 & 39.2 & 0.223 & 0.976 \\
Real & Unrelated & 80 & 0.0 & 2.0 & 0.153 & 1.381 \\
Unrelated & Real & 80 & 0.0 & 0.8 & 0.155 & 1.377 \\
\hline
\end{tabular}
\caption{RAPT-style diagnostic probe. RAPT reveals modality attention shifts but is not sufficient as a standalone mirage detector. Real/no-question inputs can show high image RAPT, while unrelated image/question pairs can still receive nontrivial image attention.}
\label{tab:rapt_probe}
\end{table*}

The key observation is that RAPT is diagnostic but not decisional. It shows whether the decoder allocates relative attention to image or question tokens, but it does not directly verify that the attended image content semantically answers the question. For example, unrelated image--question pairs still receive image attention, and real images without questions can produce high image RAPT even though no VQA answer is meaningful. This motivated TC-LIA's focus on patch--text semantic alignment rather than aggregate modality attention.

\begin{figure*}[t]
\centering
\includegraphics[width=0.95\textwidth]{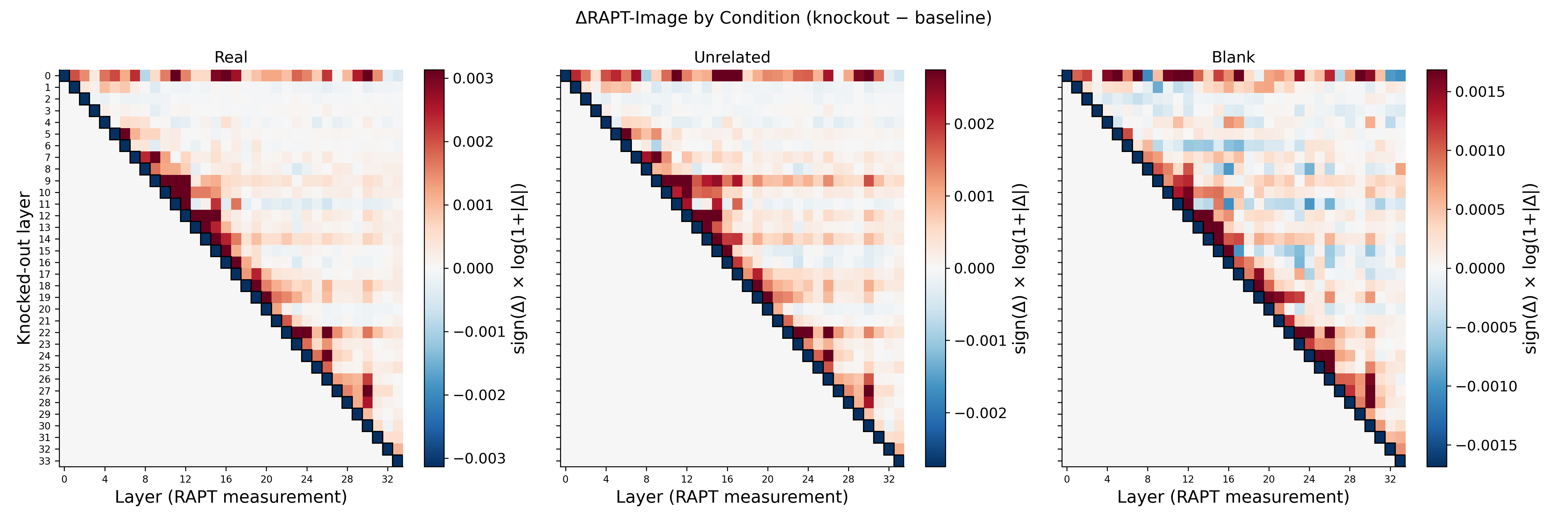}
\caption{Condition-wise RAPT shifts relative to the matched image-question setting. The perturbations clearly alter modality allocation, but the resulting patterns remain insufficient for deciding whether question-relevant visual evidence is actually present.}
\label{fig:appendix_rapt_delta_condition}
\end{figure*}

Figure~\ref{fig:appendix_rapt_delta_condition} complements the heatmap by summarizing how attention allocation moves under controlled perturbations. The deltas are meaningful as diagnostics, yet they still do not provide a robust criterion for separating related from unrelated-real cases.

\subsection{High-resolution Pet Dataset Sanity Check}
A larger natural-image sanity check used 100 Oxford-IIIT Pet image pairs with related and unrelated breed prompts. Final cosine and TC-LIA both performed extremely well because the task was comparatively easy. The internal-only probe also achieved high AUROC, confirming that intermediate features contain meaningful semantic information. However, this experiment did not represent the harder medical/document mirage setting, where same-domain and near-domain negatives are more challenging.

\subsection{Mixed Medical--Nonmedical 200-pair Experiment}
A mixed 200-pair experiment combined natural and medical images with related and unrelated prompts. TC-LIA features, including final cosine, late patch top-$k$, gain, slope, and internal alignment score, separated related and unrelated pairs well. This experiment motivated the final 10-domain benchmark evaluation but was still too small to support the main paper claim.

\subsection{Lessons from Failed Attempts}
\begin{itemize}
    \item Final CLIP cosine is a strong baseline but can be overly optimistic on easy negatives.
    \item Attention concentration alone does not imply semantic relevance.
    \item RAPT captures modality allocation but not whether the image evidence answers the question.
    \item GradCAM/SAM-style maps are valuable for visualization but require a scalar decision layer for deployment.
    \item The final system needs both semantic alignment and supervised fusion to handle blank, unrelated-real, and related cases simultaneously.
\end{itemize}

  \section{Detailed Differentiation from RAPT/VEA}
  \label{app:rapt_diff}

  \paragraph{What RAPT and VEA measure.}
  Relative Attention Per Token (RAPT) \citep{liu2026seeingbutnotbelieving}
  quantifies how much of a decoder's attention budget is allocated to image
  tokens versus text tokens at each layer.
  Visual Evidence Augmentation (VEA) extends this by studying how attention
  shifts under controlled perturbations.
  Both methods ask: \textit{does the model attend to the image?}
  Our method asks a different question: \textit{does the image contain
  question-relevant visual evidence?}
  This distinction determines whether a detector can separate
  \textsc{Unrelated-Real} inputs from \textsc{Related} ones.

\begin{figure*}[t]
\centering
\includegraphics[width=\linewidth]{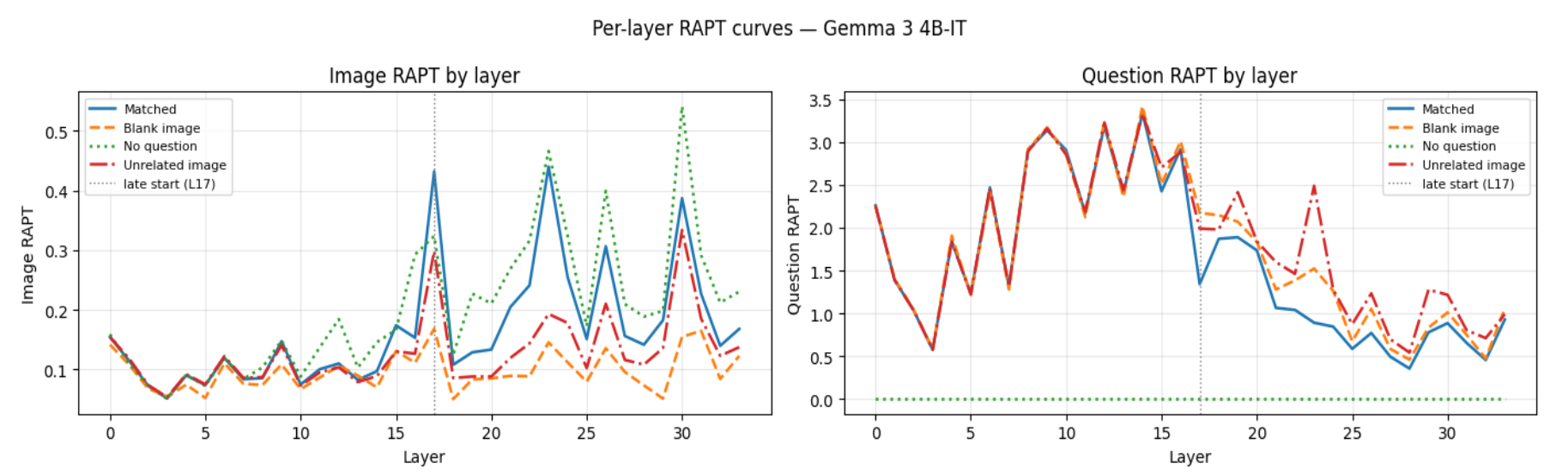}
\caption{%
  \textbf{Per-layer RAPT curves for Gemma3-4B across four input conditions.}
  \textit{Left:} Image RAPT by layer.
  \textit{Right:} Question RAPT by layer.
  The dotted vertical line marks the early/late split at layer 17.
  In the image RAPT panel, the \textit{unrelated image} condition (red)
  tracks closely with the \textit{matched} condition (blue) across all layers,
  confirming that the decoder allocates similar image attention regardless of
  whether the image is semantically relevant.
  The \textit{blank image} condition (orange) shows suppressed but non-zero
  image RAPT, while \textit{no question} (green) produces spuriously high
  image RAPT in late layers.
  In the question RAPT panel, the \textit{no question} condition collapses
  to zero as expected, while all other conditions remain indistinguishable
  from each other in late layers.
  Together, these curves show that RAPT tracks modality budget allocation
  but cannot separate related from unrelated inputs --- motivating
  TC-LIA's patch--text cosine alignment as the decisional signal.
}
\label{fig:rapt_perlayer}
\end{figure*}

  Figure~\ref{fig:rapt_perlayer} makes this limitation concrete.
  The \textsc{Unrelated-Real} curve (red) is nearly indistinguishable from
  the \textsc{Matched} curve (blue) in both image and question RAPT across
  all 34 layers of Gemma3-4B, even though the image bears no semantic
  relation to the question.
  The decoder allocates a similar attention budget to image tokens regardless
  of whether those tokens contain question-relevant content.
  RAPT therefore cannot separate the two conditions that matter most for
  mirage detection.

  \paragraph{Single-layer knockout cannot suppress visual grounding.}
  From the 400-probe observational RAPT suite (80 anchor samples $\times$ 5
  conditions across 8 VQA domains), we selected a 150-sample subset (three
  conditions: matched, unrelated, blank; 50 samples each) for the
  intervention experiment, giving $3 \times 50 \times 35 = 5{,}250$ inference
  passes on Gemma3-4B (35 states: baseline plus 34 layer knockouts).
  At each intervention, image tokens were masked from attending at one target
  layer $k$, and RAPT was re-measured at all other layers.

\begin{figure}[t]
\centering
\includegraphics[width=.7\columnwidth]{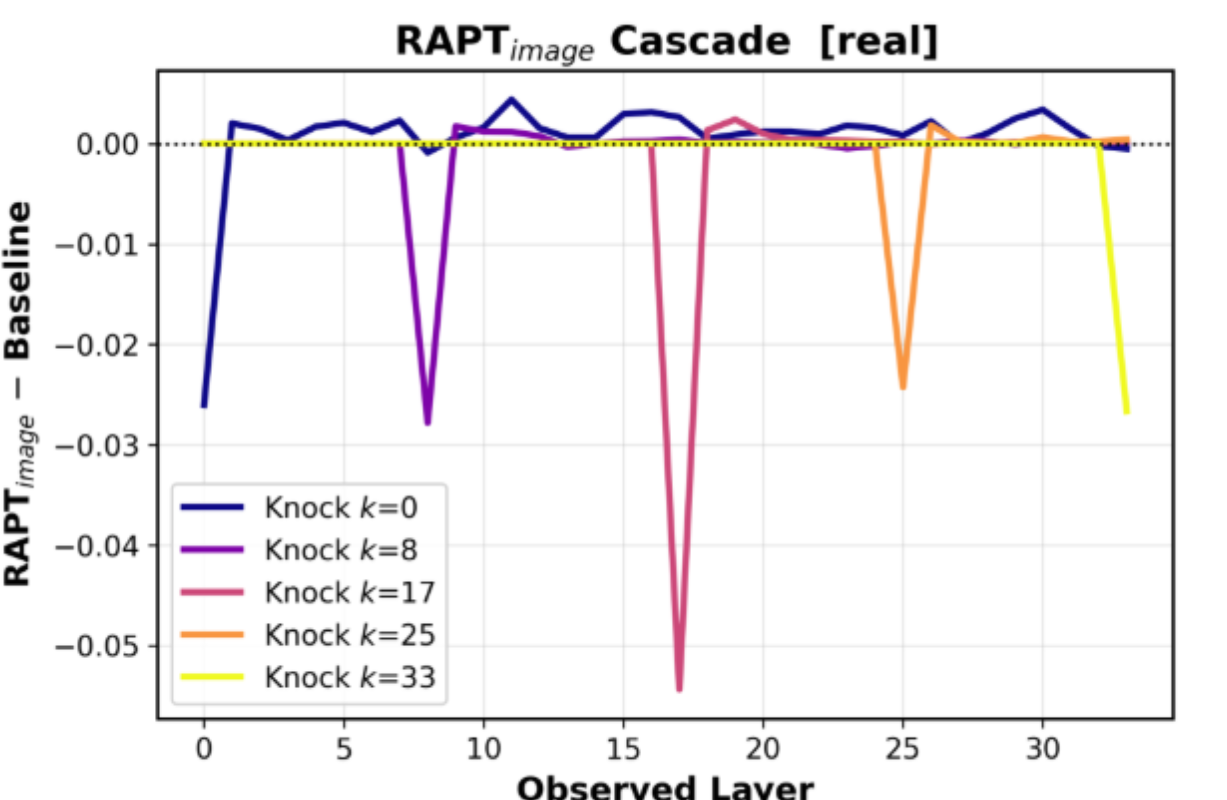}
\caption{%
  \textbf{RAPT$_{\text{image}}$ cascade under single-layer knockout (real inputs).}
  Each line shows the change in image-attention RAPT relative to baseline
  when layer $k$ is knocked out ($k \in \{0, 8, 17, 25, 33\}$).
  The knocked-out layer shows a sharp local drop (diagonal dip),
  but all other observed layers return to near-zero deviation,
  confirming that the network compensates for the blocked layer
  by redistributing image attention downstream.
  Visual grounding in Gemma3-4B is therefore distributed
  and cannot be suppressed by a single-layer intervention.
  This compensation behavior is demonstrated on Gemma3-4B, a concatenated
  prefix decoder-only architecture; whether cross-attention or resampler-based
  vision--language interfaces (e.g., Flamingo-style gated cross-attention,
  Q-Former resamplers) exhibit identical downstream compensation remains open
  for future work.
}
\label{fig:rapt_cascade}
\end{figure}

  Figure~\ref{fig:rapt_cascade} shows the result for five representative
  knockout layers ($k \in \{0, 8, 17, 25, 33\}$) on real inputs.
  Each knockout produces a sharp local dip only at the intervened layer;
  all other observed layers return immediately to near-zero deviation from
  baseline.
  The network compensates for the blocked layer by redistributing image
  attention downstream, a \textit{forward compensation} effect.

\begin{figure}[t]
\centering
\includegraphics[width=0.7\columnwidth]{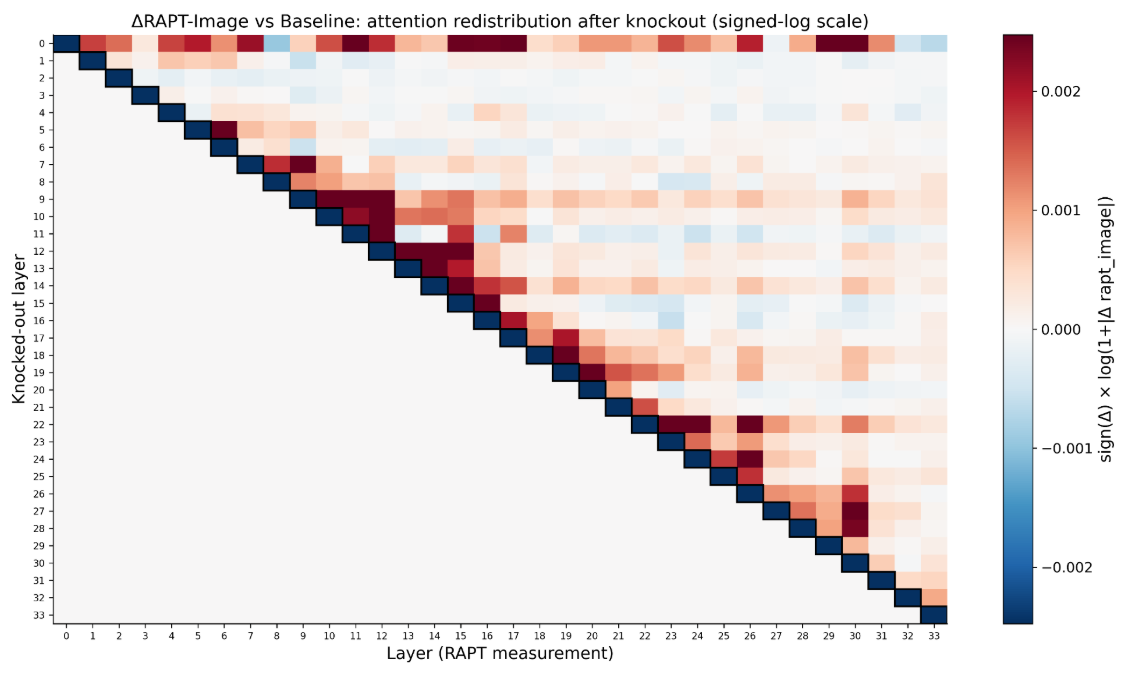}
\caption{%
  \textbf{$\Delta$RAPT$_{\text{image}}$ heatmap under attention knockout
  (signed-log scale, real inputs).}
  Rows are knocked-out layers; columns are RAPT measurement layers.
  The dark-blue diagonal marks the local suppression at the intervened
  layer. The red upper-triangle shows forward compensation:
  downstream layers increase their image-attention allocation
  to recover the blocked signal.
  Early knockouts (rows 0--10) produce the widest and strongest
  compensation because more downstream layers are available to
  redistribute attention.
}
\label{fig:rapt_heatmap}
\end{figure}

  Figure~\ref{fig:rapt_heatmap} generalises this across all 34 layers.
  The dark-blue diagonal marks local suppression at each knocked-out layer.
  The red upper-triangle confirms systematic forward compensation:
  downstream layers increase their image-attention allocation to recover
  the blocked signal, with early knockouts (rows 0--10) triggering the
  broadest redistribution because more downstream layers are available.
  No layer knockout systematically improved ROUGE-L or reduced mirage rate,
  confirming that visual grounding in Gemma-3 is distributed across the
  network and cannot be eliminated by single-layer intervention.

  \paragraph{How TC-LIA differs.}
  Where RAPT tracks \textit{how much} attention flows to image tokens,
  TC-LIA tracks \textit{whether question-conditioned patch evidence emerges
  in late vision-encoder layers}.
  Specifically, TC-LIA computes top-$k$ patch--text cosine similarity at
  each of the 32 ViT-H/14 transformer blocks using the frozen text embedding
  of the question, not the decoder's attention weights.
  A related image--question pair exhibits a rising late-layer alignment
  trajectory; an unrelated pair shows a flat or declining one, regardless
  of how the decoder allocates its attention budget.
  
  This design sidesteps both failure modes of RAPT.
  First, it operates on the encoder before any decoder attention is computed,
  so there is no mechanism for downstream layers to compensate for a blocked
  signal.
  Second, it conditions alignment on the specific question text rather than
  measuring aggregate modality attention, making it sensitive to semantic
  mismatch rather than just image presence.
  TC-LIA is therefore \textit{decisional}: its scalar features feed
  directly into an ensemble classifier, rather than being merely diagnostic.

\begin{table*}[h]
\centering
\small
\begin{tabular}{p{0.20\linewidth}p{0.36\linewidth}p{0.36\linewidth}}
\hline
Aspect & RAPT/VEA-style work & Our TC-LIA mirage detector \\
\hline
Primary question & Does the VLM attend to visual evidence and use it correctly? & Is there question-relevant visual evidence present before generation? \\
Input assumption & Usually assumes the image is relevant and evidence exists. & Explicitly includes related, unrelated-real, and blank/noise inputs. \\
Signal & Decoder-side attention mass over text/image/evidence tokens. & Vision-encoder patch embeddings aligned with text across layers. \\
Main operation & Highlight or mask evidence regions to improve answer generation. & Classify image--question pair and decide answer vs abstain. \\
Failure addressed & ``Seeing but not believing'': evidence is present but underutilized. & Mirage risk: answer is generated despite absent or mismatched evidence. \\
Primary metric & EM/F1 improvement and evidence attribution quality. & Mirage rate, three-class accuracy, macro-F1, related recall. \\
Use in our paper & Related work and auxiliary diagnostic. & Main method. \\
\hline
\end{tabular}
\caption{Conceptual distinction between RAPT/VEA-style evidence utilization and TC-LIA mirage detection.}
\label{tab:rapt_diff}
\end{table*}

\section{SAM3 Grounding as a Diagnostic Rather than a Detector}
\label{app:sam3}
As a qualitative sanity check, we also visualize prompt-conditioned SAM3-style grounding maps \citep{carion2025sam3}. For related
examples, the segmentation/grounding mask should cover plausible question-relevant image regions. For
unrelated-real examples, the mask is expected to be absent, diffuse, low-confidence, or semantically inconsistent
with the prompt. For blank/noise examples, no stable grounding should be produced.

\begin{figure*}[t]
\centering
\includegraphics[width=\textwidth]{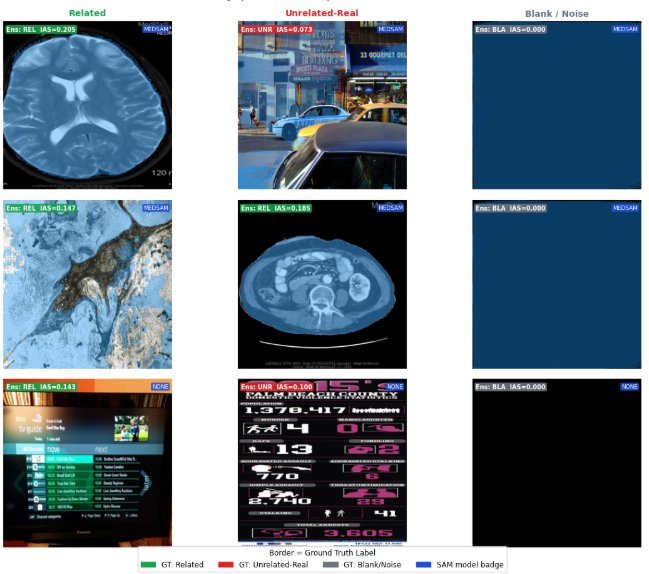}
\caption{SAM3-style prompt-conditioned grounding sanity check. Related examples may not always
produce stable evidence regions, whereas unrelated and blank/noise cases can produce
unstable grounding.}
\label{fig:sam3}
\end{figure*}

Figure~\ref{fig:sam3} shows representative SAM3-style grounding maps across all three input conditions.
In each panel, the question (or a key phrase extracted from it) is the
grounding prompt for the segmenter, and the returned mask is visualised on the input
image.

Although SAM3-style grounding provides useful qualitative evidence, we found that it is not sufficiently reliable as a standalone mirage detector. For some \textsc{Related} examples, the generated masks missed clinically or semantically salient regions, especially when the relevant evidence was subtle, diffuse, or not easily described by a short phrase. Conversely, for \textsc{Unrelated-Real} or even \textsc{Blank/Noise} examples, the grounding model occasionally produced spurious masks on visually salient but question-irrelevant regions, indicating that mask presence alone does not guarantee question-relevant evidence. These failure modes suggest that grounding visualizations are valuable for interpretation and error analysis, but they should not be treated as a decisive answerability test computed ahead of generation. We therefore use SAM3-style masks as auxiliary diagnostic evidence, while the main mirage detector relies on TC-LIA alignment features and ensemble fusion.


  \section{Qualitative Result Cards}
  \label{app:qual_cards}
\begin{figure*}[t]
  \centering

  \fbox{%
  \begin{minipage}{0.96\linewidth}
  \textbf{Example A: Medical Question / Related Image} \hfill \textbf{Model: Gemma3-4B}\\[0.5em]
  \begin{minipage}{0.23\linewidth}
  \centering
  \fbox{%
  \includegraphics[width=0.95\linewidth]{figures/chest_vqa_related.jpg}
  }
  \end{minipage}
  \hfill
  \begin{minipage}{0.72\linewidth}
  \textbf{Question:} "is there airspace consolidation on the left side?"\\
  \textbf{Raw VLM output:} "Yes, there appears to be airspace consolidation (likely pneumonia) on the left side of the lung field."\\
  \textbf{TC-LIA:}\\
  final\_cos=0.142278\\
  late=0.124805\\
  gain=0.028265\\
  slope=0.001867\\
  IAS=0.106767\\
  \textbf{VLM class:} "RELATED"\\
  \textbf{Ensemble decision:} \textsc{Unrelated-Real} (over-refusal).\\
  \textbf{Released output:} "I cannot answer based on the given image."
  \end{minipage}
  \end{minipage}}

  \vspace{0.5em}

  \fbox{%
  \begin{minipage}{0.96\linewidth}
  \textbf{Example B: Medical Question / Unrelated-Real Image} \hfill \textbf{Model: Gemma3-4B}\\[0.5em]
  \begin{minipage}{0.23\linewidth}
  \centering
  \fbox{%
  \includegraphics[width=0.95\linewidth]{figures/chest_vqa_unrealted.jpg}
  }
  \end{minipage}
  \hfill
  \begin{minipage}{0.72\linewidth}
  \textbf{Question:} "is there airspace consolidation on the left side?"\\
  \textbf{Raw VLM output:} "Yes, there appears to be airspace consolidation (likely representing pneumonia) in the left lower lobe of the lung."\\
  \textbf{TC-LIA:}\\
  final\_cos=0.110989\\
  late=0.111715\\
  gain=0.013821\\
  slope=0.001162\\
  IAS=0.085613\\
  \textbf{VLM class:} "RELATED"\\
  \textbf{Ensemble decision:} \textsc{Unrelated-Real}.\\
  \textbf{Released output:} "I cannot answer based on the given image."
  \end{minipage}
  \end{minipage}}

  \vspace{0.5em}

  \fbox{%
  \begin{minipage}{0.96\linewidth}
  \textbf{Example C: Blank/noise input} \hfill \textbf{Model: Gemma3-4B}\\[0.5em]
  \begin{minipage}{0.23\linewidth}
  \centering
  \fbox{%
  \includegraphics[width=0.95\linewidth]{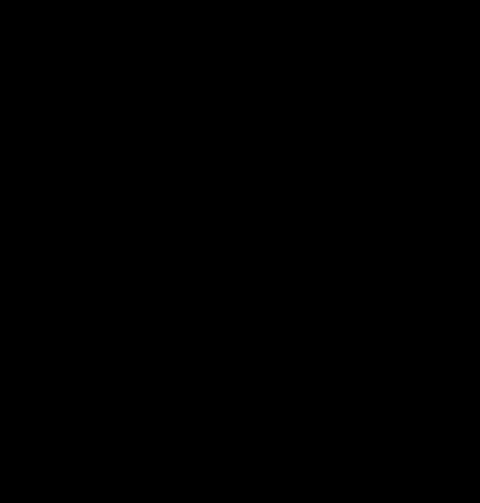}
  }
  \end{minipage}
  \hfill
  \begin{minipage}{0.72\linewidth}
  \textbf{Question:} "is there airspace consolidation on the left side?"\\
  \textbf{Raw VLM output:} "Yes, there appears to be airspace consolidation on the left side of the image."\\
  \textbf{Stage 1:} \\
  image\_std=0.000\\
  spectral\_flatness=0.000\\
  blank flag = blank\\
  \textbf{Ensemble decision:} \textsc{Blank/Noise}.\\
  \textbf{Released output:} "I cannot answer based on the given image."
  \end{minipage}
  \end{minipage}}

  \caption{Qualitative result cards for all three input conditions (Related, Unrelated-Real, Blank/Noise)
  on the same question using Gemma3-4B. On the related CT image the ensemble over-refuses a
  confident answer (the residual over-refusal cost quantified in Appendix~\ref{app:confusion}). On the
  unrelated real image the VLM hallucinates a confident finding and its own self-assessment is fooled,
  but the TC-LIA trajectory features drop and the ensemble rejects the response. The blank input is
  deterministically blocked by the pixel-statistics gate.}
  \label{fig:result_gemma}
  \end{figure*}

\begin{figure*}[t]
\centering

\fbox{%
\begin{minipage}{0.96\linewidth}
\textbf{Example A: Medical Question / Related Image} \hfill \textbf{Model: LLaVA-NeXT-110B}\\[0.5em]
\begin{minipage}{0.23\linewidth}
\centering
\fbox{%
\includegraphics[width=0.95\linewidth]{figures/result_card/images_chest_vqa_real_167.jpg}
}
\end{minipage}
\hfill
\begin{minipage}{0.72\linewidth}
\textbf{Question:} "how is the patient positioned for this scan?"\\
\textbf{Raw VLM output:} "The patient is positioned lying down on their back."\\
\textbf{TC-LIA:}\\
final\_cos=0.25632190704345703\\
late=0.2388894772157073\\
gain=0.14442294416949153\\
slope=0.008294089832493374\\
IAS=0.2103761734343284\\
\textbf{VLM class:} "RELATED"\\
\textbf{Ensemble decision:} \textsc{RELATED}.\\
\textbf{Released output:} "The patient is positioned lying down on their back."
\end{minipage}
\end{minipage}}

\vspace{0.5em}

\fbox{%
\begin{minipage}{0.96\linewidth}
\textbf{Example B: Medical Question / Unrelated-Real Image} \hfill \textbf{Model: LLaVA-NeXT-110B}\\[0.5em]
\begin{minipage}{0.23\linewidth}
\centering
\fbox{%
\includegraphics[width=0.95\linewidth]{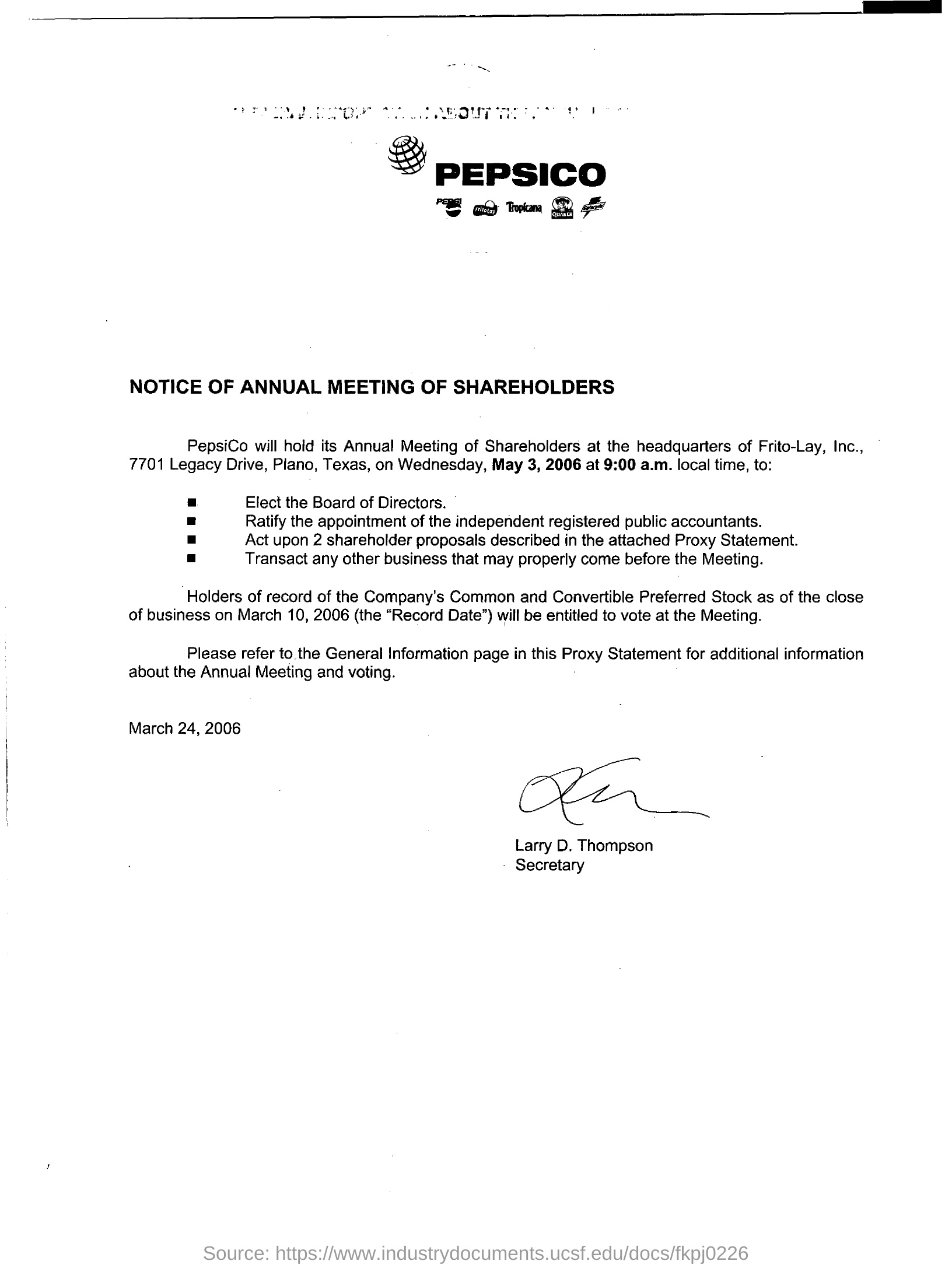}
}
\end{minipage}
\hfill
\begin{minipage}{0.72\linewidth}
\textbf{Question:} "how is the patient positioned for this scan?"\\
\textbf{Raw VLM output:} "The patient is positioned lying down on their back."\\
\textbf{TC-LIA:}\\
final\_cos=0.08261966705322266\\
late=0.12089351005852222\\
gain=0.02212412329390645\\
slope=0.0014618255643179523\\
IAS=0.07499801209175964\\
\textbf{VLM class:} "RELATED"\\
\textbf{Ensemble decision:} \textsc{Unrelated\_Real}.\\
\textbf{Released output:} "I cannot answer based on the given image."
\end{minipage}
\end{minipage}}

\vspace{0.5em}

\fbox{%
\begin{minipage}{0.96\linewidth}
\textbf{Example C: Medical Question / Blank/noise input} \hfill \textbf{Model: LLaVA-NeXT-110B}\\[0.5em]
\begin{minipage}{0.23\linewidth}
\centering
\fbox{%
\includegraphics[width=0.95\linewidth]{figures/blank.jpg}
}
\end{minipage}
\hfill
\begin{minipage}{0.72\linewidth}
\textbf{Question:} "how is the patient positioned for this scan?"\\
\textbf{Raw VLM output:} "The patient is lying down on their back."\\
\textbf{Stage 1:} \\
image\_std=0.0\\
spectral\_flatness=0.0\\
blank flag = blank\\
\textbf{Ensemble decision:} \textsc{Blank/Noise}.\\
\textbf{Released output:} "I cannot answer based on the given image."
\end{minipage}
\end{minipage}}

\caption{Qualitative result cards for all three input conditions (Related, Unrelated-Real,
Blank/Noise) on the same question using LLaVA-NeXT-110B.}
\label{fig:result_cards_llava_next}
\end{figure*}

\begin{figure*}[t]
    \centering

    \fbox{%
    \begin{minipage}{0.96\linewidth}
    \textbf{Example A: Medical Question / Related Image} \hfill \textbf{Model: Aya-Vision-32B}\\[0.5em]
    \begin{minipage}{0.23\linewidth}
    \centering
    \fbox{\includegraphics[width=0.95\linewidth]{figures/result_card/images_chest_vqa_real_17.jpg}}
    \end{minipage}
    \hfill
    \begin{minipage}{0.72\linewidth}
    \textbf{Question:} ``what are these opacities anterior to the right kidney?''\\
    \textbf{Raw VLM output:} ``Based on the image provided, the opacities anterior to the right kidney are likely to be lymph nodes. These structures
  are typically seen as well-defined, round or oval shapes with varying densities, and they can be located along the renal hilus or in the perirenal
  space. [\ldots]''\\
    \textbf{TC-LIA:}
    final\_cos=0.3303,\;
    late=0.2134,\;
    gain=0.1199,\;
    slope=0.0072,\;
    IAS=0.2372\\
    \textbf{VLM class:} \textsc{Related}\\
    \textbf{Ensemble decision:} \textsc{Related} \checkmark\\
    \textbf{Released output:} ``[\ldots] the opacities anterior to the right kidney are likely to be lymph nodes. [\ldots]''
    \end{minipage}
    \end{minipage}}

    \vspace{0.5em}

    \fbox{%
    \begin{minipage}{0.96\linewidth}
    \textbf{Example B: Medical Question / Unrelated-Real Image} \hfill \textbf{Model: Aya-Vision-32B}\\[0.5em]
    \begin{minipage}{0.23\linewidth}
    \centering
    \fbox{\includegraphics[width=0.95\linewidth]{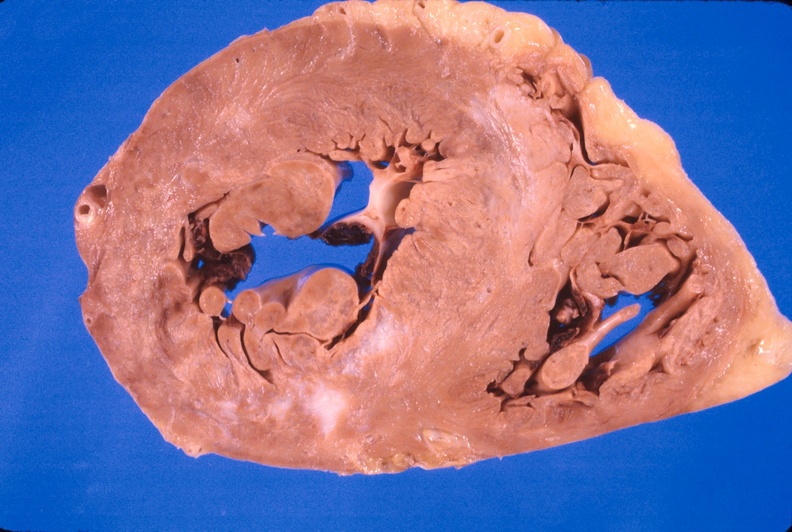}}
    \end{minipage}
    \hfill
    \begin{minipage}{0.72\linewidth}
    \textbf{Question:} ``what are these opacities anterior to the right kidney?''\\
    \textbf{Raw VLM output:} ``Based on the image provided, the opacities anterior to the right kidney appear to be calcifications. These are likely
  calcium deposits that have formed within the kidney tissue or along the kidney's surface. [\ldots]'' \textit{(base-prompt mirage)}\\
    \textbf{TC-LIA:}
    final\_cos=0.1880,\;
    late=0.1838,\;
    gain=0.1021,\;
    slope=0.0064,\;
    IAS=0.1559\\
    \textbf{VLM class:} \textsc{Related} \textit{(incorrect)}\\
    \textbf{Ensemble decision:} \textsc{Unrelated-Real} \checkmark\\
    \textbf{Released output:} ``I cannot answer based on the given image.''
    \end{minipage}
    \end{minipage}}

    \vspace{0.5em}

    \fbox{%
    \begin{minipage}{0.96\linewidth}
    \textbf{Example C: Medical Question / Blank Image} \hfill \textbf{Model: Aya-Vision-32B}\\[0.5em]
    \begin{minipage}{0.23\linewidth}
    \centering
    \fbox{\includegraphics[width=0.95\linewidth]{figures/blank.jpg}}
    \end{minipage}
    \hfill
    \begin{minipage}{0.72\linewidth}
    \textbf{Question:} ``what are these opacities anterior to the right kidney?''\\
    \textbf{Raw VLM output:} ``Based on the image provided, the opacities anterior to the right kidney are not clearly identifiable. [\ldots] They
  could potentially represent a variety of structures or conditions, such as bowel gas, soft tissue densities, or even artifacts from the imaging
  process. [\ldots]'' \textit{(base-prompt mirage)}\\
    \textbf{TC-LIA:} \textit{Stage-1 blank gate triggered} \\
    image\_std=0.0\\
spectral\_flatness=0.0\\
blank flag = blank\\\\
    \textbf{VLM class:} \textsc{Related} \textit{(incorrect)}\\
    \textbf{Ensemble decision:} \textsc{Blank/Noise} \checkmark\\
    \textbf{Released output:} ``I cannot answer based on the given image.''
    \end{minipage}
    \end{minipage}}

    \caption{Qualitative result cards for all three input conditions (\textsc{Related},
    \textsc{Unrelated-Real}, \textsc{Blank/Noise}) on the same medical question
    (Aya-Vision-32B, \texttt{chest\_vqa}, sample~17).
    The base-prompt VLM generates confident diagnoses (lymph nodes, calcifications)
    even for unrelated and blank inputs; the ensemble correctly identifies all three conditions.}
    \label{fig:result_cards_aya32b}
  \end{figure*}

\begin{figure*}[t]
    \centering

    \fbox{%
    \begin{minipage}{0.96\linewidth}
    \textbf{Example A: Medical Question / Related Image} \hfill \textbf{Model: BLIP2-OPT-2.7B}\\[0.5em]
    \begin{minipage}{0.23\linewidth}
    \centering
    \fbox{\includegraphics[width=0.95\linewidth]{figures/result_card/images_chest_vqa_real_221.jpg}}
    \end{minipage}
    \hfill
    \begin{minipage}{0.72\linewidth}
    \textbf{Question:} ``what is the condition?''\\
    \textbf{Raw VLM output:} ``a tumor''\\
    \textbf{TC-LIA:}
    final\_cos=0.2446,\;
    late=0.1809,\;
    gain=0.0357,\;
    slope=0.0020,\;
    IAS=0.1731\\
    \textbf{VLM class:} \textsc{Related}\\
    \textbf{Ensemble decision:} \textsc{Related} \checkmark\\
    \textbf{Released output:} ``a tumor''
    \end{minipage}
    \end{minipage}}

    \vspace{0.5em}

    \fbox{%
    \begin{minipage}{0.96\linewidth}
    \textbf{Example B: Medical Question / Unrelated-Real Image} \hfill \textbf{Model: BLIP2-OPT-2.7B}\\[0.5em]
    \begin{minipage}{0.23\linewidth}
    \centering
    \fbox{\includegraphics[width=0.95\linewidth]{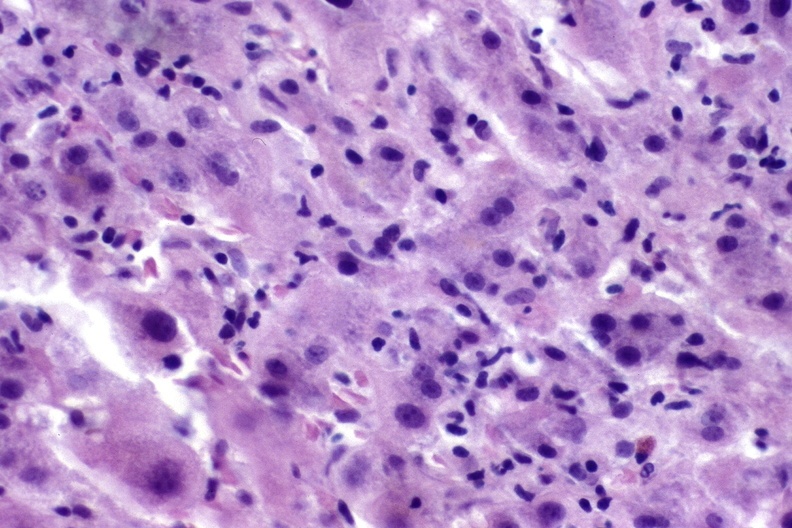}}
    \end{minipage}
    \hfill
    \begin{minipage}{0.72\linewidth}
    \textbf{Question:} ``what is the condition?''\\
    \textbf{Raw VLM output:} ``cancer'' \textit{(base-prompt mirage)}\\
    \textbf{TC-LIA:}
    final\_cos=0.2368,\;
    late=0.1747,\;
    gain=0.0448,\;
    slope=0.0024,\;
    IAS=0.1690\\
    \textbf{VLM class:} \textsc{Related} \textit{(incorrect)}\\
    \textbf{Ensemble decision:} \textsc{Unrelated-Real} \checkmark\\
    \textbf{Released output:} ``I cannot answer based on the given image.''
    \end{minipage}
    \end{minipage}}

    \vspace{0.5em}

    \fbox{%
    \begin{minipage}{0.96\linewidth}
    \textbf{Example C: Medical Question / Blank Image} \hfill \textbf{Model: BLIP2-OPT-2.7B}\\[0.5em]
    \begin{minipage}{0.23\linewidth}
    \centering
    \fbox{\includegraphics[width=0.95\linewidth]{figures/blank.jpg}}
    \end{minipage}
    \hfill
    \begin{minipage}{0.72\linewidth}
    \textbf{Question:} ``what is the condition?''\\
    \textbf{Raw VLM output:} ``the condition is the condition'' \textit{(base-prompt mirage)}\\
    \textbf{TC-LIA:} \textit{Stage-1 blank gate triggered} \\
        image\_std=0.0\\
spectral\_flatness=0.0\\
blank flag = blank\\\\
    \textbf{VLM class:} \textsc{Related} \textit{(incorrect)}\\
    \textbf{Ensemble decision:} \textsc{Blank/Noise} \checkmark\\
    \textbf{Released output:} ``I cannot answer based on the given image.''
    \end{minipage}
    \end{minipage}}

    \caption{Qualitative result cards for all three input conditions (\textsc{Related},
    \textsc{Unrelated-Real}, \textsc{Blank/Noise}) on the same medical question
    (BLIP2-OPT-2.7B, \texttt{chest\_vqa}, sample~221).
    The base-prompt VLM produces confident but wrong diagnoses for non-related inputs
    (``cancer'', ``the condition is the condition'');
    the ensemble correctly identifies all three conditions.}
    \label{fig:result_cards_blip2}
  \end{figure*}

  The main paper (Fig.~\ref{fig:result_cards_main}) shows result cards for GPT-5.4 on
matched and mismatched inputs with the same medical question;
Figures~\ref{fig:result_gemma} and~\ref{fig:result_cards_internvl} show result cards
for Gemma3-4B and InternVL2-8B respectively, each covering the \textsc{Related},
\textsc{Unrelated-Real}, and \textsc{Blank/Noise} conditions on the same question.
Additional result cards for LLaVA-NeXT-110B (Fig.~\ref{fig:result_cards_llava_next}) and for
Qwen2.5-VL-32B, Aya-Vision-32B, BLIP2-OPT-2.7B, LLaVA-1.5-7B, and InternVL3-38B are shown in
Figures~\ref{fig:result_cards_qwen32b}--\ref{fig:result_cards_internvl3_38b}.
These cards are illustrative demonstrations. Qwen2.5-VL-32B and BLIP2-OPT-2.7B belong to the
evaluated 14-backbone roster; Aya-Vision-32B, LLaVA-NeXT-110B, LLaVA-1.5-7B, and
InternVL3-38B do not, and no quantitative claim in this paper is based on them.



\begin{figure*}[t]
  \centering

  \fbox{%
  \begin{minipage}{0.96\linewidth}
  \textbf{Example A: Medical Question / Related Image} \hfill \textbf{Model: InternVL2-8B}\\[0.5em]
  \begin{minipage}{0.23\linewidth}
  \centering
  \fbox{%
  \includegraphics[width=0.95\linewidth]{figures/chest_3_r.jpg}
  }
  \end{minipage}
  \hfill
  \begin{minipage}{0.72\linewidth}
  \textbf{Question:} "which side of the heart border is obscured?"\\
  \textbf{Raw VLM output:} "The left border of the heart is obscured in the image."\\
  \textbf{TC-LIA:}\\
  final\_cos=0.21062666177749634\\
  late=0.17084469366818666\\
  gain=0.03115342278033495\\
  slope=0.001489122176974401\\
  IAS=0.1528464299405425\\
  \textbf{VLM class:} "RELATED"\\
  \textbf{Ensemble decision:} \textsc{Related}.\\
  \textbf{Released output:} "The left border of the heart is obscured in the image."
  \end{minipage}
  \end{minipage}}

  \vspace{0.5em}

  \fbox{%
  \begin{minipage}{0.96\linewidth}
  \textbf{Example B: Medical Question / Unrelated-Real Image} \hfill \textbf{Model: InternVL2-8B}\\[0.5em]
  \begin{minipage}{0.23\linewidth}
  \centering
  \fbox{%
  \includegraphics[width=0.95\linewidth]{figures/chest_3_u.jpg}
  }
  \end{minipage}
  \hfill
  \begin{minipage}{0.72\linewidth}
  \textbf{Question:} "which side of the heart border is obscured?"\\
  \textbf{Raw VLM output:} "The left side of the heart border is obscured."\\
  \textbf{TC-LIA:}\\
  final\_cos=0.07188746333122253\\
  late=0.15706725232303143\\
  gain=0.0027110567316412926\\
  slope= 0.0002593882955457832\\
  IAS=0.0756431420856699\\
  \textbf{VLM class:} "UNRELATED\_REAL"\\
  \textbf{Ensemble decision:} \textsc{Unrelated\_Real}.\\
  \textbf{Released output:}  "I cannot answer based on the given image."
  \end{minipage}
  \end{minipage}}

  \vspace{0.5em}

  \fbox{%
  \begin{minipage}{0.96\linewidth}
  \textbf{Example C: Blank/noise input} \hfill \textbf{Model: InternVL2-8B}\\[0.5em]
  \begin{minipage}{0.23\linewidth}
  \centering
  \fbox{%
  \includegraphics[width=0.95\linewidth]{figures/blank.jpg}
  }
  \end{minipage}
  \hfill
  \begin{minipage}{0.72\linewidth}
  \textbf{Question:} "which side of the heart border is obscured?"\\
  \textbf{Raw VLM output:} "The left border of the heart is obscured."\\
  \textbf{Stage 1:} image\_std=0\\
   spectral\_flatness=0\\
  blank flag = blank\\
  \textbf{Ensemble decision:} \textsc{Blank/Noise}.\\
  \textbf{Released output:} "I cannot answer based on the given image."
  \end{minipage}
  \end{minipage}}

  \caption{Qualitative result cards for all three input conditions (Related, Unrelated-Real, Blank/Noise)
  on the same question using InternVL2-8B.}
  \label{fig:result_cards_internvl}
  \end{figure*}

  \begin{figure*}[t]
    \centering

    \fbox{%
    \begin{minipage}{0.96\linewidth}
    \textbf{Example A: Medical Question / Related Image} \hfill \textbf{Model: Qwen2.5-VL-32B}\\[0.5em]
    \begin{minipage}{0.23\linewidth}
    \centering
    \fbox{\includegraphics[width=0.95\linewidth]{figures/result_card/images_chest_vqa_real_153.jpg}}
    \end{minipage}
    \hfill
    \begin{minipage}{0.72\linewidth}
    \textbf{Question:} ``what is the condition in which the ventricles are enlarged as seen in this image?''\\
    \textbf{Raw VLM output:} ``Hydrocephalus''\\
    \textbf{TC-LIA:}
    final\_cos=0.3814,\;
    late=0.2409,\;
    gain=0.1422,\;
    slope=0.0085,\;
    IAS=0.2731\\
    \textbf{VLM class:} \textsc{Related}\\
    \textbf{Ensemble decision:} \textsc{Related} \checkmark\\
    \textbf{Released output:} ``Hydrocephalus''
    \end{minipage}
    \end{minipage}}

    \vspace{0.5em}

    \fbox{%
    \begin{minipage}{0.96\linewidth}
    \textbf{Example B: Medical Question / Unrelated-Real Image} \hfill \textbf{Model: Qwen2.5-VL-32B}\\[0.5em]
    \begin{minipage}{0.23\linewidth}
    \centering
    \fbox{\includegraphics[width=0.95\linewidth]{figures/result_card/images_chest_vqa_unrelated_153.jpg}}
    \end{minipage}
    \hfill
    \begin{minipage}{0.72\linewidth}
    \textbf{Question:} ``what is the condition in which the ventricles are enlarged as seen in this image?''\\
    \textbf{Raw VLM output:} ``Hydrocephalus'' \textit{(base-prompt mirage)}\\
    \textbf{TC-LIA:}
    final\_cos=0.1929,\;
    late=0.2108,\;
    gain=0.1015,\;
    slope=0.0060,\;
    IAS=0.1650\\
    \textbf{VLM class:} \textsc{Related} \textit{(incorrect)}\\
    \textbf{Ensemble decision:} \textsc{Unrelated-Real} \checkmark\\
    \textbf{Released output:} ``I cannot answer based on the given image.''
    \end{minipage}
    \end{minipage}}

    \vspace{0.5em}

    \fbox{%
    \begin{minipage}{0.96\linewidth}
    \textbf{Example C: Medical Question / Blank Image} \hfill \textbf{Model: Qwen2.5-VL-32B}\\[0.5em]
    \begin{minipage}{0.23\linewidth}
    \centering
    \fbox{\includegraphics[width=0.95\linewidth]{figures/blank.jpg}}
    \end{minipage}
    \hfill
    \begin{minipage}{0.72\linewidth}
    \textbf{Question:} ``what is the condition in which the ventricles are enlarged as seen in this image?''\\
    \textbf{Raw VLM output:} ``Dilated cardiomyopathy'' \textit{(base-prompt mirage)}\\
    \textbf{TC-LIA:} \textit{Stage-1 blank gate triggered} \\ image\_std=0.0\\
spectral\_flatness=0.0\\
blank flag = blank\\\\
    \textbf{VLM class:} \textsc{Blank\_or\_Noise}\\
    \textbf{Ensemble decision:} \textsc{Blank/Noise} \checkmark\\
    \textbf{Released output:} ``I cannot answer based on the given image.''
    \end{minipage}
    \end{minipage}}

    \caption{Qualitative result cards for all three input conditions (\textsc{Related},
    \textsc{Unrelated-Real}, \textsc{Blank/Noise}) on the same medical question
    (Qwen2.5-VL-32B, \texttt{chest\_vqa}, sample~153).
    The base-prompt VLM hallucinates a diagnosis for both non-related inputs;
    the ensemble correctly identifies all three conditions.}
    \label{fig:result_cards_qwen32b}
  \end{figure*}



  \begin{figure*}[t]
    \centering

    \fbox{%
    \begin{minipage}{0.96\linewidth}
    \textbf{Example A: Medical Question / Related Image} \hfill \textbf{Model: LLaVA-1.5-7B}\\[0.5em]
    \begin{minipage}{0.23\linewidth}
    \centering
    \fbox{\includegraphics[width=0.95\linewidth]{figures/result_card/images_chest_vqa_real_316.jpg}}
    \end{minipage}
    \hfill
    \begin{minipage}{0.72\linewidth}
    \textbf{Question:} ``the hypointense nature of the hemorrhage signifies what in terms of the timeline?''\\
    \textbf{Raw VLM output:} ``The hypointense nature of the hemorrhage signifies that it is an old hemorrhage.''\\
    \textbf{TC-LIA:}
    final\_cos=0.2859,\;
    late=0.1899,\;
    gain=0.0868,\;
    slope=0.0051,\;
    IAS=0.2040\\
    \textbf{VLM class:} \textsc{Related}\\
    \textbf{Ensemble decision:} \textsc{Related} \checkmark\\
    \textbf{Released output:} ``The hypointense nature of the hemorrhage signifies that it is an old hemorrhage.''
    \end{minipage}
    \end{minipage}}

    \vspace{0.5em}

    \fbox{%
    \begin{minipage}{0.96\linewidth}
    \textbf{Example B: Medical Question / Unrelated-Real Image} \hfill \textbf{Model: LLaVA-1.5-7B}\\[0.5em]
    \begin{minipage}{0.23\linewidth}
    \centering
    \fbox{\includegraphics[width=0.95\linewidth]{figures/result_card/images_chest_vqa_unrelated_316.jpg}}
    \end{minipage}
    \hfill
    \begin{minipage}{0.72\linewidth}
    \textbf{Question:} ``the hypointense nature of the hemorrhage signifies what in terms of the timeline?''\\
    \textbf{Raw VLM output:} ``The hypointense nature of the hemorrhage signifies that it is a very early stage of the bleeding.''
  \textit{(base-prompt mirage)}\\
    \textbf{TC-LIA:}
    final\_cos=0.0649,\;
    late=0.1344,\;
    gain=0.0247,\;
    slope=0.0012,\;
    IAS=0.0699\\
    \textbf{VLM class:} \textsc{Related} \textit{(incorrect)}\\
    \textbf{Ensemble decision:} \textsc{Unrelated-Real} \checkmark\\
    \textbf{Released output:} ``I cannot answer based on the given image.''
    \end{minipage}
    \end{minipage}}

    \vspace{0.5em}

    \fbox{%
    \begin{minipage}{0.96\linewidth}
    \textbf{Example C: Medical Question / Blank Image} \hfill \textbf{Model: LLaVA-1.5-7B}\\[0.5em]
    \begin{minipage}{0.23\linewidth}
    \centering
    \fbox{\includegraphics[width=0.95\linewidth]{figures/blank.jpg}}
    \end{minipage}
    \hfill
    \begin{minipage}{0.72\linewidth}
    \textbf{Question:} ``the hypointense nature of the hemorrhage signifies what in terms of the timeline?''\\
    \textbf{Raw VLM output:} ``The hypointense nature of the hemorrhage signifies that it is an old hemorrhage.'' \textit{(base-prompt mirage)}\\
    \textbf{TC-LIA:} \textit{Stage-1 blank gate triggered} \\
            image\_std=0.0\\
spectral\_flatness=0.0\\
blank flag = blank\\\\
    \textbf{VLM class:} \textsc{Related} \textit{(incorrect)}\\
    \textbf{Ensemble decision:} \textsc{Blank/Noise} \checkmark\\
    \textbf{Released output:} ``I cannot answer based on the given image.''
    \end{minipage}
    \end{minipage}}

    \caption{Qualitative result cards for all three input conditions (\textsc{Related},
    \textsc{Unrelated-Real}, \textsc{Blank/Noise}) on the same medical question
    (LLaVA-1.5-7B, \texttt{chest\_vqa}, sample~316).
    The base-prompt VLM hallucinates timeline-specific diagnoses for both
    non-related inputs; the IAS drop from 0.204 (related) to 0.070 (unrelated)
    illustrates the discriminative power of TC-LIA.}
    \label{fig:result_cards_llava15_7b}
  \end{figure*}


  \begin{figure*}[t]
    \centering

    \fbox{%
    \begin{minipage}{0.96\linewidth}
    \textbf{Example A: Medical Question / Related Image} \hfill \textbf{Model: InternVL3-38B}\\[0.5em]
    \begin{minipage}{0.23\linewidth}
    \centering
    \fbox{\includegraphics[width=0.95\linewidth]{figures/result_card/images_chest_vqa_real_81.jpg}}
    \end{minipage}
    \hfill
    \begin{minipage}{0.72\linewidth}
    \textbf{Question:} ``from what structure is the large hypodense mass emanating from?''\\
    \textbf{Raw VLM output:} ``The large hypodense mass is emanating from the pituitary gland.''\\
    \textbf{TC-LIA:}
    final\_cos=0.3106,\;
    late=0.1985,\;
    gain=0.0864,\;
    slope=0.0051,\;
    IAS=0.2184\\
    \textbf{VLM class:} \textsc{Related}\\
    \textbf{Ensemble decision:} \textsc{Related} \checkmark\\
    \textbf{Released output:} ``The large hypodense mass is emanating from the pituitary gland.''
    \end{minipage}
    \end{minipage}}

    \vspace{0.5em}

    \fbox{%
    \begin{minipage}{0.96\linewidth}
    \textbf{Example B: Medical Question / Unrelated-Real Image} \hfill \textbf{Model: InternVL3-38B}\\[0.5em]
    \begin{minipage}{0.23\linewidth}
    \centering
    \fbox{\includegraphics[width=0.95\linewidth]{figures/result_card/images_chest_vqa_unrelated_81.jpg}}
    \end{minipage}
    \hfill
    \begin{minipage}{0.72\linewidth}
    \textbf{Question:} ``from what structure is the large hypodense mass emanating from?''\\
    \textbf{Raw VLM output:} ``The large hypodense mass is emanating from the pituitary gland.'' \textit{(base-prompt mirage)}\\
    \textbf{TC-LIA:}
    final\_cos=0.1384,\;
    late=0.1358,\;
    gain=0.0216,\;
    slope=0.0013,\;
    IAS=0.1065\\
    \textbf{VLM class:} \textsc{Related} \textit{(incorrect)}\\
    \textbf{Ensemble decision:} \textsc{Unrelated-Real} \checkmark\\
    \textbf{Released output:} ``I cannot answer based on the given image.''
    \end{minipage}
    \end{minipage}}

    \vspace{0.5em}

    \fbox{%
    \begin{minipage}{0.96\linewidth}
    \textbf{Example C: Medical Question / Blank Image} \hfill \textbf{Model: InternVL3-38B}\\[0.5em]
    \begin{minipage}{0.23\linewidth}
    \centering
    \fbox{\includegraphics[width=0.95\linewidth]{figures/blank.jpg}}
    \end{minipage}
    \hfill
    \begin{minipage}{0.72\linewidth}
    \textbf{Question:} ``from what structure is the large hypodense mass emanating from?''\\
    \textbf{Raw VLM output:} ``The large hypodense mass is emanating from the pituitary gland.'' \textit{(base-prompt mirage)}\\
    \textbf{TC-LIA:} \textit{Stage-1 blank gate triggered} \\
        image\_std=0.0\\
spectral\_flatness=0.0\\
blank flag = blank\\\\
    \textbf{VLM class:} \textsc{Related} \textit{(incorrect)}\\
    \textbf{Ensemble decision:} \textsc{Blank/Noise} \checkmark\\
    \textbf{Released output:} ``I cannot answer based on the given image.''
    \end{minipage}
    \end{minipage}}

    \caption{Qualitative result cards for all three input conditions (\textsc{Related},
    \textsc{Unrelated-Real}, \textsc{Blank/Noise}) on the same medical question
    (InternVL3-38B, \texttt{chest\_vqa}, sample~81).
    Despite having no relevant visual evidence, the base-prompt VLM confidently
    produces the identical answer ``pituitary gland'' for all three conditions;
    the ensemble correctly identifies each case.}
    \label{fig:result_cards_internvl3_38b}
  \end{figure*}





\section{Classifier-Level Diagnostics}
\label{app:classifier_diagnostics}

\subsection{Confusion Matrix Summary}
\label{app:confusion}

Table~\ref{tab:confusion_summary} summarises the per-class error distribution across
all 14 VLM backbones.
\textsc{Blank} inputs are perfectly separated in every case (zero misclassifications),
confirming that blank/noise detection is effectively deterministic and is not the
bottleneck.
All residual errors are concentrated between \textsc{Related} and
\textsc{Unrelated}. The last two columns report raw error \emph{counts}, and the
class supports are unequal: the two unrelated conditions together contain roughly
twice as many test samples as the related condition, so equal counts would already
imply a higher per-sample error rate on related inputs. Normalising by class
support, the asymmetry is clear for every backbone: the dominant error direction
is \textsc{Related}$\to$\textsc{Unrelated} (i.e.\ the ensemble over-refuses, routing
genuinely answerable inputs to refusal), while \textsc{Unrelated}$\to$\textsc{Related}
(mirage errors) have a lower per-sample rate but are more consequential, since they
correspond to unsupported answers being accepted.
Qwen2.5-VL-72B achieves the best balance, with the highest \textsc{Related} recall
($80.7\%$) and strong \textsc{Unrelated} recall ($86.5\%$), while weaker models
(BLIP2-OPT-2.7B, Phi-3.5-Vision, Gemma3-4B) show the lowest \textsc{Related} recall
($63.5\%$--$64.6\%$), indicating more over-refusal on genuinely answerable inputs.
\textsc{Unrelated} recall is notably consistent across all backbones
($84.8$--$87.8\%$), showing that mirage suppression is robust regardless of
the VLM backbone.

\paragraph{Blank-gate accounting.}
The three-class arithmetic above treats the Stage-1 blank/noise gate as part of the
predictor: records flagged by the gate (all blank-condition inputs, plus the small
set of non-blank records whose pixel statistics trigger the blank/noise criterion)
are assigned \textsc{Blank} deterministically and are never scored by the ensemble
classifier. These flows are image-side quantities, so they are identical for every
backbone. On the 15{,}004-record test split they are: all 3{,}751 \textsc{Blank}
inputs are flagged and refused (100\% blank recall, zero \textsc{Blank}
misclassifications; hence blank-condition cells are omitted from
Table~\ref{tab:confusion_summary}); 35 \textsc{Related} and 57 \textsc{Unrelated}
test records are additionally flagged (0.9\% of related and 0.8\% of unrelated
support). The 35 constitute over-refusals and are charged against \textsc{Related}
recall; the 57 are refusals on genuinely out-of-scope inputs (the desired
behavior), but the strict per-class accounting charges them to \textsc{Unrelated}
recall, which is why the two recall columns fall slightly below the
Related$\to$Unrelated/Unrelated$\to$Related counts alone would suggest. Neither
flow can produce a mirage error (a refused answer is never accepted), and the
mirage-rate arithmetic of Table~\ref{tab:main} counts only
\textsc{Unrelated}$\to$\textsc{Related} cells. With these two gate columns
included, the confusion arithmetic closes exactly for every backbone; for example,
Qwen2.5-VL-72B: \textsc{Related} $3{,}026 + 690 + 35 = 3{,}751$,
\textsc{Unrelated} $6{,}489 + 956 + 57 = 7{,}502$,
\textsc{Blank} $3{,}751 + 0 + 0 = 3{,}751$.

\begin{table*}[h]
\centering
\small
\caption{Per-class error summary for the best ensemble per VLM backbone.
\textsc{Blank} inputs are perfectly classified by all models (the Stage-1 gate flags
all of them) and are omitted; the gate additionally hard-refuses 92 non-blank test
records (35 \textsc{Related}, 57 \textsc{Unrelated}) that are charged to the recall
denominators and are identical for every backbone (see the accounting paragraph
below).
Rel~Rec = \textsc{Related} recall; Unrel~Rec = \textsc{Unrelated} recall.
Rel$\to$Unrel = related inputs incorrectly refused;
Unrel$\to$Rel = unrelated inputs incorrectly accepted (mirage errors).
All rows are computed with the released training-pipeline configuration
(identical features, split, and hyperparameters); the InternVL2-8B row is
recomputed over all 10 domains.}
\label{tab:confusion_summary}
\setlength{\tabcolsep}{5pt}
\renewcommand{\arraystretch}{0.82}
\begin{tabular*}{\textwidth}{@{\extracolsep{\fill}} l c cc rr}
\toprule
Model & Clf & Rel Rec$\uparrow$ & Unrel Rec$\uparrow$ & Rel$\to$Unrel$\downarrow$ & Unrel$\to$Rel$\downarrow$ \\
\midrule
Qwen2.5-VL-72B     & RF  & \textbf{80.7\%} & 86.5\% &  690 &  956 \\
Claude-Opus-4.8      & Ada & 79.4\%          & 86.8\% &  738 &  936 \\
Qwen2.5-VL-7B      & RF  & 74.6\%          & \textbf{87.3\%} &  917 &  899 \\
Qwen2.5-VL-32B     & Ada & 77.7\%          & 84.8\% &  803 & 1081 \\
Kimi-K3            & RF  & 80.3\%          & 85.6\% &  704 & 1024 \\
GPT-5.4            & RF  & 73.2\%          & 85.6\% &  969 & 1020 \\
Llama-4-Scout      & RF  & 69.9\%          & 86.4\% & 1093 &  963 \\
Llama-3.2-11B      & RF  & 64.6\%          & \textbf{87.8\%} & 1292 &  861 \\
LLaVA-OV-Qwen2-7B  & Ada & 65.8\%          & 87.0\% & 1248 &  918 \\
Gemma3-4B          & RF  & 64.6\%          & 87.0\% & 1293 &  920 \\
Phi-3.5-Vision     & RF  & 63.8\%          & 87.4\% & 1324 &  889 \\
BLIP2-OPT-2.7B     & RF  & 63.5\%          & 87.2\% & 1334 &  905 \\
MiniCPM-V-2.6      & RF  & 68.8\%          & 86.8\% & 1136 &  935 \\
InternVL2-8B        & RF & 69.2\%          & 87.3\% & 1122 &  895 \\
\bottomrule
\end{tabular*}
\end{table*}

\subsection{Classifier Comparison}
\label{app:classifier_comparison}

Table~\ref{tab:appendix_classifier_comparison_all} reports held-out accuracy and
5-fold cross-validation accuracy for all five ensemble classifiers across fourteen
VLM backbones.
Two observations stand out.
First, the performance spread across classifiers is remarkably narrow for every
backbone ($0.29$--$0.86$ percentage points in held-out accuracy), confirming that
the ensemble's effectiveness is driven by the feature set rather than classifier
choice.
Second, Random Forest is the best classifier for eleven of fourteen
backbones and is used as the default throughout the paper; AdaBoost leads on the
remaining three (Claude-Opus-4.8, Qwen2.5-VL-32B, LLaVA-OV-Qwen2-7B).
Cross-validation accuracy tracks held-out accuracy closely across all backbones,
with a consistent gap of $2.3$--$3.8$ percentage points and low standard deviation
($0.76$--$2.07\%$), indicating stable generalisation rather than overfitting to the
held-out split.
The tightest CV variance is observed for BLIP2-OPT-2.7B ($\pm 0.76\%$), while the largest is for Qwen2.5-VL-7B and
Qwen2.5-VL-32B ($\pm 2.07\%$), likely reflecting greater sensitivity to fold
composition in higher-capability models.

\begin{table*}[h]
\centering
\small
\caption{Held-out accuracy (\%) and 5-fold CV accuracy (\%, mean$\pm$std) across
five ensemble classifiers and 14 VLM backbones.
\textbf{Bold} marks the best classifier per backbone.}
\label{tab:appendix_classifier_comparison_all}
\setlength{\tabcolsep}{4pt}
\renewcommand{\arraystretch}{0.82}
\begin{tabular*}{\textwidth}{@{\extracolsep{\fill}} l ccccc c}
\toprule
\multirow{2}{*}{Model} & \multicolumn{5}{c}{Held-out Accuracy (\%)} & \multirow{2}{*}{CV Acc (best clf)} \\
\cmidrule(lr){2-6}
 & RF & AdaBoost & GradBoost & XGBoost & LightGBM & \\
\midrule
Qwen2.5-VL-72B     & \textbf{88.42} & 88.08 & 88.06 & 88.08 & 88.00 & $85.67 \pm 1.84\%$ \\
Claude-Opus-4.8      & 88.18 & \textbf{88.23} & 88.04 & 87.99 & 87.88 & $85.30 \pm 1.96\%$ \\
Qwen2.5-VL-7B      & \textbf{87.28} & 87.03 & 86.90 & 86.81 & 86.83 & $84.22 \pm 2.07\%$ \\
Qwen2.5-VL-32B     & 86.82 & \textbf{86.83} & 86.37 & 86.50 & 86.25 & $84.32 \pm 2.07\%$ \\
GPT-5.4            & \textbf{86.13} & 85.88 & 85.81 & 85.44 & 85.50 & $83.14 \pm 1.75\%$ \\
Llama-4-Scout      & \textbf{85.68} & 85.13 & 85.26 & 84.86 & 84.93 & $82.70 \pm 1.77\%$ \\
Llama-3.2-11B      & \textbf{85.04} & 84.78 & 84.76 & 84.18 & 84.30 & $81.46 \pm 1.08\%$ \\
LLaVA-OV-Qwen2-7B  & 84.89 & \textbf{84.95} & 84.63 & 84.46 & 84.48 & $81.12 \pm 1.19\%$ \\
Gemma3-4B          & \textbf{84.64} & 84.54 & 84.42 & 84.03 & 83.84 & $81.66 \pm 1.17\%$ \\
Phi-3.5-Vision     & \textbf{84.64} & 84.38 & 84.35 & 84.00 & 83.98 & $81.39 \pm 1.01\%$ \\
BLIP2-OPT-2.7B     & \textbf{84.46} & 84.42 & 84.20 & 83.89 & 83.70 & $81.56 \pm 0.76\%$ \\
MiniCPM-V-2.6      & \textbf{85.58} & 85.18 & 85.32 & 84.89 & 85.01 & $82.47 \pm 1.70\%$ \\
InternVL2-8B       & \textbf{85.94} & 85.78 & 85.62 & 85.46 & 85.44 & $82.87 \pm 0.99\%$ \\
Kimi-K3            & \textbf{87.87} & 87.78 & 87.79 & 87.58 & 87.61 & $85.53 \pm 1.88\%$ \\
\bottomrule
\end{tabular*}
\end{table*}





\section{Full TC-LIA Score Comparison}

\begin{figure*}[h]
\centering
\includegraphics[width=\linewidth]{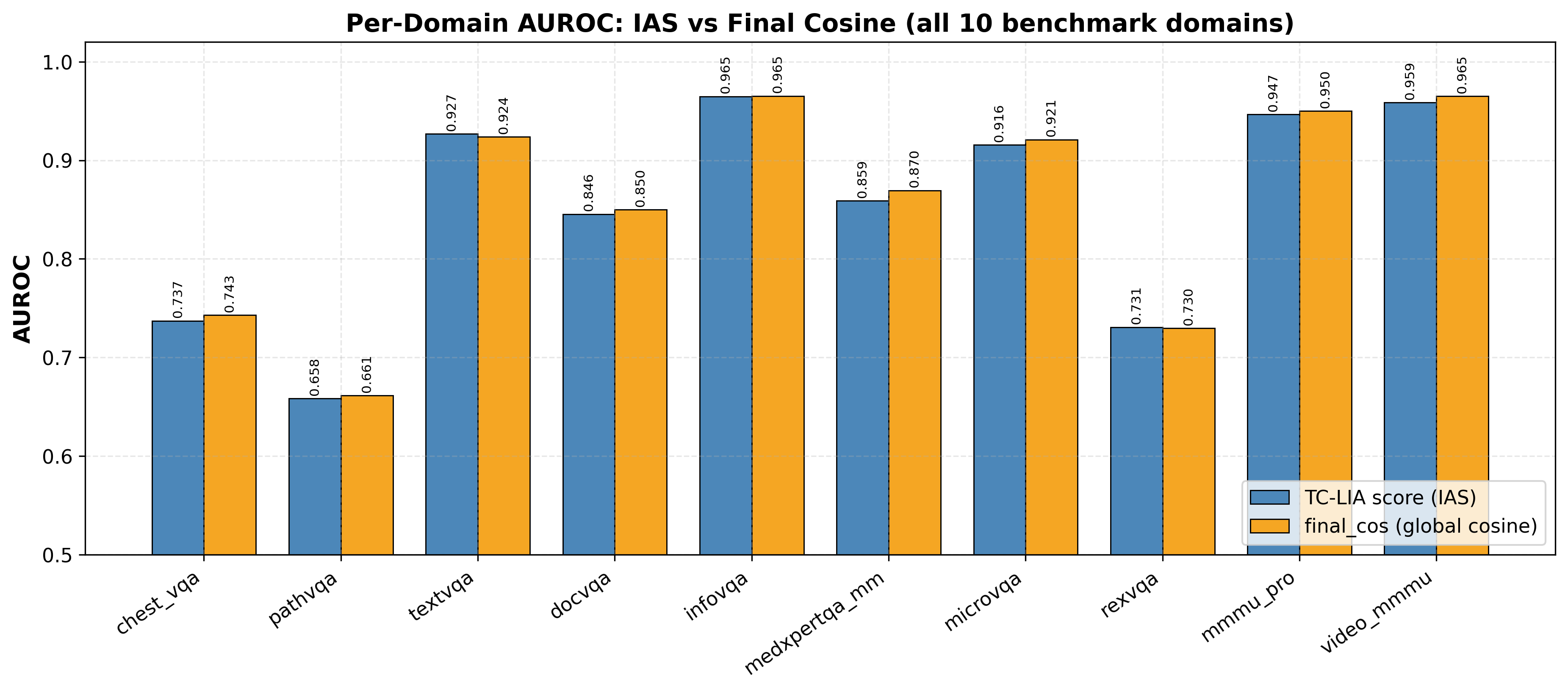}
\caption{
Per-domain AUROC for binary \textsc{Related} vs.\ \textsc{Unrelated-Real} detection
(in-domain and out-of-domain unrelated images pooled) using \texttt{final\_cos}
(ViT-H/14 global cosine) alone versus the Internal Alignment Score (IAS), on all ten
benchmark domains. The two scores are near-identical everywhere (largest absolute
difference $0.010$, on \texttt{medxpertqa\_mm}); the trajectory components that IAS
summarises contribute through the unsupervised threshold rule and the learned ensemble
rather than through a higher single-scalar AUROC.
}
\label{fig:domain_auroc}
\end{figure*}

Figure~\ref{fig:domain_auroc} compares IAS against \texttt{final\_cos} alone across all ten
benchmark domains (binary \textsc{Related} vs.\ pooled \textsc{Unrelated-Real} detection,
computed over the full 19{,}004-record benchmark with blank/noise records excluded). The two
scalars track each other closely: the per-domain AUROC difference never exceeds $0.011$ in
either direction ($|\Delta|\leq0.011$, mean $\Delta=-0.004$). IAS therefore preserves
essentially all of the global cosine's per-domain discriminative power; its added value is
not a higher single-scalar AUROC but the trajectory summary (late top-$k$, gain, slope) that
the unsupervised threshold rule (V09) and the learned ensemble exploit
(Table~\ref{tab:tclia_feature_auroc}).

\section{Per-Condition Answer Quality}
\label{app:quality_detail}

Table~\ref{tab:quality_detail} reports per-condition answer quality across all fourteen
VLM backbones. For \textsc{Related} inputs, BLEU and ROUGE-L occasionally decrease
under ensemble fusion, as the detector routes borderline related inputs to refusal
rather than rewriting answers: the ensemble is optimised for safe release, not
answer quality. BERTScore remains stable, confirming that accepted related answers are
semantically preserved. For \textsc{Unrelated} and \textsc{Blank} inputs the ensemble
achieves near-perfect refusal scores across all backbones, with \textsc{Blank}
detection being fully deterministic (1.000 across all models and metrics).

Averaged across all three conditions (Table~\ref{fig:quality} in the main text),
base-prompt BLEU and ROUGE-L are low because most non-related inputs receive
confident but incorrect answers rather than refusals; ensemble fusion produces a
large uniform improvement (${\sim}+0.6$ BLEU, ${\sim}+0.55$ ROUGE-L) by generating
well-formed refusals for \textsc{Unrelated} and \textsc{Blank} inputs, while
BERTScore is preserved or slightly improved on accepted \textsc{Related}
inputs---ensemble filtering does not degrade semantic quality. All ensemble
scores fall within a narrow band ($0.620$--$0.648$ BLEU, $0.659$--$0.753$
ROUGE-L), consistent across backbones.

\section{Risk--Coverage Trade-off of Answer Release}
\label{app:risk_coverage}

\begin{figure*}[h]
\centering
\includegraphics[width=\linewidth]{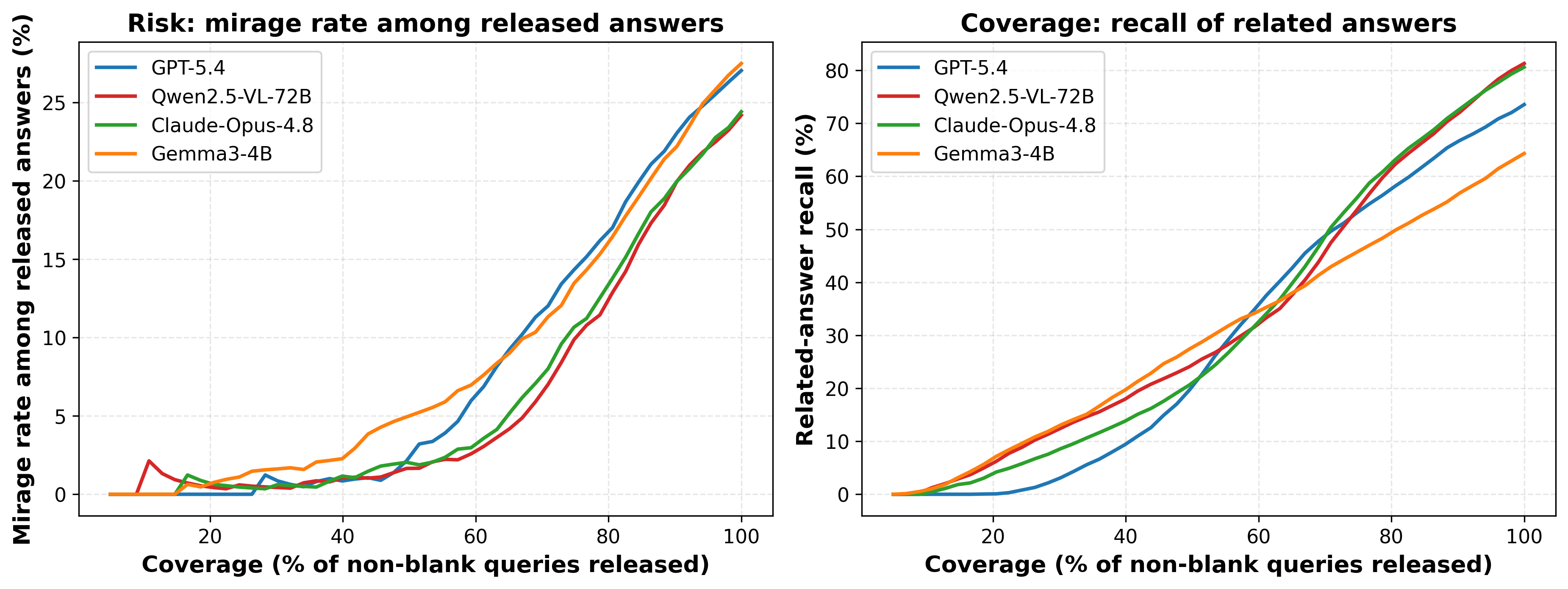}
\caption{Selective risk--coverage trade-off on the non-blank test set (released
training-pipeline configuration, harmonised feature snapshot). Left: mirage rate among
released answers. Right: recall of related answers. Releasing by ensemble confidence
suppresses mirages faster than coverage shrinks: withholding the least-confident 30\% of
queries removes more than half of the residual mirage risk.}
\label{fig:risk_coverage}
\end{figure*}

Because the ensemble doubles as a selective predictor (an answer is released only when
the classifier predicts \textsc{Related}), we report the full risk--coverage trade-off on
the non-blank test set. Queries are ranked by the ensemble's maximum class probability and
released in descending order of confidence; Figure~\ref{fig:risk_coverage} traces the
mirage rate among released answers and the recall of related answers as coverage grows,
for four backbones spanning the capability range. Under this recomputation, releasing
every non-blank answer yields a 24--28\% mirage rate among released answers; restricting
release to the 80\% highest-confidence queries lowers this to 13--17\%, and 70\% coverage
to 7--12\%, while related-answer recall falls correspondingly (from 64--81\% at full
coverage to 50--63\% at 80\%). The operating point is thus an explicit deployment choice:
every point on the curve trades mirage suppression against false abstention on answerable
inputs, and the confidence ordering concentrates mirage errors among the first answers to
be withheld. Curves are recomputed under the released training-pipeline configuration on
the harmonised feature snapshot (cf.\ the $^\ddagger$ rows of
Table~\ref{tab:tclia_feature_auroc}); absolute levels therefore differ from the published
per-backbone operating points in Table~\ref{tab:main}, while the within-curve trade-off is
the object of interest.

\begin{figure*}[t]
\centering
\includegraphics[width=0.7\textwidth]{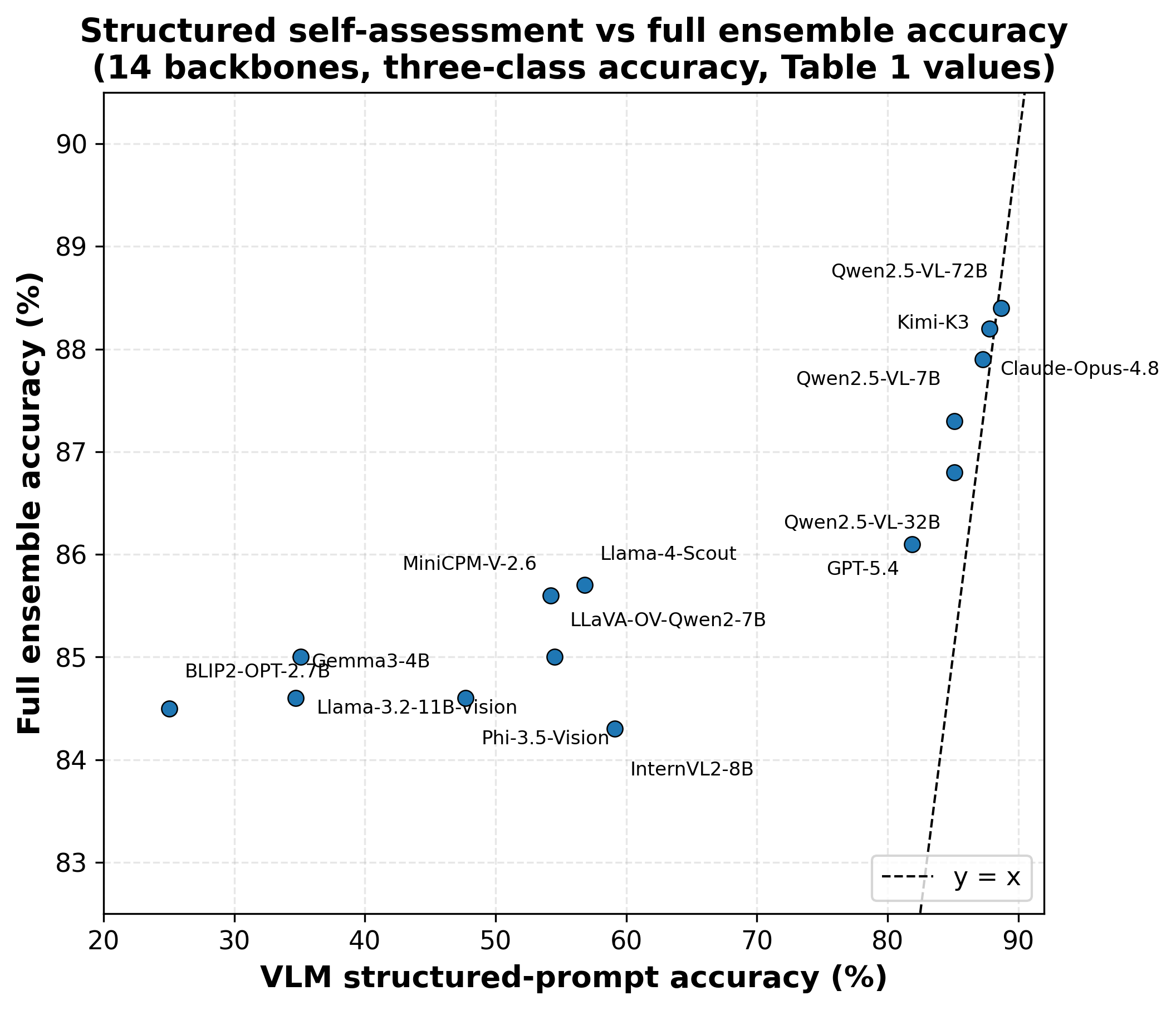}
\caption{Structured-prompt (self-assessment) accuracy vs.\ full ensemble accuracy for all
fourteen evaluated backbones, plotted from the values of Table~\ref{tab:main} (three-class
accuracy on the ten-domain benchmark). Points far above the diagonal are backbones whose
structured self-assessment is weak (e.g., BLIP2-OPT-2.7B at 25.0\%); the ensemble lifts all
of them into the 84.5--88.4\% band, while frontier models cluster near the diagonal.}
\label{fig:struc_vs_ens}
\end{figure*}

\section{Structured-Prompt Diagnostics}
\label{app:domain_structured_diagnostics}


Figure~\ref{fig:struc_vs_ens} compares the accuracy of the VLM's structured self-assessment alone against the full ensemble. Points above the diagonal indicate cases where TC-LIA, domain-routing, and pixel-statistic features improve over the VLM class prediction alone. The consistent gap demonstrates that the proposed detector is not merely relying on VLM self-refusal or self-classification; instead, the ensemble gains additional discriminative signal from layer-wise image--text alignment.

\section{Computational Overhead and Latency Analysis}
\label{app:latency}

We measure per-stage wall-clock latency (mean $\pm$ std) for the full pipeline to quantify the inference cost added on top of a base VLM call. All measurements are obtained on a single NVIDIA Tesla T4 GPU (15.6\,GB VRAM, driver 580.159.04) with 4 vCPU Intel Xeon 2.00\,GHz and 31\,GB RAM, using Qwen2.5-VL-7B-Instruct (NF4-quantised) as the reference VLM for the self-assessment stage; the frozen CLIP ViT-H/14 encoder runs on the same GPU. We time 10 images per condition (40 per pipeline) in the chest\_vqa domain after 3 warm-up runs per stage.

\begin{figure*}[h]
\centering
\includegraphics[width=\linewidth]{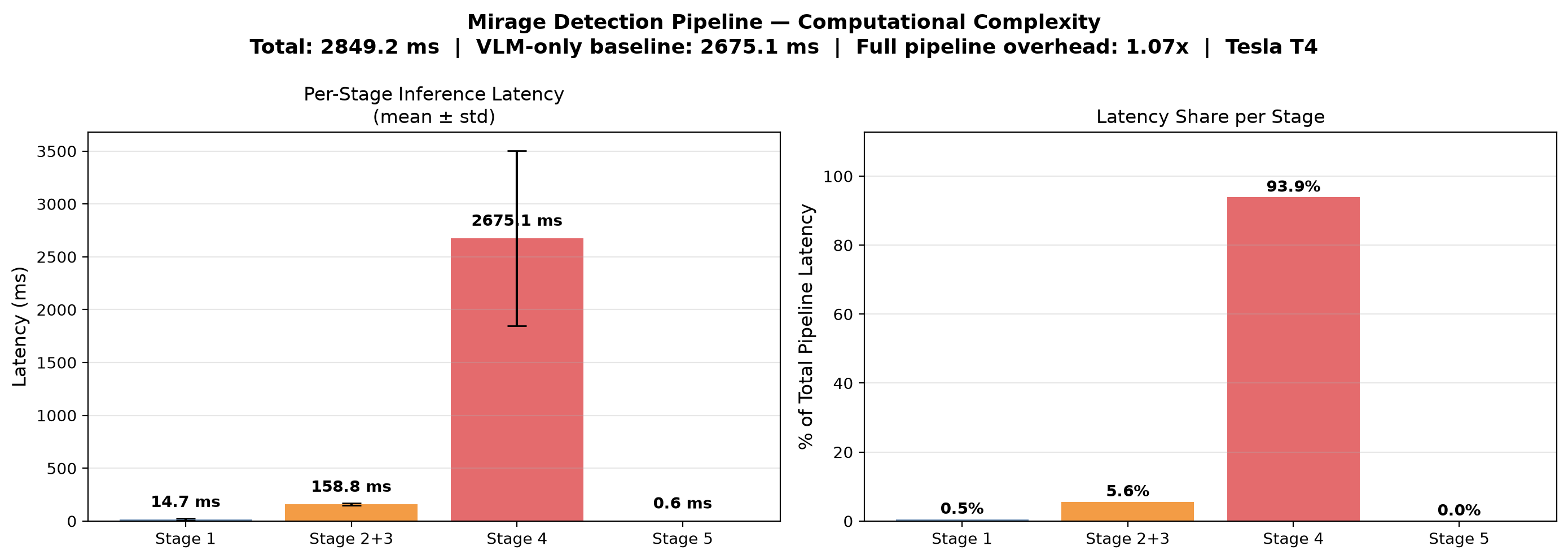}
\caption{Per-stage inference latency (mean $\pm$ std) on a single NVIDIA Tesla T4, with
Qwen2.5-VL-7B-Instruct (NF4) as the reference VLM for the self-assessment stage. The
legend reports each stage's share of total pipeline latency; the detector-side stages
(pixel statistics, CLIP ViT-H/14, XGBoost) together account for 6.5\% of the VLM-only
cost.}
\label{fig:latency}
\end{figure*}

As shown in Figure~\ref{fig:latency}, the VLM-only baseline costs 2,675.1 ms, while the full pipeline (pixel-statistic blank/noise detection, CLIP-based domain routing and TC-LIA probing, the VLM self-assessment, and the XGBoost ensemble) costs 2,849.2 ms, an overhead of only 1.065$\times$. The added components together contribute 174.1 ms (6.5\% of the VLM-only cost): pixel statistics run on the CPU (14.7 ms), the single frozen CLIP ViT-H/14 forward pass takes 158.8 ms, and the XGBoost prediction step is negligible (0.6 ms). The VLM self-assessment stage dominates end-to-end latency regardless of whether a detector is used at all (93.9\% of total pipeline time; real inputs are the slowest self-assessments at 3,361.6 ms on average, since the VLM then generates full answers, versus 2,352.1--2,543.9 ms for unrelated inputs and 2,442.9 ms for blanks). These results indicate that the practical cost of pre-generation mirage detection is small relative to the cost of querying the VLM itself. Absolute latencies naturally scale with the reference backbone and hardware, but the detector-side overhead (one frozen-encoder forward pass plus a 0.6 ms classifier call) is independent of the VLM and stays in the ${\sim}175$ ms range on this hardware.

\section{IAS Weighting Hyperparameter Sensitivity}
\label{app:ias_sensitivity}
\subsection{Weighting Sensitivity}
\begin{figure*}[h]
\centering
\includegraphics[width=\linewidth]{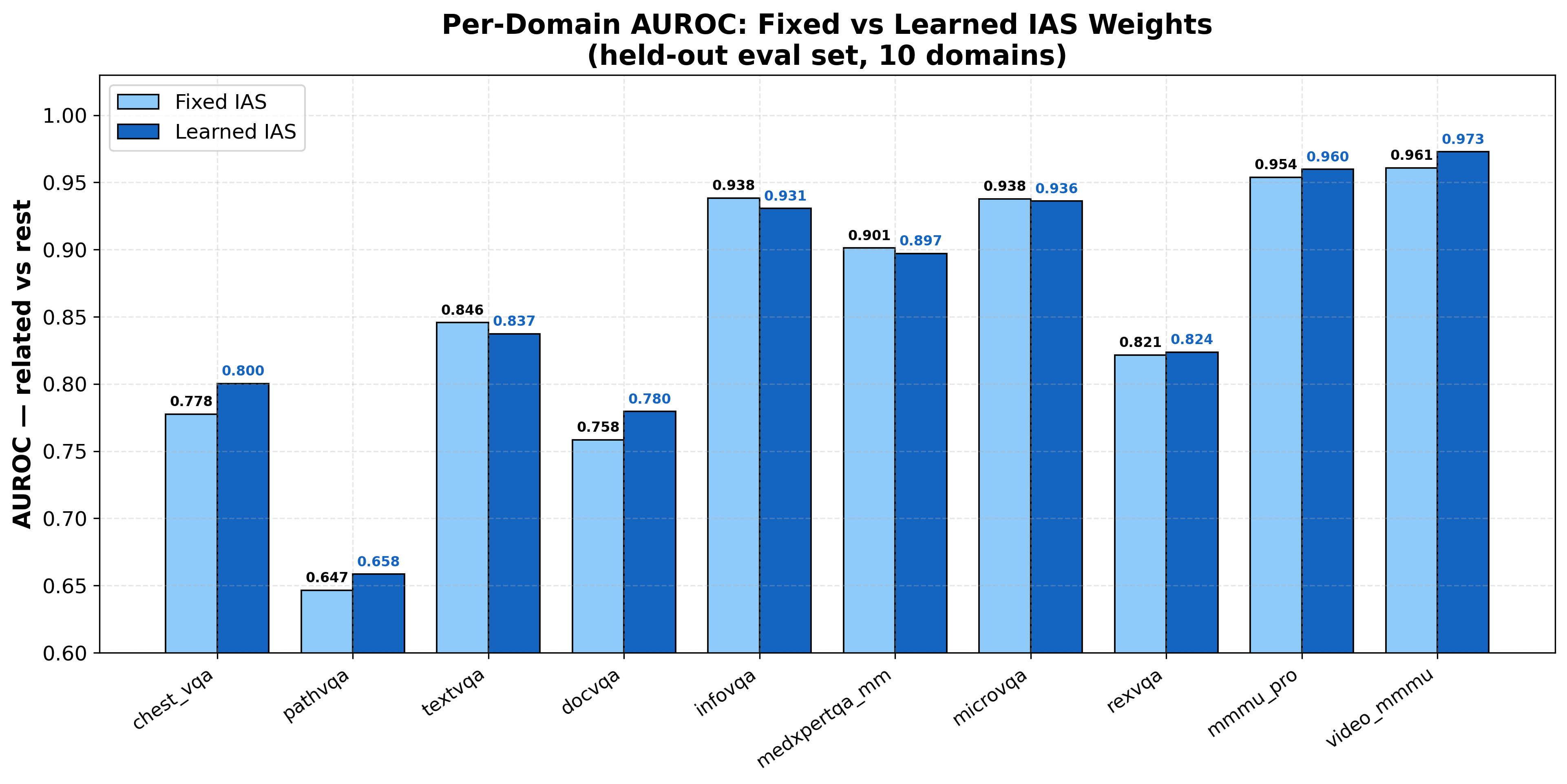}
\caption{Per-domain AUROC for fixed vs. learned IAS weights on the held-out evaluation set.}
\label{fig:ias_sensitivity}
\end{figure*}

The Internal Alignment Score (IAS) combines final cosine similarity, late-layer top-$k$ alignment, early-to-late gain, and layer-wise slope using a fixed, hand-set weighting scheme. To test whether this design choice requires careful manual tuning, we replace the fixed weights with weights learned via logistic regression on the training split, using the same features and held-out evaluation protocol, and compare per-domain AUROC. As shown in Figure~\ref{fig:ias_sensitivity}, the difference between the hand-set weighting and a fully data-driven alternative never exceeds 0.022 AUROC across all 10 domains, with a mean absolute difference of 0.010. This indicates that IAS is not brittle to the specific choice of weights: a learned weighting recovers essentially the same performance as the fixed scheme, so the fixed weights are a reasonable, low-cost default rather than a sensitive hyperparameter.  We also tested the opposite direction---per-domain weights, fitted on each domain's non-blank training rows by a simplex-grid AUROC maximisation (step $0.05$) and frozen for that domain's test rows.  The per-domain optima are inconsistent across domains (e.g.\ pure final-cosine weighting for the document domains versus gain-dominated weights for ReXVQA), and applying them degrades the pooled non-blank AUROC from $0.825$ to $0.782$ (paired $\Delta$ $-0.043$, $p<.001$) because per-domain weighting breaks the cross-domain calibration of the shared score; within-domain gains are negligible (mean $+0.001$).  The fixed global weighting is therefore retained.

\subsection{Sensitivity to Top-$k$ and the Early/Late Layer Split}
\label{app:k_split_sensitivity}
\begin{figure}[h]
\centering
\includegraphics[width=0.85\linewidth]{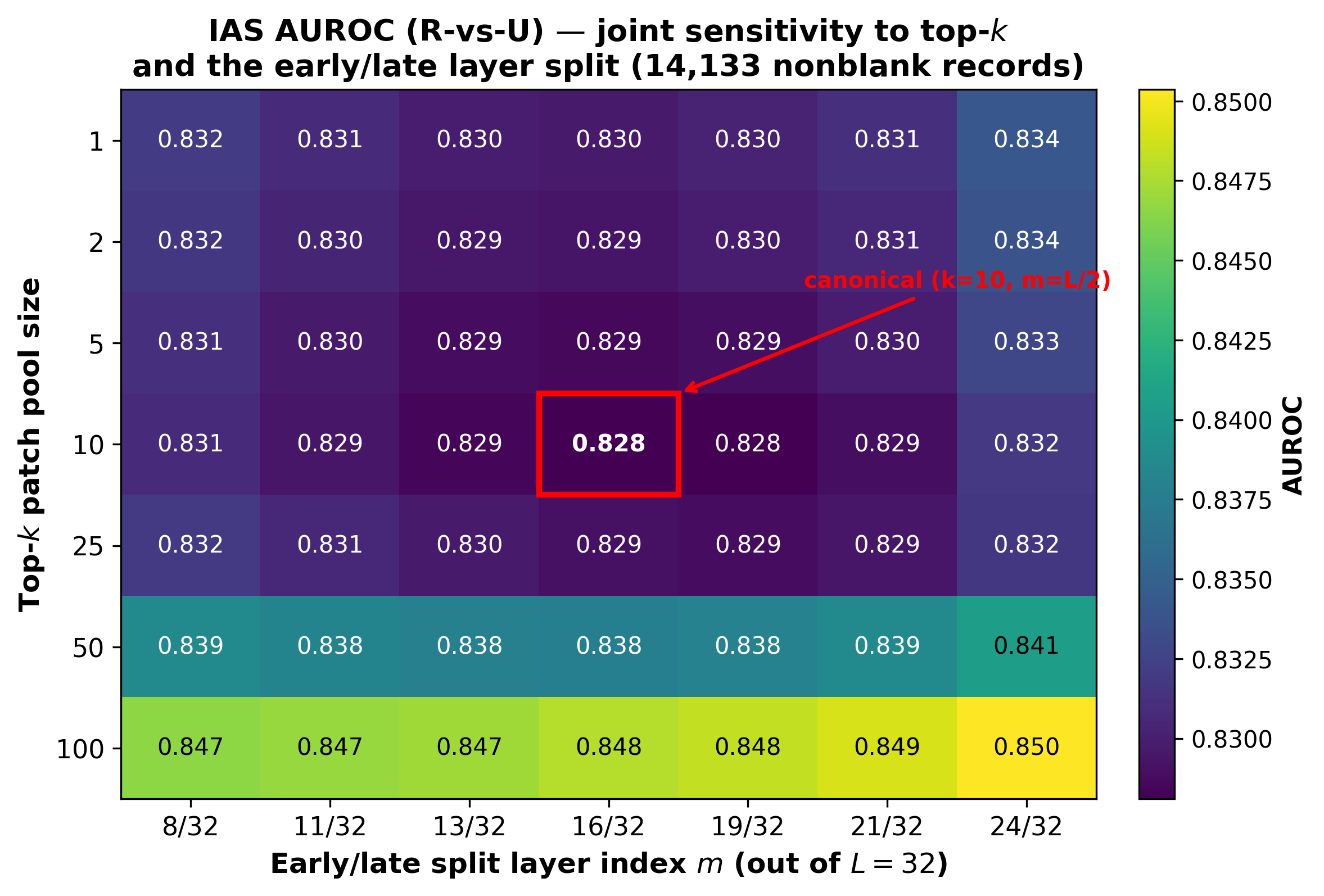}
\caption{\footnotesize IAS AUROC (R-vs-U) on the 14{,}133-record nonblank
subset, jointly sweeping the top-$k$ patch pool size (rows) and the
early/late layer split index $m$ out of $L{=}32$ layers (columns). The
canonical setting ($k{=}10$, $m{=}16$, red box) is the minimum of the grid;
every other cell is within $0.022$ AUROC of it.}
\label{fig:k_split_sensitivity}
\end{figure}

We additionally test sensitivity to the two \emph{structural} hyperparameters
underlying IAS's late/gain/slope features, independent of the weighting
analysis above: the patch pool size $k$ used when averaging the top-$k$
per-layer patch similarities, and the layer index $m$ at which we split the
32 ViT-H/14 layers into an ``early'' block ($<m$) and a ``late'' block
($\geq m$) for the gain and late-layer terms. We sweep $k \in
\{1,2,5,10,25,50,100\}$ and $m \in \{8,11,13,16,19,21,24\}$ (the canonical
setting is $k{=}10$, $m{=}16{=}L/2$) on the pre-registered 14{,}133-record
nonblank subset, recomputing IAS and its domain-stratified bootstrap AUROC
(1{,}000 resamples) for all 49 combinations. As shown in
Figure~\ref{fig:k_split_sensitivity}, AUROC varies by only $0.022$ across the
entire grid (0.828--0.850). The canonical setting is not the best cell but
sits within $0.020$ of the maximum; $k$ has a larger effect than the split
point (larger patch pools, $k{\geq}50$, help slightly), while the split
point alone shifts AUROC by at most $0.006$ at any fixed $k$. IAS is
therefore robust to both hyperparameters, and the hand-set canonical values
are a reasonable default rather than a tuned optimum. 95\% CIs
(domain-stratified bootstrap, 1{,}000 resamples) are $\pm 0.006$--$0.007$
throughout; per-cell CIs and idHN/odHN are in the released artifact
\texttt{k1\_split\_ksweep.csv}.

\section{Additional Diagnostic Experiments}
\label{app:diagnostics}

This appendix reports supplementary diagnostics, all computed on the released
artifacts under the canonical protocol of Appendix~\ref{app:Implementation}
(five-classifier zoo, StratifiedKFold(5), seed 42).  Uncertainty statements
use a domain-stratified bootstrap with $1{,}000$ resamples (seed 42), and
gated rows score at the bottom of the ranking for AUROC computation.

\paragraph{Joint fusion of both alignment encoders.}
The encoder swap (Appendix~\ref{app:encoder_gen}) replaces the CLIP
trajectory features with the SigLIP-SO400M final cosine; here we instead
\emph{add} the SigLIP cosine to the canonical 11 ensemble features as a 12th
input and retrain the full zoo under the identical protocol.  The joint
model improves non-blank AUROC from $0.918$ to $0.929$ (paired $\Delta$
$+0.011$ $[+0.008,+0.013]$, $p<.001$) and deployed AUROC from $0.939$ to
$0.946$ ($\Delta$ $+0.007$, $p<.001$), with accuracy $+0.5$ points
($p=.004$) and macro-F1 $+0.5$ points ($p=.006$); the mirage rate is
unchanged ($5.99\% \to 5.95\%$, $\Delta$ $-0.0005$, $p=.74$).  A
matched-classifier control (the same classifier trained on both feature
sets) confirms the accuracy gain for all five zoo classifiers.  The
improvement therefore lies in ranking and accuracy rather than mirage
suppression, and the released system retains the canonical feature set.

\paragraph{Feature distributions and class separation.}
On the $14{,}133$ non-blank rows the excess kurtosis of every trajectory
feature is consistent with sub-Gaussian tails---between $-0.94$ and $+0.18$
across features and classes (late top-$k$ $-0.70$/$-0.24$, gain
$-0.91$/$+0.18$, slope $-0.94$/$+0.01$, final cosine $-0.24$/$-0.42$, IAS
$-0.54$/$-0.35$ for related/unrelated respectively), with no
heavier-than-Gaussian tails---and class separation is large (Cohen's $d$:
$1.05$ late top-$k$, $0.91$ gain, $0.96$ slope, $1.44$ final cosine, $1.38$
IAS).

\paragraph{Blank/noise gate detail.}
On the $15{,}004$-record test split the gate flags $3{,}843$ inputs ($3{,}751$
true blank/noise plus $92$ real), and over the full $19{,}004$ records it
flags $4{,}871$ ($4{,}751$ true, $120$ real).  Per-domain precision ranges
from $0.906$ (MicroVQA) to $0.994$ (TextVQA) with recall $1.000$ everywhere
(Table~\ref{tab:gate_per_domain}).  The false flags concentrate exactly where
one would expect: low-contrast clinical and scanned content---flagged real
images have median grayscale std $10.2$ versus $51.1$ for passing real
images---with the largest counts in micrographs (MicroVQA, $52$) and
uniform-background documents (DocVQA, $24$).

\begin{table}[t]
\centering
\small
\setlength{\tabcolsep}{4pt}
\renewcommand{\arraystretch}{0.95}
\begin{tabular}{lrrrrrr}
  \hline
  Domain & Records & True B/N & Flagged & FP & Precision & Recall \\
  \hline
  ChestVQA        & 1804 & 451 & 456 &  5 & 0.989 & 1.000 \\
  PathVQA         & 2000 & 500 & 511 & 11 & 0.978 & 1.000 \\
  TextVQA         & 2000 & 500 & 503 &  3 & 0.994 & 1.000 \\
  DocVQA          & 2000 & 500 & 524 & 24 & 0.954 & 1.000 \\
  InfoVQA         & 2000 & 500 & 504 &  4 & 0.992 & 1.000 \\
  MedXpertQA-MM   & 2000 & 500 & 506 &  6 & 0.988 & 1.000 \\
  MicroVQA        & 2000 & 500 & 552 & 52 & 0.906 & 1.000 \\
  ReXVQA          & 2000 & 500 & 504 &  4 & 0.992 & 1.000 \\
  MMMU-Pro        & 2000 & 500 & 506 &  6 & 0.988 & 1.000 \\
  Video-MMMU      & 1200 & 300 & 305 &  5 & 0.984 & 1.000 \\
  \hline
  All             & 19004 & 4751 & 4871 & 120 & 0.975 & 1.000 \\
  \hline
\end{tabular}
\caption{Blank/noise gate per-domain performance over all $19{,}004$ records.
Flagged = images routed to refusal; FP = real images flagged ($120$ over all
records: $40$ \textsc{Related} and $80$ \textsc{Unrelated-Real}; on the test
split, $35$ \textsc{Related} and $57$ \textsc{Unrelated-Real}).  The gate
never misses a blank/noise input (recall $1.000$ everywhere).}
\label{tab:gate_per_domain}
\end{table}

\paragraph{Per-domain confusion and in-domain near-misses.}
On the test split, the deployed system (gate $+$ ensemble threshold at
$p(\textsc{Related})>0.5$) produces the aggregate confusion matrix (rows =
ground truth, columns = prediction):
\textsc{Related} $(2799, 917, 35)$; \textsc{Unrelated-Real} $(899, 6546,
57)$; \textsc{Blank-or-Noise} $(0, 0, 3751)$.  Residual errors concentrate in
\textsc{Related}$\leftrightarrow$\textsc{Unrelated-Real} confusion and in the
in-domain hard-negative band: of the $3{,}751$ test in-domain negatives,
$752$ ($20.0\%$) are predicted \textsc{Related}, but these near-misses are
genuinely similar inputs (mean final cosine $0.270$ vs.\ $0.149$ for
correctly rejected negatives; IAS $0.205$ vs.\ $0.119$), concentrated in
ReXVQA ($266$), MicroVQA ($145$), ChestVQA ($142$), and PathVQA ($86$)---the
same domains where the in-domain-negative AUROC is lowest.

\paragraph{Multi-image/video subset.}
Restricting the deployed system to the four multi-image/video domains
(MedXpertQA-MM, MicroVQA, MMMU-Pro, Video-MMMU; $5{,}600$ test records,
$37.3\%$ of the benchmark) yields $90.7\%$
accuracy, $3.9\%$ mirage rate, and $0.947$ non-blank AUROC, versus
$85.3\%$/$7.2\%$/$0.896$ on the six single-image domains: the first-frame
simplification costs nothing measurable here, though full multi-frame
aggregation remains future work.

\paragraph{Learned selectors and learned per-layer readouts.}
Logistic-regression selective heads trained on the same $2{,}972$ non-blank
training rows---with ground-truth relatedness labels, i.e.\ stronger
supervision than self-labeled correctness signals would
provide---reach at most $0.848$ non-blank AUROC (all five TC-LIA features)
versus $0.918$ for the full ensemble on identical rows (deployed accuracy
$82.3$--$83.8\%$ and mirage rate $7.2$--$7.4\%$, vs.\ $87.3\%$ and
$6.0\%$).  A learned head at this budget thus recovers only a fraction of
the fusion gain, which comes from boosting over the heterogeneous feature
set rather than from the choice of base learner.  Replacing the fixed
terminal projection with a learned per-layer readout (logistic regression on
the 32-dim top-$k$ curve, same budget) reaches $0.846$, or $0.864$ with the
final cosine appended ($+0.021$/$+0.039$ over the fixed IAS, $p<.001$): the
fixed projection leaves some separability on the table, but the learned
variant remains far below the ensemble and sacrifices the training-free
property that defines \textsc{TC-LIA}.

\paragraph{Diagnosing self-assessment transfer.}
Whether the \texttt{vlm\_class\_enc} feature helps or hurts cross-backbone
transfer can be auto-diagnosed from a deployment-available statistic: the
source's \emph{overtrust rate} $P(\text{vlm says \textsc{Related}} \mid
\text{gt} \ne \textsc{Related})$ on its non-blank training rows, which
ranges from $0.140$ (Qwen2.5-VL-7B) to $0.365$ (InternVL2-8B).  Retraining
the zoo without this feature collapses all five source models to a single
backbone-independent transfer system ($84.6\%$ accuracy, $6.1\%$ mirage
rate, macro over the 13 held-out backbones): accuracy improves for four of
five sources (up to $+3.3$ points) and the macro mirage rate drops for all
five ($-0.5$ to $-7.0$ points), at a maximum accuracy cost of $0.6$ points
(InternVL2-8B).  The overtrust rate predicts the per-source benefit of
dropping the feature (Pearson $r=-0.89$ for accuracy, $r=+0.99$ for mirage
rate; $n=5$), so a conservative source self-assessment on the training rows
signals that the feature should be down-weighted or dropped for transfer.

\paragraph{Per-condition complementarity and feature importances.}
Appendix Figure~\ref{fig:complementary}(a,b) contrasts TC-LIA, VLM structured
self-assessment, and their ensemble across the two negative types. For
\textsc{Out-of-Domain} negatives (panel b), TC-LIA is the primary workhorse
($0.913$ overall AUROC, exceeding $0.93$ on seven of ten domains) while VLM
structured self-assessment is substantially weaker ($0.772$ overall), falling to
near-chance on \texttt{rexvqa} ($0.500$) and \texttt{pathvqa} ($0.612$); the
ensemble adds a modest but consistent gain ($0.968$ overall), primarily by
correcting the VLM's failures. For \textsc{In-Domain} hard negatives (panel a),
neither component alone is sufficient---TC-LIA achieves $0.744$ and VLM
self-assessment $0.738$ overall, both near chance on \texttt{rexvqa} ($0.497$
and $0.500$)---and the ensemble lifts overall AUROC to $0.884$, $+0.14$ over
either: the two signals are genuinely complementary for the harder semantic
discrimination task. The residual in-domain misses are near-misses on genuinely
similar inputs: the $752$ test in-domain negatives predicted \textsc{Related}
carry mean final cosine $0.270$ versus $0.149$ for correctly rejected ones (IAS
$0.205$ vs.\ $0.119$). The feature importances (panel c) mirror this split:
in-domain, \texttt{vlm\_class\_enc} dominates (gain $0.585$) with
\texttt{clip\_domain\_enc} second ($0.179$); out-of-domain,
\texttt{vlm\_class\_enc} stays top ($0.444$) but TC-LIA features gain weight
(\texttt{final\_cos} $0.278$, \texttt{internal\_alignment\_score} $0.130$,
together $>40\%$ of gain).

\paragraph{Negative probes rejected as standalone solutions.}
Several approaches were considered and rejected as standalone mirage detectors
(Appendix~\ref{app:negative}), supporting the central claim that detection
requires text-conditioned semantic evidence rather than attention concentration
or output overlap: toy CLIP/BioMedCLIP relevance scoring; GradCAM/attention-only
metrics (Appendix~\ref{appsec:gradcam}); SAM3-style grounding
\citep{carion2025sam3} (Appendix~\ref{app:sam3}); RAPT image/question attention
ratios (Appendix~\ref{rapt-failed}); and small-scale easy datasets where final
cosine performed nearly perfectly.

\section{Use of AI Assistants}
\label{app:ai_assistance}

The complete AI use statement appears in the main paper, immediately before the
Ethics Statement. In brief: generative AI tools assisted with theoretical
critique, code verification and quantitative result checking, literature
identification, and manuscript composition and editing; they were not used to
fabricate or alter experimental observations, and the authors take full
responsibility for the final manuscript and released materials.




\bibliography{anthology,custom}
\bibliographystyle{plainnat}

\end{document}